\documentclass{article}

    \PassOptionsToPackage{numbers, compress}{natbib}
\usepackage[preprint]{neurips/neurips_2026}

\usepackage[utf8]{inputenc} % allow utf-8 input
\usepackage[T1]{fontenc}    % use 8-bit T1 fonts
\usepackage{hyperref}       % hyperlinks
\usepackage{url}            % simple URL typesetting
\usepackage{booktabs}       % professional-quality tables
\usepackage{amsfonts}       % blackboard math symbols
\usepackage{nicefrac}       % compact symbols for 1/2, etc.
\usepackage{microtype}      % microtypography
\usepackage{xcolor}         % colors

\usepackage{graphicx}
\usepackage{subcaption}
\usepackage{comment}
\usepackage{enumitem}
\usepackage{xurl}
\usepackage{amsmath}
\usepackage{algpseudocode}  % for pseudo code
\usepackage{algorithm}

\usepackage{amssymb}
\usepackage{mathtools}
\usepackage{amsthm}
\usepackage{multirow}
\usepackage{array}
\usepackage{wrapfig}
\usepackage{titletoc}

\usepackage[capitalize,noabbrev]{cleveref}

\theoremstyle{plain}

\theoremstyle{definition}

\theoremstyle{remark}

\usepackage{bbm}

\title{Bounding Hallucinations: Merlin-Arthur Protocols\\ for Mutual-Information Bounds in Language Models}
\author{
   Björn Deiseroth\footnotemark[1]\\
   Aleph Alpha Research\\Lab1141 \\
   \And
   Max Henning Höth\footnotemark[1]\\
   Aleph Alpha Research \\ Lab1141\\
   \And
   Kristian Kersting \\
   TU Darmstadt \\ Hessian.AI \\ Lab1141 \\
   \And
   Letitia Parcalabescu\footnotemark[1] \\
   Aleph Alpha Research \\ Lab1141\\
}

\begin{document}

\maketitle

\renewcommand{\thefootnote}{\fnsymbol{footnote}}
\footnotetext[1]{Equal contribution.}
\renewcommand{\thefootnote}{\arabic{footnote}}

\begin{abstract}

Retrieval-augmented generation (RAG) relies on retrieved context to guide large language models (LLM), yet treats the retrieval as a heuristic rather than verifiable evidence -- leading to unsupported answers, hallucinations, and reliance on spurious context.
We introduce a novel evaluation, data augmentation, and training framework that treats the RAG pipeline  as an interactive proof system by adapting the Merlin-Arthur (M/A) protocol: Arthur (the generator LLM) receives context of unknown provenance and Merlin gives helpful evidence, while Morgana injects adversarial, misleading context. 
We implement them both with an XAI method to self-assess and modify evidence most influential to Arthur. 
Based on this framework we propose the Explained Information Fraction (EIF) score, that disentangles explanation fidelity from model predictive errors and imperfect benchmarks, and normalizes M/A mutual-information lower bounds to realistic empirical settings.
When trained with those contexts, Arthur learns to answer when evidence supports the answer and abstain when evidence is insufficient.
Across five QA benchmarks and four LLMs (1B to 32B parameters), M/A reduces incorrect answers under insufficient context by up to 35\,pp during training and by 18-20\,pp over vanilla finetuning. Abstention emerges even \emph{without any manually annotated unanswerable example or preference pair}. We improve $\mathrm{EIF}_{\mathrm{cond}}$ by 0.1-0.4 during M/A training and by 0.33-0.38 over vanilla finetuning. Reusing Merlin/Morgana contexts as automatic hard positives and negatives also raises retriever Recall@1 by 2\,pp. 
While high accuracy does not guarantee entropy flow from context to answer, our EIF scores -- to our knowledge, the first context-to-answer information bound for RAG generators -- show that autonomous interactive-proof-style supervision enables RAG systems that treat retrieved documents as verifiable evidence.

\end{abstract}

\section{Introduction}

Retrieval-augmented generation (RAG) combines a context retriever with a context augmented answer generator. Being increasingly deployed in high-stakes settings with large language models (LLMs) acting autonomously, it must reason reliably over external evidence. Yet current RAG systems often break: (1)~retrievers often struggle selecting the correct document out of similar ones, (2)~generators hallucinate under incomplete or misleading evidence \citep{sun2025redeep,wang2025retrievalaugmentedgenerationconflictingevidence}, (3)~change answers under noise e.g., exchanged or erased tokens \citep{cao2025styleragsfragilitylinguistic}, and (4) rely on spurious parts of context that do not warrant their predictions \citep{wang-etal-2022-identifying}. 
A root cause is that current RAG architectures tend to treat context merely as a heuristic cue, instead of enforcing strict grounding in aligned, verifiable evidence.

To address this, we reframe the RAG pipeline %-- with both retriever and generator -- 
as an interactive proof system
%that guarantees mutually information exchange
by extending the Merlin-Arthur protocol. We probe the generator LLM (Arthur) against (i) a helpful prover (Merlin) supplying valid context, and (ii) an adversarial prover (Morgana) shows misleading evidence, forcing the model to distinguish between reliable and deceptive inputs without knowing the source. This measures robustness against hallucinations and reject behavior even on datasets without annotated reject samples or preferences. 
% We also incorporate Merlin and Morgana masks into retriever training and improve the retriever.
This adversarial formulation establishes a lower bound on the mutual information between context and generated answer. % , providing strict behavioral guarantees.
Specifically, we assess whether the model satisfies (i) \textit{completeness} (answering correctly under valid evidence), and (ii) \textit{soundness} (resisting misleading context). 
% Consequently, the system learns to act as a discerning verifier and answer when the evidence is grounded and reject otherwise, improving robustness and interpretability -- without losing accuracy relative to standard finetuning.

Our contributions are as follows:
\begin{enumerate}[topsep=2pt, itemsep=2pt, parsep=0pt, leftmargin=*]
    % \item \textbf{Extending the M/A Theoretical Framework:\bd{building a lower bound mutual bla}}
    %We extend the M/A framework to retrieval-augmented open-ended language generation. We introduce conditional evaluation to disentangle explanation fidelity from predictive errors; we \textit{propose and measure} the Explained Information Fraction (EIF) to normalize mutual-information lower bounds relative to model capability and imperfect benchmarks.

    \item \textbf{Information-Theoretic Foundations for RAG.} We extend the M/A framework to open-ended retrieval-augmented generation to provide a mutual-information lower bound that quantifies the information flow from retrieved context to generated answer. By introducing a conditional formulation of the bounds, the Explained Information Fraction (EIF) score, we disentangle explanation fidelity from base predictive error, ensuring the bound remains informative on empirical settings with noisy benchmarks and imperfect models.

    \item \textbf{Evaluation Framework.} We use the derived EIF$\in [0,1]$ as an evaluation metric to audit LLM \textit{groundedness}. Unlike existing metrics that rely on LLM-as-a-judge or simple string matching, EIF provides a formal measure of an answer's context-dependence.

    \item \textbf{Autonomous Abstention without Manual Labels.} We re-frame abstention as an emergent property of the interactive proof protocol rather than a supervised task. Unlike standard training that requires vast quantities of (human-)annotated preference pairs to teach a model where to refuse, our framework induces robust rejection behavior autonomously. By using the adversarial Morgana prover to mask decisive evidence, we teach the model to identify insufficient context and abstain without a single manually annotated ``unanswerable'' sample.

    \item \textbf{Adaptive M/A Training and Feedback Loops.} We use M/A as a training framework where Merlin and Morgana provers operate on-the-fly, ``patching'' model weaknesses as the LLM evolves to increase soundness and groundedness. We further close the loop from generation to retrieval by using the converged generator to produce automated hard positives and negatives, creating a model-driven feedback loop that improves both generator behavior and retriever recall, underlining the quality of the augmentations.
    
    %\item \textbf{M/A generator training:} We use Merlin- and Morgana-generated contexts to adversarially train generator LLMs with \textit{increased soundness, groundedness and explicit reject behavior}, without annotated unanswerable questions or preference data.

    %\item \textbf{M/A retriever training:} We feed Merlin- and Morgana-generated contexts -- produced by a converged generator -- back into retriever training, forming a feedback loop from generation to retrieval, yielding automated hard positives and hard negatives to improve the retriever. \bd{merge with point 3}
\end{enumerate}

Across a diverse range of datasets, model architectures and sizes (1B to 32B), we consistently improve groundedness, completeness, soundness, and retriever performance. Specifically, we observe:
\begin{enumerate}[topsep=2pt, itemsep=2pt, parsep=0pt, leftmargin=*]    
    \item \textbf{Higher robustness, fewer hallucinations:} The LLM learns to abstain when retrieved context does not support an answer and be more robust to misleading evidence. The reject behavior emerges even \textit{without explicit unanswerable questions or preference data} and improves upon instructing LLMs to reject under insufficient evidence. Under insufficient context, M/A training reduces overall incorrect answers by up to 35\,pp relative to an instructed LLM and by 18-20\,pp compared to vanilla training, yet matches it in accuracy (model utility). The retriever learns to separate useful from harmful context better than standard training; Recall@1 improves by 2\,pp across datasets.
    
    \item \textbf{Improved interpretability with mutual-information bounds:} We obtain information exchange bounds between context and answer of \mbox{$\textrm{EIF}_\textrm{cond} \geq 0.55$} on 5/6 datasets. Specifically, we improve $\mathrm{EIF}_{\mathrm{cond}}$ by 0.1-0.4 over  instructed LLMs and by 0.33-0.38 over vanilla training. The increased scores translate into observable model behavior: while baseline model attributions often misalign with evidence, M/A models reflect grounded reasoning. We will release our code upon publication.

\end{enumerate}

\section{Related Work}

\textbf{Reject Behavior in RAG.}
A growing body of work aims to elicit abstention from LLMs, including RAG \citep{10.1162/tacl_a_00754}.
Methods elicit refusal via prompting \citep{madhusudhan-etal-2025-llms, peng-etal-2025-unanswerability} with mixed success, or via uncertainty estimation \citep{tomani2024uncertaintybasedabstentionllmsimproves, lavi2025detectingunanswerabilitylargelanguage}, which remains unreliable because it is hard to calibrate LLM uncertainty.
\textit{Training} for refusal behavior uses annotated unanswerable questions, e.g., from SQuAD2.0 \citep{rajpurkar-etal-2016-squad}, Natural Questions \citep{kwiatkowski-etal-2019-natural}, or handcrafted rejection signals derived from dispreferred responses \citep{xu2024rejection, muhamed2025refusalbenchgenerativeevaluationselective}, requiring human (preference) annotations that may be unavailable in many domains. In contrast, our approach trains strong reject behavior without annotated unanswerable data: Morgana generates contexts where rejection is optimal, and Arthur learns to abstain when evidence is insufficient.

\textbf{Adversarial Training and Robustness.}
% we can put this  back for the camera ready:
% Adversarial methods have long been used to evaluate robustness in NLP \citep{jia-liang-2017-adversarial, alzantot-etal-2018-generating, wallace-etal-2019-universal} and
Recent work shows that RAG systems remain brittle to even mild linguistic variations \citep{cao2025styleragsfragilitylinguistic, wang-etal-2025-tricking, zeng2025rareretrievalawarerobustnessevaluation}. In RAG, adversarial supervision for the retriever has been explored via model-informed hard negatives for retrievers \citep{karpukhin-etal-2020-dense} and contrastive training \citep{izacard2022unsupervised}, as well as through adversarial perturbations to improve robustness \citep{EMREERGUN2025114056, HUANG2025493, app131810148, rafiei-asl-etal-2024-robustsentembed}. However, these approaches primarily treat adversariality as generic input noise or data augmentation; in contrast, we adopt an interactive-proof perspective in which adversarial and supportive contexts are generated in a model-dependent manner, adapt as the generator evolves, and are subsequently used to supervise the retriever -- closing a feedback loop from generation back to retrieval. Via this framing, we directly encourage (i) mutual information exchange between context and answer, (ii) grounding in causal rather than correlational features, and (iii) induce reject behavior.

% More recent work also studies adversarial perturbations to improve robustness in RAG systems \citep{EMREERGUN2025114056, HUANG2025493, app131810148, rafiei-asl-etal-2024-robustsentembed}, but these approaches treat adversariality as generic input noise or data augmentation, rather than enforcing evidence-faithful generation or explicit reject behavior.
% it does not use an interactive-proof framing,
% it does not reason about grounding, completeness/soundness, or mutual information,
% it typically treats adversariality as input perturbation or robustness augmentation, not as evidence faithfulness or reject behavior,
% supervise with the same method the retriever and the generator (not jointly though) with targeted model-dependent adversariers, adapting as the model trains, injesting it's own ouput so to say.

% we should not say the things below anymore
% Our work differs in two important ways: (i) The adversary modifies the context, not only the retrieval distribution, and (ii) we integrate adversarial supervision into the generator and the retriever with Merlin and Morgana providing targeted positive and negative training examples for both components.

\textbf{Merlin-Arthur Interactive Proof Theory.}
We build on the Merlin-Arthur (M/A) classifiers of \citet{waldchen2024interpretability}, a game-theoretic protocol inspired by interactive proof systems \citep{arora2009computational}.
Unlike heuristic attribution methods, M/A provides  interpretability  by bounding the mutual information between selected features and  class label via measurable test-set quantities (completeness and soundness). % we already say this in relwork: % , without requiring assumptions on the underlying data distribution -- cf. §\ref{sec:theoretical_foundations} for details.
% LP: we go into detail a lot later in section 3, I've kept it short here. Summarised this above.
% The authors adopt the Merlin-Arthur classifier, a framework designed to provide provable interpretability guarantees for deep learning models, such as neural networks. Unlike heuristic methods (e.g., saliency maps) which can be manipulated to hide biases, this approach uses a game-theoretic setup inspired by Interactive Proof Systems.
% The authors derive a quantitative lower bound on the Mutual Information between the selected features and the class label. This bound relies on measurable test-set metrics (Completeness and Soundness) and does not require modeling the complex data distribution, which is a common failure point in other formal interpretability methods.
We extend M/A (i)  from image classification to language and RAG, (ii) from static features to dynamic retrieved contexts, and (iii) introduce conditional evaluation via the Explained Information Fraction (EIF) to normalize mutual-information lower bounds by model capacity and disentangle explanation fidelity from prediction errors.
% While the authors target multi-label image classification with fixed visual features, we extend in several key ways: (i) From images to language: we adapt the M/A protocol to text, LLMs, retrieval and \textit{open-ended} generation. (ii) From static features to dynamic contexts: RAG introduces retrieved (and possibly manipulated) evidence, requiring context-level rather than feature-level guarantees. (iii) Conditional evaluation: We introduce conditioning on model correctness to disentangle explanation fidelity from baseline predictive errors and Explained Information Fraction (EIF) as a normalized measure that contextualizes theoretical guarantees relative to model capacity.

\textbf{Explanation Methods (XAI)} such as gradient-, attention-based saliency maps, and perturbation techniques \citep{pmlr-v70-sundararajan17a, Selvaraju_2017_ICCV, Chefer_2021_ICCV}, aim to identify which input tokens influence an LLM prediction.
Most relevant to our work are erasure and rationale methods that remove or select minimal subsets \citep{li2017understandingneuralnetworksrepresentation, lei-etal-2016-rationalizing} as well as counterfactual editing methods \citep{ross-etal-2021-explaining, wu-etal-2021-polyjuice}, which alter inputs and observe behavioral changes. 
% We do not need to bash on all of these, we motivate the choice of Atman twice later (section 3 and method section). Let's save space by being nice.
% However, most approaches capture local sensitivities or lack faithfulness, making them unsuitable for the counterfactual, evidence-removal reasoning required by the M/A protocol.
We build on \textsc{AtMan} \citep{deiseroth2023atman}, a linear-cost perturbation-based explainer that measures the effect of masking the attention of individual tokens on the model's probabilities. We use it to approximate the global counterfactual effect needed for Merlin and Morgana masking strategies. Crucially, existing explainability methods are mostly diagnostic, whereas our work uses XAI to train RAG systems using XAI-derived proofs as actionable feedback to inform generator and retriever training.

% \section{Theoretical Framework}
% \section{Foundations}
% \section{Prerequisites}
% \section{Fundamental Concepts}
% \section{Preliminary Work}
\begin{figure*}
    \centering
    \includegraphics[width=.9\linewidth, trim={0 1.05em 0 0.405em}, clip]{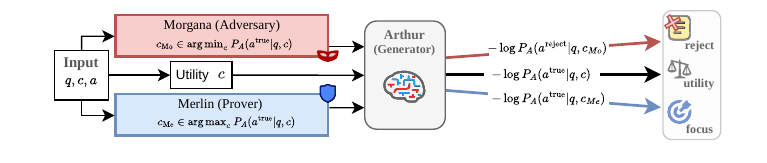}
    \caption{\textbf{Generator-LLM Training via automated M/A provers.} For each query, context, answer $(q, c, a)$ triplet, we use M/A provers to produce the $c_{\text{Me}}$ context to convince the Generator ($A$) of the correct answer and the $c_{\text{Mo}}$ context to fool $A$ into answering incorrectly. $A$ evaluates contexts $c$, $c_{\text{Me}}$, and $c_{\text{Mo}}$ separately to compute three losses. M/A training uses a weighted sum of these losses.}
    \label{fig:ma}
\end{figure*}

\section{Theoretical Foundations and Background} \label{sec:theoretical_foundations}
% LP: I know it is not nice to start a section with the subsection; but the sentence below just repeats the section header. Let's include a nice preamble in the camera ready, but save the space for review now. Thus, commenting this line out:
% We begin with the necessary theoretical foundations.

% \subsection{Verified Mutual Information Exchange via the Merlin-Arthur Protocol} % this takes 2 lines, shorter version below
\subsection{Merlin-Arthur Mutual Information Exchange}
We summarize the M/A Classifier framework of \citet{waldchen2024interpretability}, before extending it to the language and RAG (Fig.~\ref{fig:ma}) domain. It is a game-theoretic interpretability framework providing quantitative lower bounds on feature quality without requiring explicit modeling of the data distribution.
% we already say this in the related work
% inspired by Interactive Proof Systems~\citep{arora2009computational}
%the system comprises three agents: a cooperative selector \textbf{Merlin} $(M)$, an adversarial selector \textbf{Morgana} $(\hat{M})$, and a verifier \textbf{Arthur} $(A)$.

\textbf{Definitions and Agents.}
Let $D$ be a dataset from an underlying distribution $\mathcal{D}$ and a ground truth class map $\hat{y}: D \to \{-1, 1\}$.
% We denote the subset of data points belonging to a specific class $l \in \{-1,1\}$ as $D_l = \{x \in D \mid \hat{y}(x) = l\}$, and the corresponding restricted distribution as $\mathcal{D}_l$.
For each class $l \in \{-1,1\}$, we write $D_l = \{x \in D \mid \hat{y}(x) = l\}$ for the set of points with label $l$, and $\mathcal{D}_l$ for the induced distribution restricted to $D_l$.
The setup comprises three agents:
\begin{itemize}[topsep=2pt, itemsep=2pt, parsep=0pt]
    \item \textbf{Merlin} (Prover, Me) selects a feature subset $\text{Me}(x) \subseteq x$ to convince the verifier of the correct class $\hat{y}(x)$.
    \item \textbf{Morgana} (Adversary, Mo) attempts feature selection $\text{Mo}(x)$ that makes Arthur not predict the correct class.
    \item \textbf{Arthur} (Verifier, $A$) is a classifier that takes partial features $\tilde{x}$ 
    and outputs a class or abstains, i.e. $A(\tilde{x}) = \arg\max_y P_A(y \mid \tilde{x})$, with $A(\tilde{x}) \in \{-1,0,1\}$.
\end{itemize}

% Merlin selects a subset of features on the input $M(x) \subseteq x$ to convince Arthur of the correct class $c(x)$, while Morgana attempts to select features $\hat{M}(x)$ that fool Arthur into predicting the incorrect class $\neg c(x)$. Arthur can output a specific class or abstain (``Don't Know'') if the evidence is insufficient.

\textbf{Mutual-Information Interpretability Lower Bounds }
are expressed via testable errors~\citep{waldchen2024interpretability}:
%The framework's interpretability guarantees rely on two measurable metrics evaluated on a test set~\citep{waldchen2024interpretability}:
\begin{itemize}[topsep=2pt, itemsep=2pt, parsep=0pt]
    \item \textbf{Completeness Error ($\epsilon_c$):} The probability that Arthur fails to classify correctly given Merlin's features,
    $\epsilon_c = \max\nolimits_{l \in \{-1, 1\}} \mathbb{P}_{x \sim \mathcal{D}_l} [A(\text{Me}(x)) \neq \hat{y}(x)]$.
    % \label{eq:completeness}
% \end{equation*}
    \item \textbf{Strict Soundness Error ($\epsilon_s$):} The probability that 
    Arthur fails to defend against Morgana, i.e., does not abstain,
% \begin{equation*}
    $\epsilon_s = \max\nolimits_{l \in \{-1, 1\}} \mathbb{P}_{x \sim \mathcal{D}_l} [A(\text{Mo}(x)) \neq 0]$.
    % \label{eq:soundness}
% \end{equation*}
\end{itemize}

We propose a definition of soundness that is explicitly stricter than those  established in literature, that tolerate classifications of the correct class, to cover for imperfect Morgana errors. Because LLMs frequently and effectively memorize benchmarks, this stringent measure is necessary to ensure that models cannot rely on ill-formed or spurious patterns to recover samples.

% Under the assumption of Asymmetric Feature Correlation $\kappa$ and relative prover strength $\alpha$, as discussed in App.~\ref{sec:data_distribution}, 
The Average Precision $Pr_{\mathcal{D}}(M)$ of the selected features is lower-bounded by Theorem 2.11 of~\citet{waldchen2024interpretability} and common discussion of system parameters as in App.~\ref{sec:data_distribution}:
\begin{equation}
\label{eq:upper_bound_short}
    Pr_{\mathcal{D}}(M) \gtrsim 1 - \epsilon_c - \frac{\epsilon_s}{1 - \epsilon_c + \epsilon_s}\ .
\end{equation}
This precision bound directly constrains the \textbf{Mutual Information} (I) between the features and the class label via the binary entropy function $H_b$ and entropy $H$:
% Doesn't fit in the new narrow ICML columns
% \begin{equation}
% \label{eq:MI}
%     \mathbb{E}_{y \sim \mathcal{D}}[I_{x \sim \mathcal{D}}(c(x); M(y) \subseteq x)] \ge H_{x \sim \mathcal{D}}(c(x)) - H_b(Pr_{\mathcal{D}}(M))\ .
% \end{equation}
\begin{equation}
\label{eq:MI}
\begin{aligned}
\mathbb{E}_{y \sim \mathcal{D}}\!\left[
    I_{x \sim \mathcal{D}}\!\left(\hat{y}(x);\, \text{Me}(y) \subseteq x\right)
\right]\ge
H_{x \sim \mathcal{D}}\!\left(\hat{y}(x)\right)
- &H_b\!\left(Pr_{\mathcal{D}}(M)\right).
\end{aligned}
\end{equation}
In summary, low values of $\epsilon_c$ and $\epsilon_s$ guarantee with Eq.~\ref{eq:upper_bound_short}, \ref{eq:MI} that the features exchanged by the agents possess high mutual information with the ground truth class.

%\subsection{Relative Information Fidelity for Challenging Benchmarks}
\subsection{Relative Information Fidelity for Benchmarking}
While \citet{waldchen2024interpretability} applied the M/A framework directly, standard considerations often yield insubstantial bounds when the baseline model's accuracy is already low, effectively entangling \emph{explanatory failure} with \emph{predictive failure}. In such cases, poor explanation scores may reflect model error rather than a lack of fidelity, a common effect in noisy or inherently difficult benchmarks. These limitations in completeness are expected, as they arise from the stochastic nature of open-ended generation, intrinsic task difficulty and noise, and the finite capacity of the underlying model.

To address this, we introduce two theoretical extensions: (i) a Conditional Evaluation Protocol that isolates explanation fidelity from predictive performance,
and (ii) the Explained Information Fraction (EIF) that normalizes certified information gain relative to the model's predictive capability.

\textbf{Conditional Evaluation Protocol.}
To disentangle explanation fidelity from the predictive capability of the base model, we restrict evaluation to the inputs on which Arthur produces reliable predictions. 
% Let $A: \mathcal{X} \to \mathcal{Y}$ denote the classifier (Arthur) and $\mathcal{D}$ the data distribution over $\mathcal{X}$. 
Let $\mathcal{D}_{\text{acc}}$ be the conditional distribution of $x$ over correctly predicted inputs $\hat{y}(x)$:
%\begin{equation*}
$\mathcal{D}_{\text{acc}} := \mathcal{D}\big( x \mid A(x) = \hat{y}(x) \big).$
%\end{equation*}
% All Merlin-Arthur metrics are estimated under $\mathcal{D}_{\text{acc}}$.
\emph{Conditional completeness error} $(\tilde{\epsilon}_c)$ measures the probability that the explanation fails to support the correct class, conditioned on $\mathcal{D}_\text{acc}$.
%Arthur predicting the correct label on the full input.
Analogously, we define \emph{conditional soundness error} $(\tilde{\epsilon}_s)$.
%as the probability that the explanation fails to rule out incorrect classes.
% \begin{equation*}
%     \tilde{\epsilon}_c = \max_{l \in \{-1, 1\}} \mathbb{P}_{x \sim \mathcal{D}_{\text{acc}} \cap D_l}\big[ A(M(x)) \neq \hat{y}(x) \big],\
%     \tilde{\epsilon}_s = \max_{l \in \{-1, 1\}} \mathbb{P}_{x \sim \mathcal{D}_{\text{acc}} \cap D_l}\big[ A(\hat{M}(x)) = \neg \hat{y}(x) \big]\ .
% \end{equation*}
% \begin{equation*}
% \begin{aligned}
% \tilde{\epsilon}_c
% &= \max_{l \in \{-1,1\}}
% \mathbb{P}_{x \sim \mathcal{D}_{\text{acc}} \cap D_l}
% \big[ A(\text{Me}(x)) \neq \hat{y}(x) \big]\ ,
% \\
% \tilde{\epsilon}_s
% &= \max_{l \in \{-1,1\}}
% \mathbb{P}_{x \sim \mathcal{D}_{\text{acc}} \cap D_l}
% \big[ A(\text{Mo}(x)) = - \hat{y}(x) \big]\ .
% \end{aligned}
% \end{equation*}
% This conditioning isolates explanation fidelity from the model's own predictive errors: 
The resulting mutual-information lower bounds based on these errors quantify how faithfully the selected features support Arthur's \emph{correct} predictions, without being confounded by regions where Arthur is inaccurate. Finally, we define \emph{Model Coverage} as the mass of the reliable region:
%\begin{equation*}
$C := \mathbb{P}_{x \sim \mathcal{D}}\big[ A(x) = \hat{y}(x) \big]$,
%\end{equation*}
which we estimate via the accuracy on the evaluation set.
%the fraction of evaluation examples for which Arthur predicts the correct answer.

\paragraph{Explained Information Fraction.}
\label{sec:EIF}

To contextualize the strength of our derived bounds, we introduce the Explained Information Fraction (EIF), an information-theoretic analogue to the coefficient of determination ($R^2$). While $R^2$ measures the proportion of variance explained by a regression model, EIF quantifies the proportion of the baseline classifier's predictive signal that is provably captured by the explanation features. We define EIF as the ratio of the certified mutual information lower bound to the empirical mutual information of the baseline model:
\begin{equation}
    \label{eq:eif}
\text{EIF} = \frac{I_{guaranteed}}{I_{baseline}} \approx \frac{1 - H_b(Pr_{\mathcal{D}}(M))}{1 - H_b(C)}\ .
\end{equation}
The approximation $I_{baseline} := I(Y; \hat{Y}) \approx 1 - H_b(C)$ holds specifically in our setting, and is discussed in detail in App.~\ref{sec:data_distribution}. 
With the Conditional Evaluation Protocol described above it yields:
\begin{equation} 
\label{eq:eif_cond}
\text{EIF}_{\text{cond}} \approx 1 - H_b(\tilde{\epsilon}_{\text{eff}}), \: \text{where } \tilde{\epsilon}_{\text{eff}} \approx \tilde{\epsilon}_c + \frac{\tilde{\epsilon}_s}{1 - \tilde{\epsilon}_c + \tilde{\epsilon}_s}\ . 
\end{equation}

\textbf{Practical Implications.} 
To illustrate the tightness of these bounds, consider a balanced dataset where $H(\hat{y}(x)) \approx 1$ bit. If we measure completeness and soundness of 90\% each -- values we exceed for the conditioned versions in our experiments -- we obtain $H_b(Pr_{\mathcal{D}}(M))\approx 0.72$ bits. This yields with Eq.~\ref{eq:MI} (Eq.~\ref{eq:eif_cond} resp.) $0.28$ bits of mutual-information lower bound of information about the class residing within the selected features -- which is an important lower bound on mutual information exchange. 
% Even a more modest completeness and soundness of 85\% yields a $0.12$ bit guarantee.
A model coverage of $90\%$ would almost double this value to $\text{EIF}=0.52$.

% For a baseline classifier with 70\% accuracy on a balanced task with symmetric errors ($I_{baseline} \approx 0.12$ bits), a certified bound of 0.03 bits (e.g., at 80\% completeness and soundness) yields $\text{EIF}=0.03 / 0.12=0.25$ (Eq.~\ref{eq:eif}). Thus, the selected features capture at least 25\% of the model's total predictive information, providing a verified floor for explanation fidelity independent of task difficulty.
% Furthermore, we can directly measure the conditional mutual information exchange: a conditional completeness and soundness of 85\% leads to $\text{EIF}_{\text{cond}} = 0.12$, and for 90\% we obtain 0.28, which are significant lower bounds capturing the reasons for the model's generations.

\subsection{RAG Setting}
We consider the RAG pipeline with (i) \emph{context retrieval} ($q \rightarrow c$) where a retriever selects context $c$ for a query $q$ and (ii) \emph{answer generation} ($c \rightarrow a$), where an LLM generates an answer conditioned on $c$.
We use \emph{context} broadly to include documents, partial evidence, or even reasoning traces, all of which condition generation. Our goal is to certify information exchange between context and generated answer. We consider as hallucinations any non-reject answers unsupported by context evidence.

To use the M/A protocol on open-ended generation, we frame the multi-class prediction task as a binary verification problem: the positive class is the ground-truth answer, and all other outputs are treated as incorrect. This groups all hallucinations into a single negative class, preserving the protocol's completeness and soundness evaluations.
% This abstraction groups all potential hallucinations into a single ``incorrect'' state, allowing the protocol's original binary guarantees for completeness and soundness to remain mathematically valid. 

% \subsection{Using \textsc{AtMan} as a Linear Complexity Masker (Prover) for RAG Use Cases}
\subsection{Linear Complexity Prover for RAG via \textsc{AtMan}}\label{sec:linear-complexity-prover}

% In RAG, we aim to make models answer faithfully based on provided context, rather than hallucinate from parametric knowledge. %However, deploying the Merlin-Arthur protocol to verify this faithfulness faces a significant computational bottleneck: estimating the completeness and soundness errors naively requires evaluating Arthur on all possible combinatorial subsets of the context $x$, which scales exponentially with the number of tokens. 
Applying the M/A protocol directly is computationally hard, as estimating completeness and soundness through optimal Merlin and Morgana masks requires evaluation of Arthur over all -- exponentially many -- context subsets. 
% Using the M/A protocol directly is computationally hard, as finding the optimal mask requires Arthur to search over exponentially many mask subsets.
%To address this, we adopt the \textsc{AtMan} explainability (XAI) method of~\cite{deiseroth2023atman}. This perturbation-based approach utilizes token-based attention masks to measure changes in log-probabilities, conceptually akin to influence functions. Specifically, the authors demonstrated that \textsc{AtMan} effectively quantifies how specific tokens contribute in support of or contradiction to a target generation. Crucially, it requires only one probe per token (linear complexity) and allows search at various granularities, such as the token or sentence level. This serves as a valid approximation to the otherwise exhaustive tree search, allowing us to obtain a context-saliency map for a specific target token efficiently. Ultimately, this approach models the strategies of Merlin and Morgana at equal strengths.
We address this via \textsc{AtMan}~\citep{deiseroth2023atman}, a perturbation-based XAI method that measures token influence via attention masking. It requires a single isolated probe (linear complexity), supports token- and sentence-level analysis, and approximates the otherwise exhaustive subset search, enabling efficient context saliency estimation. This allows Merlin and Morgana to operate with comparable strength, while gradient-based XAI methods~\citep{pmlr-v70-sundararajan17a, Selvaraju_2017_ICCV} may capture entangled feature dependencies and do not directly model the counterfactual removals required by the M/A protocol.

% While gradient-based XAI methods \citep{pmlr-v70-sundararajan17a, Selvaraju_2017_ICCV} offer efficiency, they strictly measure local sensitivity, failing to capture the global counterfactual impact of information removal required by the Merlin-Arthur (M/A) protocol without further adapting the XAI method. 

% Finally, analyzing the M/A protocol through an XAI method introduces an inherent tradeoff between two confounding factors: the \textit{fidelity of the explanation} and the \textit{robustness of the verifier} (Arthur). Because neither the explainer nor Arthur is perfect, any evaluation reflects a mixture of both effects. It is therefore useful to distinguish three perspectives under which results can be interpreted:
Using XAI within the M/A framework introduces an inherent tradeoff between \emph{explanation fidelity} and \emph{verifier robustness}, since neither explainer nor Arthur is perfect. Thus results can be interpreted from three perspectives:
% Switch back to itemize in camera ready
% \begin{itemize}[topsep=2pt, itemsep=2pt, parsep=0pt]
%     \item \textbf{XAI Bounds:} Explanation quality assuming an ideal verifier -- isolates the fidelity of the explainer.
%     \item \textbf{Verifier Robustness:} Arthur's soundness assuming a perfect explainer -- isolates the verifier robustness.
%     \item \textbf{System Performance:} Practical regime reflecting explainer and Arthur imperfection -- reflects the real tradeoff between explanation fidelity and model robustness.
% \end{itemize}
(1) \textbf{XAI Bounds:} Explanation quality assuming an ideal verifier -- isolates the fidelity of the explainer.
(2) \textbf{Verifier Robustness:} Arthur's soundness assuming a perfect explainer -- isolates the verifier robustness.
(3) \textbf{System Performance:} Practical regime reflecting explainer and Arthur imperfection -- reflects the real tradeoff between explanation fidelity and model robustness.

\textbf{Optimality of Fixed-$x$ \textsc{AtMan} and Theoretical Headroom.}
We reduce finding of optimal M/A masks to probing with an XAI method at a fixed mask size. For any heuristic $h$, with target mask size $x$, -- our \textsc{AtMan} instantiation included -- the implied EIF satisfies:
\begin{equation}\label{eq:eif-monotonicity-main}
    \mathrm{EIF}^{h}_{x}
    \;\le\;
    \mathrm{EIF}^{\text{exh}}_{x}
    \;\le\;
    \mathrm{EIF}^{\text{exh}}_{\star}\ ,
\end{equation}
where the right-hand side is the protocol optimum. The numbers reported in this paper are thus \emph{conservative}: tighter heuristic search at the same budget, or relaxing the budget itself, can only widen the bound. 
% Empirically, we observe a $x{\approx}0.6$ peak dataset-intrinsic across models, datasets, and training regimes (§\ref{sec:mask_findings}), so we fix $x{=}0.6$ as the operating point. 
Further discussion, including parity-constrained per-sample budgets, is in App.~\ref{app:bound_optimality}.

\section{Methods for Bounding Hallucinations}

We extend the M/A protocol to RAG and use it as a \emph{training framework} for generator and  retriever.
% We build on the Merlin-Arthur (M/A) protocol and extend it to the RAG setting to train (a) the Generator and (b) the Retriever to behave reliably under both helpful and adversarial evidence. Our goal is to shape a system that for the generator (i) answers only when the context justifies an answer, (ii) rejects when evidence is insufficient or misleading, and (iii) focuses its attention on the evidence that actually supports its predictions. Likewise the retriever should only select context that is relevant for the generator to answer to the prompt.
% %yielding more interpretable behavior - which we analyse via input attribution. %this is method

% The M/A protocol involves three parties: Arthur (\texttt{A}), Merlin (\texttt{Me}), and Morgana (\texttt{Mo}). Arthur is the model being trained; Merlin is a \emph{helpful} prover that supplies evidence to help Arthur answer correctly; and Morgana is an \emph{adversarial} prover whose goal is to mislead Arthur. Arthur does not know which prover supplied the context and is trained to answer correctly.

\subsection{Merlin-Arthur Training for RAG}\label{sec:ma-for-rag}

We train Arthur, an autoregressive LLM, to (i) answer correctly ($a^{\text{true}}$) when provided helpful context, (ii) reject ($a^{\text{reject}}$) or avoid incorrect answers when receiving misleading context.

\textbf{Arthur} ($A$), an autoregressive LLM, receives a question $q$ and context $c'$, which may be the original retrieval $c$ or a Merlin or Morgana variant; the provenance of $c'$ is unknown to Arthur.
We compute Arthur's answer probabilities $P_A\big(a
 \mid q, c'\big)$ and accuracies under teacher forcing: after generating the first token, we condition on the ground-truth answer prefix to read the predicted next-token distribution.
\textbf{Merlin} (Me), the helpful prover, masks the retrieved context $c$ (cf.~§\ref{sec:context-masking}) to produce $c_{\text{Me}}$, maximizing Arthur's probability of the correct answer $a^{\text{true}}$. 
\textbf{Morgana} (Mo), the adversarial prover, masks the context $c$ to produce $c_{\text{Mo}}$ to fool Arthur, i.e., answers that are not the correct sequence:
\begin{equation}\label{eq:me-mo-goals}
c_{\text{Me}} \in \arg\max_{c} P_A(a^{\text{true}} \mid q, c), \quad c_{\text{Mo}} \in \arg\min_{c} P_A(a^{\text{true}} \mid q, c)\ .
\end{equation}
%\begin{equation} % The equivalence of min and max formulations are obvious, showing just one
%\label{eq:morgana-goal}
%c_{\text{Mo}} \in {\arg\min}_{c}\; \Sigma_{a \in \{a^{\text{true}}, a^{\text{reject}}\}}
%P_A\big( a \mid q, c \big)\ .
%\end{equation}
During training, Arthur minimizes a weighted sum of three cross-entropy losses:
\begin{equation}
\label{eq:arthur-loss,eq:morgana-loss-in-arthur-training}
\mathcal{L}_A
=
\lambda_{\text{util}}
\,\mathcal{L}_{\text{util}}
+
\lambda_{\text{Me}}
\,\mathcal{L}_\text{Me}\
+
\lambda_{\text{Mo}}\,\mathcal{L}_{\text{Mo}}\ , 
\end{equation}
with (1) the utility loss over the original context $c$ as $\mathcal{L}_{\text{util}}=-\log P_A(a^{\text{true}} \mid q,c)$, (2) Merlin loss over context $c_{\text{Me}}$ as $\mathcal{L}_{\text{Me}} = -\log P_A(a^{\text{true}} \mid q,c_{\text{Me}})$, and
(3) Morgana loss over  $c_{\text{Mo}}$ as \mbox{$\mathcal{L}_{\text{Mo}}
    = - \log P_A(a^{\text{reject}} \mid q,c_{\text{Mo}}) 
        $}. More details in Algorithm~\ref{alg:rag_training}.
% For Morgana, we treat both $a^{\text{true}}$ and $a^{\text{reject}}$ as valid targets -- rationale in App~\ref{app:why-not-just-reject}.
% \begin{equation*}
%    with utility loss $\mathcal{L}_{\text{util}}=-\log P_A(a^{\text{true}} \mid q,c)$, Merlin loss 
%    $\mathcal{L}_{\text{Me}} = -\log P_A(a^{\text{true}} \mid q,c_{\text{Me}})$, 
%    and Morgana loss \mbox{$\mathcal{L}_{\text{Mo}}
%    = \tfrac{1}{2}\Sigma_{a \in \{a^{\text{true}}, a^{\text{reject}}\}}
%        (- \log P_A(a \mid q,c_{\text{Mo}})) 
%        .$}
% \end{equation*}

\textbf{M/A-Specific Evaluation.} M/A induces three evaluation criteria of Arthur’s behavior under helpful, adversarial, and unmodified contexts at test time:
% Switch back to enum in camera ready
% \begin{enumerate}[topsep=2pt, itemsep=2pt, parsep=0pt, leftmargin=*]
%     \item \textbf{Groundedness}: the ground-truth answer is in the provided context -- measured separately for Me and Mo.
%     \item \textbf{Completeness}: Arthur answers correctly when Merlin provides supportive context.
%     \item \textbf{Soundness}: Arthur avoids wrong answers under  Morgana's adv. contexts by answering correctly or rejecting.
% \end{enumerate}
(1) \textbf{Groundedness}: the provided Me/Mo context supports the ground-truth answer (App.~\ref{app:rag-metrics}). 
(2) \textbf{Completeness}: Arthur answers correctly when Merlin provides supportive context.
(3) \textbf{Soundness}: Arthur avoids wrong answers under  Morgana's adversarial contexts by rejecting.
Together, these three criteria characterize a RAG system that answers when evidence justifies an answer and rejects otherwise, yielding models that are accurate (complete), robust (sound), and evidence-aligned (grounded).

% \begin{comment}
% \begin{figure}
%     \centering
%     \includegraphics[width=.9\linewidth]{draft_figure/architecture/merlin.pdf}
%     \caption{Inside workflow of Merlin. \mh{needs to be adjusted for new formula} \lp{This needs to prettier and clearer to make it into the appendix. Commenting it out atm, doesn't add anything compared to the pseudocode; does not ease cognitive effort to understand the method.}}
%     \label{fig:merlin}
% \end{figure}

% \begin{figure}
%     \centering
%     \includegraphics[width=.9\linewidth]{draft_figure/architecture/morgana.pdf}
%     \caption{Inside workflow of Morgana. \mh{needs to be adjusted for new formula}\lp{This needs to prettier and clearer to make it into the appendix. Commenting it out atm, doesn't add anything compared to the pseudocode; does not ease cognitive effort to understand the method.}}
%     \label{fig:morgana}
% \end{figure}
% \end{comment}

\subsection{Context Masking}\label{sec:context-masking}

% In this section, we specify concrete implementations of Merlin, and Morgana which transform an input context $c$ into new a context $c_{\text{Me}}$ or $c_{\text{Mo}}$ that changes Arthur's answer probabilities toward their respective goals - defined in Equations \ref{eq:merlin-goal} and \ref{eq:morgana-goal}.  
% We give both provers access to the original context $c$ and let them \emph{mask} a fixed fraction $x\%$ of the context in either a helpful (Merlin) or adversarial (Morgana) manner.
Merlin and Morgana transform a context $c$ into $c_{\text{Me}}$ and $c_{\text{Mo}}$ by masking a  token fraction $x\%$ to respectively support or adversarially disrupt Arthur’s prediction (Eq.~\ref{eq:me-mo-goals}) as follows (cf. Algorithm~\ref{alg:masking}).

\textbf{Masking Granularity.}
Naively, one could mask tokens independently.  
However, to use the sentence structure of natural language, we support both: 1)~\textbf{token-level masking}: each token constitutes a candidate mask unit, and 2)~\textbf{sentence-level masking}: tokens from the same sentence are a mask unit.
We use the term `unit' to refer to either individual tokens (token-level masking) or sentences (sentence-level masking), depending on the chosen granularity.

\textbf{Selection Strategy.}
Merlin and Morgana identify the most influential units to mask via \textbf{top-$k$ selection}, providing a good balance between runtime and effectiveness.
For each unit $i$ in a context $c$ of length $N$, we mask only $i$ to obtain $c_i$ and evaluate Arthur.

\textit{Merlin scores} via $p^{\text{Me}}_i = P_A(a^{\text{true}} \mid q,c_i)\ $ and
% switch back to equation env in camera ready
% \begin{equation}
%     p^{\text{Me}}_i = P_A(a^{\text{true}} \mid q,c_i)\ .
% \end{equation}
\textit{Morgana} via: $p^{\text{Mo}}_i = 1 - P_A(a^{\text{true}}\mid q,c_i)\ $.
% switch back to equation env in camera ready
% \begin{equation}
%     p^{\text{Mo}}_i = 1 - P_A(a^{\text{true}}\mid q,c_i) - P_A(a^{\text{reject}} \mid q,c_i)\ .
% \end{equation}
We select the $k$ highest-scoring unit positions, where $k$ represents the masking ratio $x\%$:
\begin{equation}
    v_{\text{Me/Mo}} = \mathrm{topk}\big([p^{\text{Me/Mo}}_0, \ldots, p^{\text{Me/Mo}}_N],\, k\big)\ ,
    \label{eqn:topk}
\end{equation}
Me and Mo construct their masked contexts via 
$c_{\text{Me}/\text{Mo}} = \mathrm{mask}(c, v_{\text{Me}/\text{Mo}})$, with masking as follows.

% This approach provides a good balance between runtime and effectiveness: masking the top-$k$ influential units reliably shifts Arthur's behavior toward Merlin's and Morgana's respective goals.

\textbf{Masking Strategies} \label{masking_strats}
define how to produce $c_{\text{Me}}$ and $c_{\text{Mo}}$ from the selected unit set $v$.
Rather than replacing tokens with a fixed \texttt{[MASK]} token (e.g., `...'), we use a more principled alternative via \textbf{attention-level masking via \textsc{AtMan}}:
% \paragraph{String replacement.}
% \mh{We are not doing this anymore -> remove}
% The simplest approach replaces tokens of each selected unit with a fixed mask token \texttt{[MASK]}, such as `...':
% \begin{equation}
% c_j =
% \begin{cases}
% \texttt{[MASK]} & \text{if } j \in v, \\
% c_j & \text{otherwise}.
% \end{cases}
% \end{equation}
% 
% This approach is straightforward but can introduce artifacts: the replacement token may itself have meaning (e.g., `...') or an atypical embedding, and inserting it repeatedly can shift the model's predictions for reasons unrelated to the removed content.
%\textbf{Attention Masking (\textsc{AtMan}).}
% To avoid altering the input semantics with artificial \texttt{[MASK]} tokens and preserve the original positional embeddings, we follow   \cite{deiseroth2023atman} and apply attention-level masking.  
% This approach leaves the input tokens unchanged but modifies the attention scores so that masked tokens cannot influence the model's computation. Finally, the authors demonstrated that this approach works for both the objectives of Merlin, and Morgana.
Following \citet{deiseroth2023atman}, we use attention-level masking to avoid introducing artificial \texttt{[MASK]} tokens and to preserve positional embeddings. 
Tokens remain, but their attention contributions are suppressed, preventing masked tokens from influencing the computation. 
The target influence is measured, supporting both Merlin’s and Morgana’s objectives.

Specifically, \textsc{AtMan} implements discrete token removal by adding column-wise suppression at the respective columns for token positions of $v$, $\mathbf{M}^{\textrm{token;}v}$ next to the causal mask:
\begin{gather*}
\mathbf{M}^{\textrm{causal}}_{ij} = 
\begin{cases} 
-\infty & \text{if } j > i,  \\
0 & \text{otherwise}
\end{cases}\;,\; 
\mathbf{M}^{\textrm{token;}v}_{ij} = -\infty \cdot \mathbb{I}(j \in v)\ ,
\end{gather*}
% \begin{equation}
\begin{align}
    \text{Attention}(\mathbf{Q}, \mathbf{K}, \mathbf{V}&; \mathbf{M}) = \text{softmax}\left(\mathbf{Q}\mathbf{K}^\top / \sqrt{d_k} + \mathbf{M} \right) \mathbf{V}, \notag 
    \text{where }  \mathbf{M}&=\mathbf{M}^{\textrm{causal}} + \mathbf{M}^{\textrm{token;}v}\ .
    \label{eqn:attention}
\end{align}
% \end{equation}

\subsection{Retriever Training with M/A-Generated Contexts}\label{sec:retriever-training}

We extend the M/A framework to form a feedback loop from generation to retrieval (cf. Fig.~\ref{fig:rag}). We include automatically generated Merlin and Morgana contexts from a converged generator as \emph{hard positives and negatives} in retriever training -- at no manual annotation cost. Full details in App.~\ref{app:retriever_training}.
% as additional supervision signals. Because the M/A training already produces these manipulated contexts as part of the RAG objective, they serve as task-specific hard positives and hard negatives examples at no manual annotation cost -- since they reflect how the generator behaves under helpful and adversarial evidence.

\section{Experimental Evaluation}
\label{sec:empirics}
% We evaluate our approach on five retrieval QA benchmarks spanning single-hop, multi-hop, and open-domain QA; short and long context. Our experiments test whether M/A training improves (i) robustness to misleading context, (ii) reject behavior reducing hallucinations, and (iii) attribution plausibility. We additionally evaluate the impact of M/A supervision on the retriever.

\textbf{Models.}
We use
Llama-3.2-1B-Instruct,
Llama-3.2-3B-Instruct,
Qwen3-4B-Instruct, Qwen2.5-32B-Instruct as advanced baselines that have undergone alignment training. See App.~\ref{app:generator-models} for model references.
For retrieval, we use a pre-trained BERT-style model, \textit{granite-embedding-small-english-r2}~\citep{awasthy2025graniteembeddingr2models}.

\textbf{Datasets.}
We evaluate on
(i) SQuAD2.0~\citep{rajpurkar-etal-2016-squad,rajpurkar-etal-2018-know} (single-hop; contains unanswerable questions),
(ii) HotpotQA~\citep{yang2018hotpotqa} (multi-hop),
and (iii) TriviaQA~\citep{2017arXivtriviaqa} (long context open-domain with noisy retrieval; contains unanswerable questions),
(iv) 2WikiMultihopQA~\citep{ho-etal-2020-constructing} (multi-hop with evidence reasoning),
and (v) MuSiQue~\citep{trivedi-etal-2022-musique} (complex multi-hop reasoning; contains unanswerable questions)
--  in App.~\ref{app:datasets} we describe the datasets in detail. 

\textbf{Prompts.}
All LLMs use the same instruction prompt shown in App.~\ref{app:prompt}, fixed across baselines and M/A training, and explicitly requesting rejection when the answer is unsupported.
% We use the same instruction prompt shown in App.~\ref{app:prompt} for all models and training conditions, including the baseline and M/A-trained variants. Since all models are instruction-tuned, this prompt explicitly defines the expected reject behavior: we instruct the model  to answer only when the answer is supported by the provided context and to output \texttt{Reject} otherwise. This prompt remains fixed throughout training and evaluation.

\textbf{Metrics.}
For generator training, we report exact-match accuracy and probability, (conditional variants of) completeness, soundness, and EIF.
We additionally report \textit{groundedness} as a secondary metric for post-hoc analysis. We use answer-span annotations for SQuAD2.0, annotated supporting-fact sentences for HotpotQA, and gold-answer string matching for the remaining datasets -- details in App.~\ref{app:rag-metrics}.
For the retriever, we report Recall@$k$ and Mean Reciprocal Rank (MRR), measuring whether and how highly the gold document is retrieved -- details in App.~\ref{app:retriever-metrics}.
% For the retriever we report Recall@$k$ and Mean Reciprocal Rank (MRR). Recall@$k$ measures how often the correct context appears within the top-$k$ retrieved results, while MRR is the mean reciprocal rank of the correct context across all queries. -> moved to appendix

\textbf{M/A Training Details RAG Generator.}
% For each dataset, we use the standard train/validation/test splits. In TriviaQA we truncate the context to a maximum sequence length of 5,000 bytes.
% We use standard train/val/test splits; we truncate TriviaQA contexts to 5{,}000 bytes.
For each sample, we construct three contexts: the original $c$, Merlin-masked $c_{\text{Me}}$, and Morgana-masked $c_{\text{Mo}}$ (§\ref{sec:context-masking}) with a masking ratio $x=0.6$ (cf. Mask Findings §\ref{sec:mask_findings}).
% [WE ALREADY SAY THIS IN METHODS] We apply teacher forcing during training to compute next-token probabilities for $a^{\text{true}}$ and $a^{\text{reject}}$.
Arthur trains with the loss defined in §\ref{sec:ma-for-rag} and $\lambda_{\text{Me}}=0.65$,
$\lambda_{\text{Mo}}=0.10$, $\lambda_{\text{util}}=0.25$, (selected based on small sweeps).
Baselines use standard finetuning ($\lambda_{\text{util}}=1$, rest 0).
% For comparison, we also train baseline models without Merlin or Morgana supervision, i.e.,
% standard finetuning in which only the original context contributes to the loss
% ($\lambda_{\text{orig}}=1,\ \lambda_{\text{Me}}=\lambda_{\text{Mo}}=0$).
We fine-tune for a maximum of 200 steps with LoRA \citep{hu2022lora} (rank 8, $\alpha=16$, dropout $=0$) for one epoch -- except MusiQue -- in \texttt{bfloat16}, AdamW \citep{loshchilov2018decoupled} and a $10^{-3}$ learning rate.
% We train on 64~GPUs and set the local masking batch size to~8, i.e., each Arthur worker first generates Merlin and Morgana masked contexts for 8 input samples before any Arthur updates occur. These masked variants are then consumed in local training batches of~2 for the actual parameter updates of Arthur. This delayed-update structure enforces partial offline training: the masks are generated using a slightly older version of Arthur, preventing the masking process from chasing every transient fluctuation in Arthur's parameters.
We train on 64 GPUs with effective batch size 128. Each GPU generates Me/Mo masks for 8 samples before parameter updates. 
% This has an off-policy regularization effect by training on masks from slightly stale Arthur versions --  preventing the masking from chasing transient fluctuations in Arthur's parameters.

\textbf{M/A Training Details Retriever.}
We compute the contrastive loss (Eq.~\ref{eq:infonce}), with dataset-specific document pools described in App.~\ref{app:retriever_training-neg-pool}. For M/A training, we replace two negatives with  Morgana variants (if soundness is 0 or Arthur predicts $a^{\text{reject}}$ with the Morgana input) at $x\!\in\!\{0.6,0.8\}$ and replace two negatives  with positive Merlin variants (if completeness is 1) at $x\!\in\!\{0.3,0.6\}$, using a converged M/A-trained generator (ablations in Fig.~\ref{fig:retriever_ablation}). Full details in App.~\ref{app:retriever_training}.

\textbf{Convention.} 
We hereafter show sentence-level attention masking for its efficiency and robustness. Comprehensive comparison results are in App.~\ref{app:rag-training-results}.

\begin{figure*}[t]
    \centering
    \includegraphics[width=\linewidth, trim={0.7em 0.7em 0.7em 0.7em}, clip]{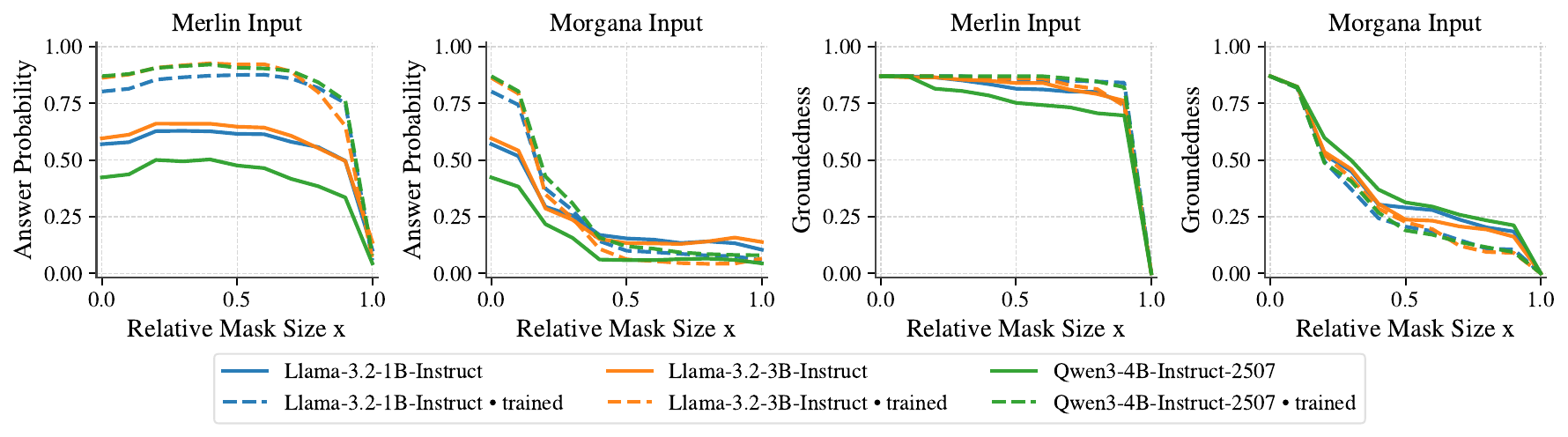}
    \caption{\textbf{Me increases accuracy and groundedness, Mo decreases them.}
    %, indicated by flatter and monotone curves for both Me/Mo, as discussed in text.
    %-- and Mo finding it more difficult to induce incorrect answers. 
    Sentence masking on HotpotQA; 
    left: Arthur’s prob. for $a^{\text{true}}$ under Me/Mo contexts with mask size~$x$;
    right: groundedness. % of Me/Mo contexts;
    %solid lines: instruct models; dashed lines: after M/A training.
    }
    \label{fig:evaluate_maskers_squad_sentence_before_after_training}
% \vspace{-4pt}
\end{figure*}

\subsection{Operational Masking Ratio Validation and Relevance Of Reject Supervision}
% \subsection{Mask Findings and Operational Masking Ratio Validation} %$ Evaluation Results}
\label{sec:mask_findings}
\textbf{The $x=0.6$ operational point is dataset-intrinsic.}
Sentence-level masking results on HotpotQA are in
Fig.~\ref{fig:evaluate_maskers_squad_sentence_before_after_training} (Fig.~\ref{fig:eval_masker_before_after} shows more metrics and SQuAD). 
Ideally, answer probability as well as groundedness should decrease monotonically with mask size, but remain higher for Merlin (reflecting evidence preservation) and drop early for Morgana (reflecting evidence removal).
We observe this trend consistently across models and the studied datasets.
Based on these findings, we set the masking ratio to $x=0.6$, where Merlin reliably maximizes and Morgana minimizes answer probability.

\begin{figure*}[!htbp]
        \centering
        \includegraphics[width=.9\linewidth, trim={0 1.1em 0em 1.1em}, clip]{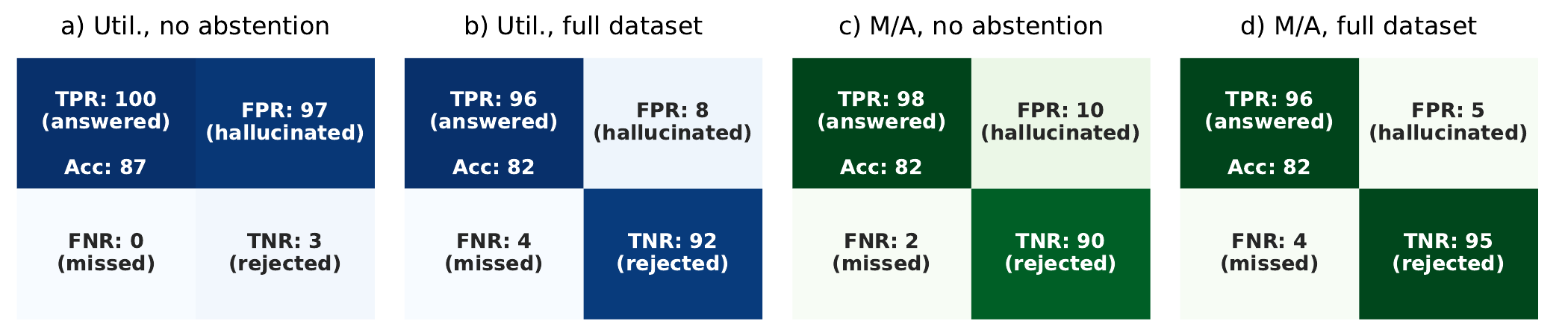}
     
    \caption{\textbf{M/A masks carry enough discriminative signal to erase hallucinations even when training with correct samples only,} while standard finetuning requires enough annotated abstain samples. Confusion Matrices for training Qwen2.5-32B on SQuAD, non-abstention and full data split.}
    \label{fig:atman_filter_peak_sec5}
\end{figure*}

\textbf{Training without abstention samples collapses rejections; M/A substitutes reject supervision.}
We ablate 20-step fine-tunings (2,560 samples) on SQuAD (1) the full dataset and (2) an ``answerable-only'' subspace lacking abstention examples. In Fig.~\ref{fig:atman_filter_peak_sec5} presents the abstention confusion matrices and overall accuracies. TPR refers to an answer being given (not necessarily the correct one) when relevant context is present; and FPR when the respective context was missing, i.e. a ``hallucination''.  
If finetuned solely on the answerable subspace using utility loss Fig.~\ref{fig:atman_filter_peak_sec5}(a), the model tends to hallucinate. While this failure is traditionally mitigated by including a large fraction ($\approx 40\%$) of explicit ``abstain'' samples (Fig.~\ref{fig:atman_filter_peak_sec5}b), M/A-augmented samples provide sufficient discriminative signal to achieve comparable results using only answerable inputs (Fig.~\ref{fig:atman_filter_peak_sec5}c).
M/A training on the full dataset further optimizes these results, allowing for precise control over the TPR/FPR trade-off via the loss weights, here $\lambda_\text{util}=\lambda_{{\text{Mo}}}=\lambda_{{\text{Me}}}=0.33$. Note the higher accuracy in (a) may be a misleading artifact of forced guessing: by never abstaining, the model occasionally hits the correct answer by chance.

%feature selection becomes discriminative with a peak, again, around $60\%$. A comparable trend is visible for supervised finetuning with $\approx 40\%$ reject samples (orange).

\subsection{M/A Protocol for Generator Training}
\label{subsec:ma-generator-training}
\begin{figure*}[t]
    \centering
    \includegraphics[width=\linewidth, trim={0.7em 0.7em 0.7em 0.7em}, clip]{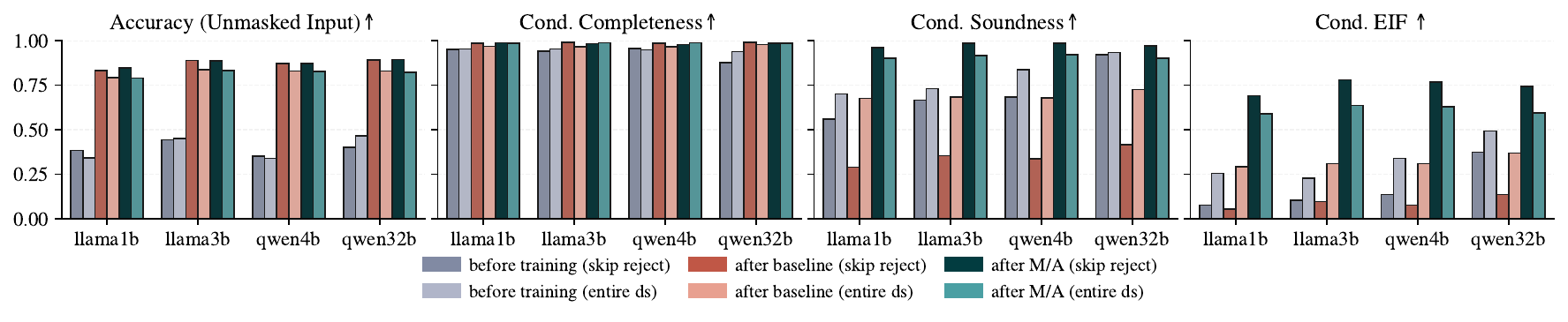}
    \caption{\textbf{With M/A training, completeness, soundness and EIF increase.} Results for all models averaged over all datasets at three stages: (1) before training, (2) after M/A training, (3) after baseline finetuning. Accuracy measured on the data split in brackets. See Fig.~\ref{fig:RAG_training_barplot_full} for metrics per dataset.}
    \label{fig:RAG_training_barplot_avg}
\end{figure*}

\begin{figure*}[ht]
    \centering
    \includegraphics[width=\linewidth, trim={0.7em 0.7em 0.7em 0.7em}, clip]{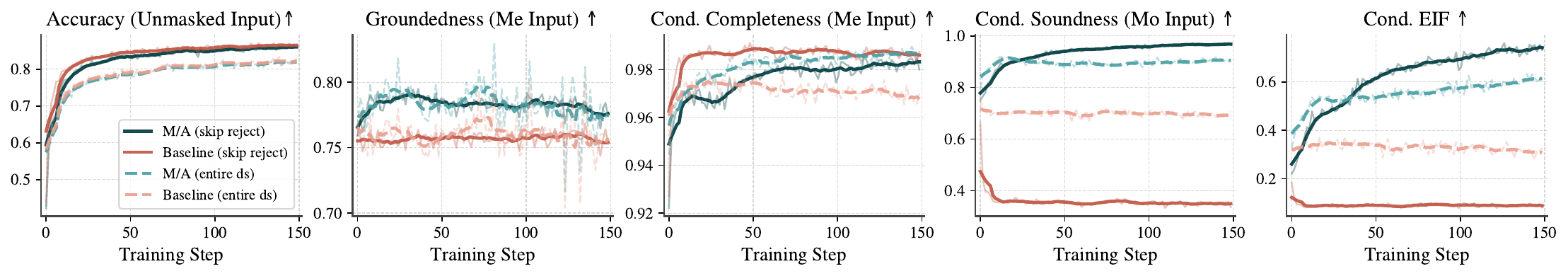}
    \caption{\textbf{M/A leads to measurable lower-bounds on datasets, especially without reject annotations.} 
    % Sentence-level M/A vs baseline finetuning for Llama-3.2-1B on HotpotQA. Curves smoothed with a rolling-window average.
    Training metrics averaged over all models and datasets. Step~0 is \emph{before} any model updates, reflecting initial performance. Results with all models, metrics and datasets
    % in Fig.~\ref{fig:RAG_train_curves_hotpot_app}. SQuAD and TriviaQA results 
    are in Figs.~\ref{fig:RAG_train_curves_squad_app}-\ref{fig:RAG_train_curves_musique_app_qwen}.}
    \label{fig:RAG_train_curves_main}
\vspace{-7pt}
\end{figure*}

We compare M/A training to two baselines: (i) the LLM \emph{before} finetuning and (ii) after standard finetuning. Figs.~\ref{fig:RAG_training_barplot_avg},\ref{fig:RAG_train_curves_main} show  accuracy, (conditional) groundedness, completeness, soundness and EIF. App. Figs.
% \ref{fig:RAG_train_curves_squad_app}-
~\ref{fig:RAG_training_barplot_full}-\ref{fig:RAG_train_curves_musique_app_qwen} show full results, training  curves, reject rates, loss components and ablations in
\ref{app:rag-training-results}.

\textbf{M/A preserves Model Utility.}
Across datasets, the M/A-trained model matches vanilla SFT accuracy on unmasked input, showing that M/A training retains model utility and -- as outlined below -- they \textit{add robustness and make the model grounded in useful evidence} under stable training.
% This -- combined with the soundness and completeness results -- confirms that M/A training preserves the model's core QA capability while adding robustness under stable training as shown by the loss.

\textbf{Soundness Increases.}
Errors under Morgana inputs correspond to hallucinations -- answers produced without sufficient evidence. M/A training markedly improves soundness: Arthur learns to avoid incorrect answers and abstain under adversarial masking, while the baseline remains highly fragile.
% and often produces incorrect answers when Morgana removes or occludes evidence.
Notably, \textbf{abstention emerges even when the dataset contains no unanswerable samples}
(c.f. HotpotQA and  2WikiMultihopQA or ``skip reject'' of datasets in Fig.~\ref{fig:RAG_training_barplot_full}).
% This demonstrates that M/A training substantially increases robustness to misleading or insufficient evidence.
 The same effect is established under controlled conditions on SQuAD via the answerable-only ablation in \S\ref{sec:mask_findings} (Fig.~\ref{fig:atman_filter_peak_sec5}d,e).

\textbf{Completeness Increases under M/A Training.}
With Merlin’s supportive contexts, Arthur completeness increases. While baselines may achieve higher completeness by exploiting spurious cues, they do so at the expense of soundness, reflecting  the inherent \textbf{soundness--completeness trade-off}:
increasing answerability raises hallucination risk, while enforcing soundness promotes rejection.
% improving completeness (i.e., answering more questions correctly when helpful evidence is present) typically reduces soundness, since pushing the model to answer more often increases the risk of accepting misleading or insufficient evidence. Conversely, enforcing high soundness naturally lowers completeness by encouraging the model to reject more frequently.

% \textbf{Reject Rate}:
% On unmanipulated inputs, the M/A model's reject rate converges toward the dataset’s natural frequency of unanswerable examples.
% The stable behavior of the M/A model reflects that the reject token is used appropriately on HotpotQA and just slighlty over-triggered on SQuAD, consistent with the training objective that encourages rejection only when evidence is unreliable.

\textbf{Controlled Grounded Behavior under M/A training.} The M/A models have higher groundedness with Merlin input compared to the baseline (Fig.~\ref{fig:RAG_train_curves_main}), thus M/A training leads to stronger grounding in evidence. This is directly reflected in more interpretable attribution patterns. 
In addition, Fig.~\ref{fig:atman_hotpot_sentence} shows \emph{sentence}-level \textsc{AtMan} attributions on HotpotQA: before training, the model fails to identify either supporting sentence; after vanilla finetuning, it selects only one; after M/A training, both required sentences are correctly highlighted, indicating improved grounding.

\textbf{M/A yields Measurable Mutual Information Exchange Bounds.}
In §\ref{sec:EIF}, we introduced the Explained Information Fraction (EIF) as a practical proxy for Mutual Information Exchange in realistic empirical setups -- where benchmarks, models and XAI methods are imperfect. As shown in Fig.~\ref{fig:RAG_train_curves_main}, \textbf{we obtain strong lower-bound information exchange, when training with autonomously augmented data under M/A}, even in the presence of only positive samples. M/A reaches $\textrm{EIF}_{cond} \approx 0.6$, while the baseline remains target-focused.
% These results indicate that the model learns to align itself to answer faithfully to the context, through the new objectives of interacting with the XAI methods. This, otherwise, can only be achieved through large amounts of diverse and explicitly labeled reject cases -- and still insufficient to induce aligned and plausible attribution patterns, as evidenced by the baseline explanations in Fig.~\ref{fig:atman_squad_mc_token} and the groundedness analysis.
%This shows that
Thus, M/A training aligns answers with context via XAI-driven interactions, achieving evidence based and faithful behavior that would otherwise require large amounts of explicitly labeled reject data -- data still being insufficient to yield plausible attributions, as shown by attributions of baselines in Fig.~\ref{fig:atman_hotpot_sentence} and groundedness in Fig.~\ref{fig:RAG_train_curves_main}.

\subsection{M/A Protocol for Retriever Training}\label{subsec:retriever-results}
\begin{wrapfigure}{r}{0.6\textwidth} 
% \begin{figure}[h!]
    \centering
     \vspace{-15pt}
     \includegraphics[width=\linewidth,trim=0cm 0.9cm 0cm 1.5cm, clip]{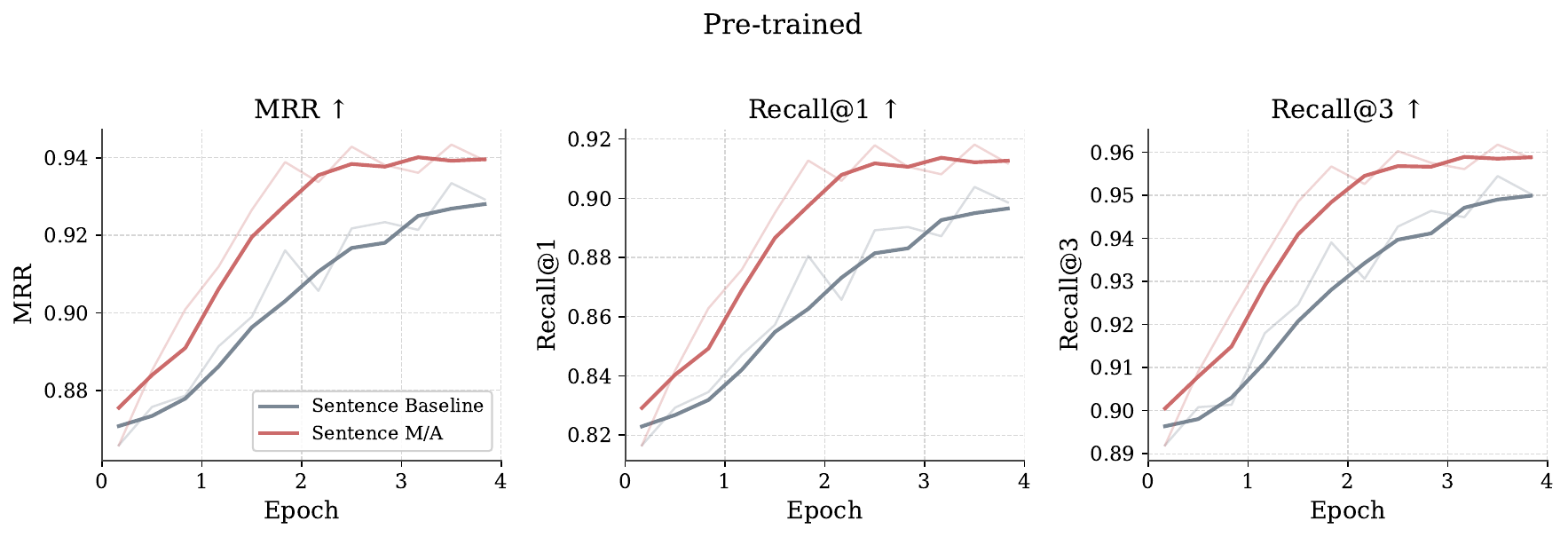}
    \caption{\textbf{Automatically generated M/A masks improve retriever training}. It achieves higher with fewer steps. Compared are sentence-level M/A augmentations with baseline training.} % [EXPLAINED IN LEGEND, can put it back in camera ready] Solid lines denote sentence-level; dashed lines token-level.}
    \label{fig:retriever_metrics}
 \vspace{-5pt}
% \end{figure}
\end{wrapfigure}

Fig.~\ref{fig:retriever_metrics} compares MRR and Recall@$k$ for (i) a baseline retriever trained on ground-truth positives and standard negatives, and (ii) our retriever trained with Merlin positives and Morgana-generated hard negatives replacing docs in the negative pool. 
% which we train on the ground-truth and the Merlin document and a pool of negative documents in which some documents are replaced by Morgana-generated confounders.
Despite using the same original documents, the M/A retriever improves faster and distinguishes across all metrics.
% Across all metrics, the M/A retriever shows faster and significant improvement during training than the baseline retriever, despite both being trained on the same original documents.
% This performance gap suggests that the inclusion of Merlin and Morgana documents provides a more informative training signal. The M/A approach enables the model to learn more robust representations that better distinguish relevant documents from challenging negatives. The Morgana confounders, in particular, appear to help the retriever develop stronger discrimination capabilities by presenting harder negative examples that share surface-level similarities with the ground-truth but lack critical information. This is further supported by the higher cosine similarities of the Morgana documents compared to other negative documents in Fig~\ref{fig:retriever_all_context_similarities}. Additionally, the Merlin documents provide another positive example, offering the retriever a richer signal for learning what constitutes a relevant document and potentially improving its robustness to variations in document quality.
This indicates that M/A contexts provide a richer supervision signal. Morgana confounders act as challenging hard negatives that are highly similar to the ground truth yet lack crucial evidence, sharpening discrimination (supported by higher cosine similarities of Morgana documents compared to other negatives, cf.~Fig.~\ref{fig:retriever_all_context_similarities}). Merlin contexts add complementary positives for learning what constitutes a relevant document.

\section{Conclusions}

%\lp{stress utility preservation, cond EIF and weights for false pos and negs}

By extending the Merlin-Arthur framework, we equip RAG systems with evidence awareness via game-theoretic supervision. While academic benchmarks often provide explicit ``unanswerable'' labels, industrial data rarely includes such samples. We bridge this gap by transforming explanations into active training signals using only curated positive target data: Merlin-Morgana interactions act as proofs that guide model behavior and improve groundedness, completeness,  soundness and allow for controllable weighting of false-positives and -negatives. Experiments across four models and five datasets confirm consistent gains in hallucination reduction and evidence reliance.

% Via our extension of the Merlin-Arthur Classifier framework, we equip RAG systems with verifiable evidence awareness. This game-theoretic supervision produces three desirable properties: higher groundedness, stronger completeness under helpful retrieval, and improved soundness under adversarial retrieval. Crucially, the model learns to align its behavior without explicit annotation of unanswerable queries or rejection labels--only from curated positive target data.

% Beyond improving reliability, our approach shows that explanations can serve as training signals rather than merely diagnostic tools. The Merlin--Morgana interactions function as structured, model-generated proofs that shape both retrieval and generation behavior. Our experiments demonstrate consistent gains across Llama-3.2-1B, Llama-3.2-3B, and Qwen3-4B on SQuAD, HotpotQA, and TriviaQA. We observe fewer hallucinations on the generator, more interpretable attribution maps, and a retriever that becomes more robust by learning from automatically generated confounders.

Finally, our theoretic adoption of Mutual Information for a realistic empirical setting -- the Conditional Explained Information Fraction -- enables rigorous measurement of context dependence in LLM-generated answers. This provides a foundation towards certified model behavior, a critical requirement as LLM-based systems become increasingly deployed in high-stakes applications, for which dataset accuracy alone is insufficient.

% \section*{Acknowledgements} % exclude for ICML review
% We thank Stephan Wäldchen for his valuable conceptual contributions, members of Aleph Alpha Research who assisted with helpful discussions -- especially Carina Kauf, Sagi Shaier, B\'eni Egressy and Felix Berkenkamp --, and members of the infrastructure support team and TU Darmstadt -- especially Patrick Schramowski.

\bibliographystyle{plainnat}
\bibliography{bib}

%%%%%%%%%%%%%%%%%%%%%%%%%%%%%%%%%%%%%%%%%%%%%%%%%%%%%%%%%%%%
\newpage
\appendix

\section*{Appendix}
\addcontentsline{toc}{subsection}{Appendix}

\startcontents[sections]
\subsection*{Contents of Appendix}
\hrule
\vspace{1em}

% This prints the TOC only for the sections following this point
\printcontents[sections]{l}{1}{\setcounter{tocdepth}{2}}

\vspace{1em}
\hrule
\vspace{2em}

\section{Impact Statement and Limitations} \label{app:impact-limitations} % does not count towards the page limit; we need this as per ICML policy https://icml.cc/Conferences/2026/CallForPapers
%This paper presents work with the goal to advance the field of machine learning by developing principled training and evaluation frameworks for reliable, evidence-aware model behavior. 

This research aims to increase the reliability of LLMs by addressing the problem of ``hallucinations.'' By implementing a proof-based system, we constrain the model to generate answers backed by verifiable evidence, which is an essential step for the safe deployment of AI in critical fields such as medicine, law, and engineering. Beyond safety, this approach enables models to verify their own outputs, potentially reducing the industry's reliance on expensive, human-labeled datasets. While we focus on Retrieval-Augmented Generation (RAG), our methodology -- including interactive-proof-style supervision and mutual-information lower bounds -- is broadly applicable to any learning system that must reason over external inputs and justify its decisions.

Despite these benefits, several limitations must be considered. The system is fundamentally limited by the quality of its sources; if the input documents contain misinformation, the model will treat them as ``proof'' and propagate those errors. There is also a risk that the model will become too conservative, refusing to answer helpful queries because it cannot find a direct citation. Additionally, the proof-verification process (and the benefits that come with it) increases computational costs and energy consumption compared to standard training, discussed extensively in \ref{sec:time_estimates}. While our framework does not directly address broader issues like data privacy or training bias, we believe it significantly contributes to the ongoing effort to make AI systems more transparent, predictable, and aligned with factual evidence.

\section{Properties of the Data Distribution}
\label{sec:data_distribution}

The full equation for Theorem 2.11 of~\citet{waldchen2024interpretability} is defined as:
\begin{equation}
\label{eq:upper_bound_full}
    Pr_{\mathcal{D}}(M) \ge 1 - \epsilon_c - \frac{\kappa \alpha^{-1} \epsilon_s}{1 - \epsilon_c + \kappa \alpha^{-1} B^{-1} \epsilon_s}\ .
\end{equation}

$H(\hat{y}(x))$ represents the entropy of the class distribution and can be calculated as $H(\hat{y}(x)) = -p \log(p) - (1-p) \log(1-p)$ with $p = \mathbb{P}_{x \sim \mathcal{D}}[\hat{y}(x) = 1]$. 
Similarly, the class imbalance can be computed with
$B = \max_{l \in \{-1, 1\}} \frac{\mathbb{P}_{x \sim \mathcal{D}}[\hat{y}(x)=l]}{\mathbb{P}_{x \sim \mathcal{D}}[\hat{y}(x)=-l]}$. 
In large scale open-ended evaluation datasets we can w.l.o.g. assume $B \approx 1$, as classes here mean concrete completion sequences, that are unlikely to be heavily imbalanced, and otherwise could be sub-sampled accordingly. In particular in web-crawled benchmarks exact sequences are statistically unlikely to occur.

The Asymmetric Feature Correlation $\kappa$ captures the degree to which features are concentrated in one class versus distributed in the other and is defined as:
% $$\kappa := \max_{l \in \{-1,1\}} \max_{F \subset D_p} \mathbb{E}_{y \sim \mathcal{D}_{l}|_{F^*}} \left[ \max_{z \subseteq y, z \in F} \frac{\mathbb{P}_{x \sim \mathcal{D}_{-l}}[z \subseteq x \mid x \in F^*]}{\mathbb{P}_{x \sim \mathcal{D}_{l}}[z \subseteq x \mid x \in F^*]} \right]\ ,$$

$$
\kappa :=
\displaystyle
\max_{l \in \{-1,1\}}
\max_{F}
\mathbb{E}_{y \sim \mathcal{D}_{l}\!\mid_{F^*}}
\left[
\max_{z \subseteq y,\; z \in F}
\frac{
\mathbb{P}_{x \sim \mathcal{D}_{-l}}\!\left[z \subseteq x \mid x \in F^*\right]
}{
\mathbb{P}_{x \sim \mathcal{D}_{l}}\!\left[z \subseteq x \mid x \in F^*\right]
}
\right],$$

for $F^*$ the set of all data points containing a feature from $F$.
For practicality Assumption 2.12 of~\citet{waldchen2024interpretability} uses $\kappa \approx 1$, meaning an even feature distribution cross classes. As we are in the RAG setting, i.e. information extraction as a Q/A faithfully from a given context, in an open-ended generation fashion, use benchmarks with thousands of random web-crawled samples, and utilize XAI methods to produce respective relevance masks on the input context, the existence of a heavily biased feature distributions is arguable negligible. It would only exist in malformed data distributions that do not contain meaningful context spans leading to answer of the corresponding questions, and may therefrom be filtered in a pre-processing step, or by the models conditioned accuracy filter. Latest when applying the framework, our adversarial training actively suppresses remaining, if existing, correlations by consistently penalizing reliance on parametric priors over retrieved evidence -- to each sample of the benchmark, the model has to find a supporting and adversarial mask.

%, for otherwise Morgana will eventually find the particular mask pattern to break Arthur during training. In particular in the discrete language domain with varying length samples being randomly sampled from web-crawled corpora, it is unlikely that the input tokens allow such a pattern without being exploited by Morgana.

% \begin{align}
%     F &\subset D_p \quad \text{(A specific subset of the partial dataset/feature space)} \\
%     F^* &:= \{x \in D \mid \exists z \in F : z \subseteq x\} \quad \text{(The set of all data points containing a feature from } F \text{)}
% \end{align}

Finally, the relative success rate $\alpha$ measures the computational capability of the adversary relative to the cooperator in finding convincing features is defined as: 
$$\alpha := \min_{l \in \{-1, 1\}} \frac{\mathbb{P}_{x \sim \mathcal{D}_{-l}}[A(\text{Mo}(x))=l \mid x \in F_{l}^*]}{\mathbb{P}_{x \sim \mathcal{D}_{l}}[A(\text{Me}(x))=l \mid x \in F_{l}^*]}\ ,$$
where $F_l := \text{Me}(D_l) \cap \{z \mid A(z) = l\} $ is the set of Features Merlin uses to successfully convince Arthur of class and $F_l^*$ the set of data points where Merlin finds successfully such a feature. We assume that Morgana is at least as powerful as Merlin, and as such approximate $\alpha \approx 1.$ Besides the large space of possible masks, this is in particular reasonable in the conditional metrics, for which we filter for samples that the model initially answered correct, and in combination with model training behavior, i.e. as the model is forced to reject on Morgana's feature samples.  

% \begin{align}
%     F_l &:= M(D_l) \cap \{z \in D_p \mid A(z) = l\} \quad \text{(Features Merlin uses to successfully convince Arthur of class } l \text{)} \\
%     F_l^* &:= \{x \in D \mid \exists z \subseteq x : z \in F_l\} \quad \text{(Data points containing a successful Merlin feature)}
% \end{align}

In summary, low values of $\epsilon_c$ and $\epsilon_s$ guarantee with Eqs.~\ref{eq:MI},~\ref{eq:upper_bound_full} that the features exchanged by the agents possess high mutual information with the ground truth class.
With this discussion, in Eq.~\ref{eq:MI} it particularly follows that $I_{baseline} := I(Y; \hat{Y}) \approx 1 - H_b(C)$ holds. 

\section{Optimality of Fixed-$x$ AtMan and Parity-Constrained Dynamic-$x$}
\label{app:fixed-x-rationale}
\label{app:bound_optimality}
This appendix expands the operating-point choice $x{=}0.6$ used throughout the main paper. The Merlin-Arthur masks of Eq.~\ref{eq:me-mo-goals} are defined as exhaustive subset-search optima -- enumerating all $\binom{N}{k}$ context subsets yields the tightest EIF the protocol admits. Any polynomial-time XAI heuristic, including our \textsc{AtMan} instantiation, yields a lower bound on this exhaustive ceiling; the last paragraph estimates the gap empirically via $\mathrm{EIF}^{\text{AtMan}}_{x=0.6}$ vs.\ $\mathrm{EIF}^{\text{exh}}_{x=0.6}$ on tractable context lengths. A natural relaxation of fixed-$x$ \textsc{AtMan} is a \emph{per-sample dynamic budget},
\begin{equation*}
    x_{\text{Me}}^\star \in \arg\max_{x} P_A(a^{\text{true}} \mid c_x^{\text{Me}}),
    \qquad
    x_{\text{Mo}}^\star \in \arg\min_{x} P_A(a^{\text{true}} \mid c_x^{\text{Mo}}),
\end{equation*}
optionally subject to a parity constraint $|x_{\text{Me}}\!-\!x_{\text{Mo}}|\!\le\!\delta$. The \citet{waldchen2024interpretability} bound is agnostic to the mask-selection mechanism, so for any heuristic scorer $h$ and any fixed budget $x$ the following monotonicity chain holds by strict relaxation:
\begin{equation}\label{eq:eif-monotonicity}
    \mathrm{EIF}^{h}_{x}
    \;\le\;
    \mathrm{EIF}^{h}_{\text{dyn}}
    \;\le\;
    \mathrm{EIF}^{\text{exh}}_{\text{dyn}},
    \qquad
    \mathrm{EIF}^{h}_{x}
    \;\le\;
    \mathrm{EIF}^{\text{exh}}_{x}
    \;\le\;
    \mathrm{EIF}^{\text{exh}}_{\text{dyn}}.
\end{equation}

\paragraph{Why Eq.~\ref{eq:eif-monotonicity} does not imply ``always use dynamic-$x$''.}
Read in isolation, the chain suggests the relaxation is free: dynamic-$x$ never decreases the implied bound. Two caveats prevent that conclusion.

\emph{(i) The bound's hidden parameters are not all mask-family invariant.} Eq.~\ref{eq:upper_bound_full} depends on the asymmetric feature correlation $\kappa$ and the relative prover strength $\alpha$ (App.~\ref{sec:data_distribution}). $\kappa$ is a property of the data distribution and is unchanged under any change of mask family. $\alpha$ is not: if Merlin and Morgana can choose budgets independently, Morgana's strict-reject optimum collapses onto the unmasked context $x_{\text{Mo}}{=}1$ on every sample, driving $\alpha\to\infty$. In that regime, Eq.~\ref{eq:MI} attributes mutual information to mask \emph{volume} rather than feature \emph{content}, voiding the certificate. Restricting the relaxation to a \emph{parity-constrained} dynamic-$x$ with $|x_{\text{Me}}\!-\!x_{\text{Mo}}|\!\le\!\delta$ preserves $\alpha$ up to $O(\delta)$. The choice $\delta{=}0$ recovers fixed-$x$ exactly.

\emph{(ii) The empirical headroom over fixed-$x{=}0.6$ appears small.} The gain of dynamic over fixed scales with the across-sample variance $\mathrm{Var}(x^\star)$ of the per-sample optimum, while compute grows by a factor of $K$ (the number of candidate budgets searched). On single-doc QA the \textsc{AtMan}-Merlin accuracy curve peaks at $x\!\approx\!0.6$ across models, datasets, training regimes, and durations (Fig.~\ref{fig:atman_filter_peak_sec5}), so the population-level optimum is stable; per-sample optima cluster around this peak (cf.\ Fig.~\ref{fig:evaluate_maskers_squad_sentence_before_after_training}), making the expected lift modest at substantial extra cost.

\paragraph{Choice of $x$ for Merlin, Morgana, and parity-constrained dynamic-$x$.}
The two provers have asymmetric optima with respect to $x$:
\begin{itemize}[topsep=2pt, itemsep=2pt, parsep=0pt]
    \item \textbf{Merlin} ($x_{\text{Me}}^\star$) maximizes $P_A(a^{\text{true}}\!\mid\!c_x^{\text{Me}})$. On HotpotQA (Fig.~\ref{fig:evaluate_maskers_squad_sentence_before_after_training}) the curve is approximately unimodal in $x$, with a population peak near $x\!\approx\!0.6$ and a plateau over $x\!\in\![0.4,0.7]$. The same shape replicates across SQuAD/TriviaQA and across Llama-3.2-1B/3B and Qwen3-4B (Fig.~\ref{fig:atman_filter_peak_sec5}).
    \item \textbf{Morgana} ($x_{\text{Mo}}^\star$) minimizes the same probability. Morgana's curve drops earliest -- masking the answer-bearing sentence is sufficient to disrupt Arthur, so the strongest-attack region typically lies at $x_{\text{Mo}}^\star\!\in\![0.4,0.7]$ as well, slightly biased toward larger $x$ than Merlin's peak. 
    % The right tail $x\!\to\!1$ is excluded as degenerate: with no context, full-SQuAD-trained models reject (Fig.~\ref{fig:atman_filter_peak_sec5}b,c), and answerable-only baselines fall back to parametric memory (Fig.~\ref{fig:atman_filter_peak_sec5}e), so $x_{\text{Mo}}{=}1$ measures memorization rather than feature removal.
\end{itemize}
The horizontal gap between Merlin's peak and Morgana's trough in Fig.~\ref{fig:evaluate_maskers_squad_sentence_before_after_training} sits within $\approx\!0.2$ across both pre- and post-training curves, suggesting $\delta\!\lesssim\!0.2$ as a defensible parity budget. Smaller $\delta$ tightens the $O(\delta)$ slack in $\alpha$ at the cost of empirical headroom; larger $\delta$ moves Morgana into the right-tail region where $x_{\text{Mo}}\!\to\!1$ collapses the certificate (cf.\ paragraph above). Concretely:
\begin{itemize}[topsep=2pt, itemsep=2pt, parsep=0pt]
    \item $\delta{=}0$: recovers fixed-$x$ \textsc{AtMan}; tightest $\alpha$, no per-sample adaptation. Our operating point.
    \item $\delta\!\in\![0.05,0.10]$: small $\alpha$ slack; captures the per-sample variation around $x\!\approx\!0.6$ visible in Fig.~\ref{fig:evaluate_maskers_squad_sentence_before_after_training}.
    \item $\delta\!\in\![0.10,0.20]$: covers the full empirical Merlin/Morgana optimum spread; recommended upper bound for parity-constrained dynamic-$x$.
    \item $\delta\!>\!0.20$: lets Morgana approach the unmasked-context regime; not defensible as a certificate.
\end{itemize}

\paragraph{Adopted operating point.} We adopt fixed-$x{=}0.6$ ($\delta{=}0$) \textsc{AtMan} for both provers in the main experiments, corresponding to Merlin's population peak and to Morgana's strong-attack region. By Eq.~\ref{eq:eif-monotonicity}, both the $\delta$-parity dynamic-$x$ AtMan EIF and the exhaustive-search ceiling lie at or above the values we report; our numbers are therefore conservative under both relaxations.

\paragraph{On optimality of $\mathrm{EIF}^{\textsc{AtMan}}_{x=0.6}$.} To validate our mask fidelity, we compared our masks against an \textit{exhaustive search} on 100 random samples. We observe near-total overlap, averaged of all Merlin and Morgana masks, at a 20\% masking rate (99\% for SQuAD; 97\% for HotpotQA), which remains robust at 40\% (90\%; 93\%) and 60\% (73\%; 74\%). Notably, the lower overlap at 60\% does not imply a loss of precision regarding core evidence; rather, it reflects the increased entropy of selecting from the remaining irrelevant context once the primary evidence is already identified: we observe that Merlin optimization leaves the ground-truth evidence intact until the highest compression levels (supported by the high groundedness scores for Merlin in Fig.~\ref{fig:evaluate_maskers_squad_sentence_before_after_training}), while Morgana masks it early on. As such, the EIF scores remain comparable.
This makes the \textsc{AtMan} instantiation a high-fidelity proxy for the optimal prover. By yielding a conservative approximation -- effectively a lower bound of the theoretical lower bound (Eq.~\ref{eq:eif-monotonicity-main}) -- the framework provides trustworthy numbers on EIF, soundness and completeness of practical relevance.

\section{Masking Procedure}\label{app:evaluate_maskers}

We describe the masking procedure for Merlin and Morgana in Algorithm~\ref{alg:masking}.
\begin{algorithm}
%\footnotesize
\caption{Context Masking  for Merlin and Morgana. Procedure \textsc{MaskContext}}
\label{alg:masking}
\begin{algorithmic}[1]
    \Require $q, c, a^{\text{true}}, x\%$
    %\Ensure $c_{\text{Me}}, c_{\text{Mo}}$
    
    \State $N \gets \text{length}(c)$
    \State $k \gets \lfloor N \cdot x\% \rfloor$
    \State $p^{\text{Me}} \gets [0, \dots, 0]$ \Comment{Merlin scores; token or sentence level}
    \State $p^{\text{Mo}} \gets [0, \dots, 0]$ \Comment{Morgana scores}
    
    \State \textbf{Phase 1:} Estimate importance of each token
    \For{$i = 0$ \textbf{to} $N - 1$}
        \State $c_{\text{masked}} \gets \text{copy}(c)$
        \State $c_{\text{masked}} \gets \text{mask}(c_{\text{masked}}, i)$ \Comment{Masks the $i$-th token}
        \State $p_{\text{true}} \gets P(A(q, c_{\text{masked}}) = a^{\text{true}})$
        %\State $p_{\text{reject}} \gets P(A(q, c_{\text{masked}}) = a^{\text{reject}})$
        \State $p^{\text{Me}}[i] \gets p_{\text{true}}$
        \State $p^{\text{Mo}}[i] \gets 1 - p_{\text{true}}$ \Comment{Probability of incorrect answer}
    \EndFor
    
    \State \textbf{Phase 2:} Select and mask top-$k$ influential tokens
    \State $v_{\text{Me}} \gets \text{topk}(p^{\text{Me}}, k)$
    \State $v_{\text{Mo}} \gets \text{topk}(p^{\text{Mo}}, k)$
    
    \State $c_{\text{Me}} \gets \text{mask}(c, v_{\text{Me}})$ \Comment{Merlin-masked context}
    \State $c_{\text{Mo}} \gets \text{mask}(c, v_{\text{Mo}})$ \Comment{Morgana-masked context}
    
    \State \Return $c_{\text{Me}}, c_{\text{Mo}}$
\end{algorithmic}
\end{algorithm}

\textbf{\textit{We list the following results:}}

\begin{figure*}[t]
    \centering
    \begin{subfigure}[t]{\linewidth}
        \centering
        \includegraphics[width=\linewidth]{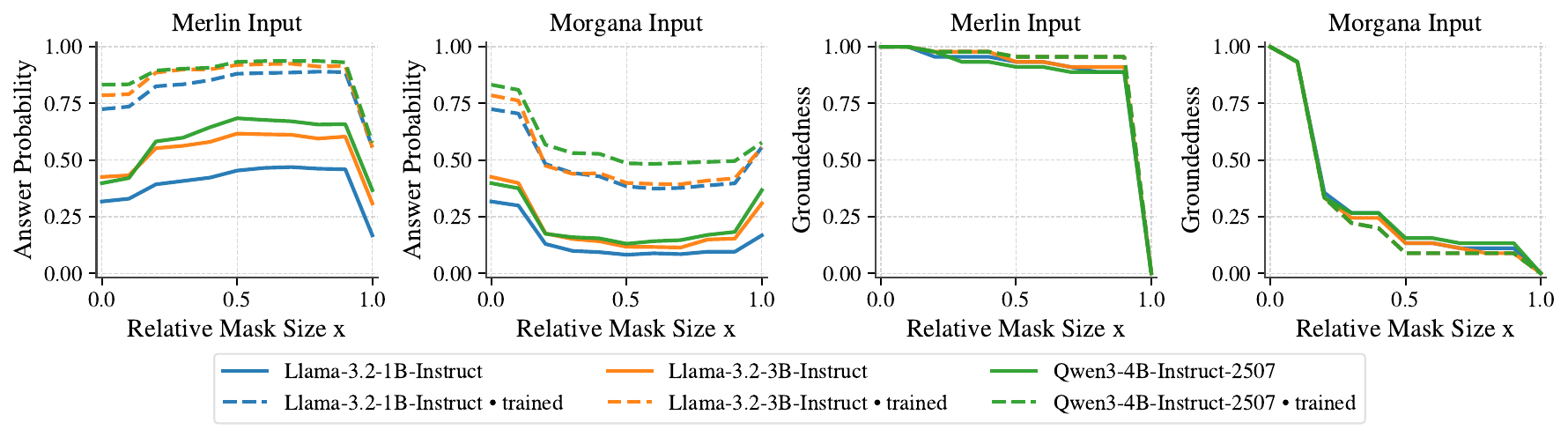}
        \caption{\textbf{Sentence}-level masking on \textbf{SQuAD}.}
        \label{fig:evaluate_maskers_squad_token_before_after_training}
    \end{subfigure}
    
    \vspace{0.5em}

    \begin{subfigure}[t]{\linewidth}
        \centering
        \includegraphics[width=\linewidth]{figures/eval_maskers/HotpotQA_open_sentence_before_after_training.pdf}
        \caption{\textbf{Sentence}-level masking on \textbf{HotpotQA}.}
        \label{fig:evaluate_maskers_hotpotqa_token_before_after_training}
    \end{subfigure}
    
    %\vspace{0.5em}
    
    %\begin{subfigure}[t]{\linewidth}
    %    \centering
    %    \includegraphics[width=\linewidth]{figures/eval_maskers/TriviaQA_open_sentence_before_after_training.pdf}
    %    \caption{\textbf{Sentence}-level masking on \textbf{TriviaQA}.}
    %    \label{fig:evaluate_maskers_triviaqa_token_before_after_training}
    %\end{subfigure}
    
    \caption{\textbf{Me increases accuracy and groundedness, Mo pushes them down.} Comparison of Atman masking behavior \textbf{before and after M/A training} on two datasets. 
    Left: The probability Arthur’s assigns to $a^{\text{true}}$ under Merlin- and Morgana-masked contexts as a function of mask size $x$. 
    Right: Groundedness of Merlin and Morgana contexts. 
    Solid lines show pretrained models; dashed lines show models after M/A training.}
    \label{fig:eval_masker_before_after}
\end{figure*}

In Fig.~\ref{fig:eval_masker_before_after} we show our masking evaluation across two datasets with both \textbf{token}- and \textbf{sentence}-level masking.

\begin{figure*}[t]
    \centering
    \begin{subfigure}[t]{\linewidth}
        \centering
        \includegraphics[width=\linewidth]{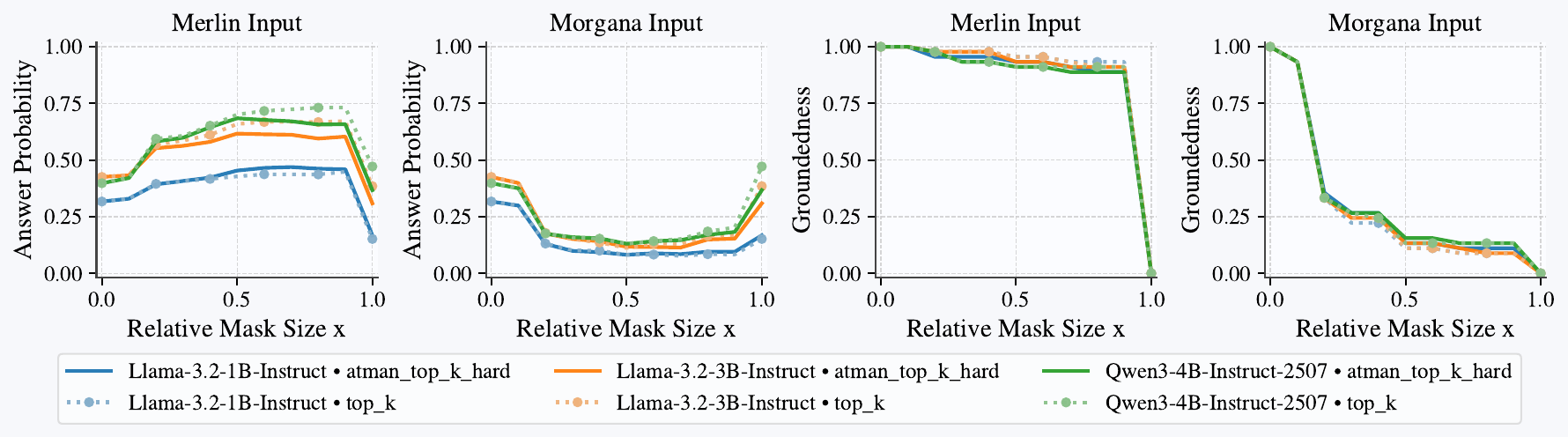}
        \caption{\textbf{Sentence}-level masking on \textbf{SQuAD}.}
        \label{fig:evaluate_maskers_squad_sentence_algo_comparison}
    \end{subfigure}
    
    %\vspace{0.5em}
    
    %\begin{subfigure}[t]{\linewidth}
    %    \centering
    %    \includegraphics[width=\linewidth]{figures/eval_maskers/SQuAD_open_token_atman-topk_vs_topk.pdf}
    %    \caption{\textbf{Token}-level masking on \textbf{SQuAD}.}
    %    \label{fig:evaluate_maskers_squad_token_algo_comparison}
    %\end{subfigure}

    \vspace{0.5em}

    \begin{subfigure}[t]{\linewidth}
        \centering
        \includegraphics[width=\linewidth]{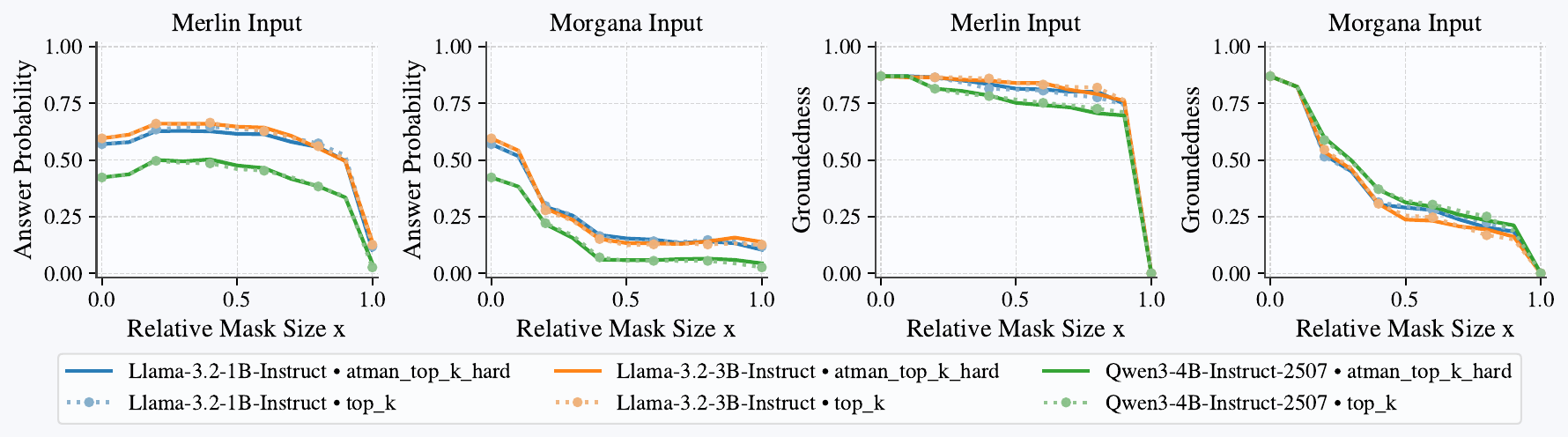}
        \caption{\textbf{Sentence}-level masking on \textbf{HotpotQA}.}
        \label{fig:evaluate_maskers_hotpot_sentence_algo_comparison}
    \end{subfigure}
    
    %\vspace{0.5em}
    
    %\begin{subfigure}[t]{\linewidth}
    %    \centering
    %    \includegraphics[width=\linewidth]{figures/eval_maskers/TriviaQA_open_sentence_atman-topk_vs_topk.pdf}
    %    \caption{\textbf{Sentence}-level masking on \textbf{TriviaQA}.}
    %    \label{fig:evaluate_maskers_triviaqa_token_algo_comparison}
    %\end{subfigure}
    
    \caption{Comparison between \textbf{attention- and string-based masking strategies}  on two datasets. Left: probability Arthur assigns to $a^{\text{true}}$ under Merlin- and Morgana-masked contexts as a function of mask size $x$. 
    Right: groundedness of the masked contexts.
    Solid curves correspond to top-$k$ attention masking; dashed curves correspond to 
    top-$k$ token replacement (``...'').}
    \label{fig:eval_masker_masking_comparison}
\end{figure*}

In Fig.~\ref{fig:eval_masker_masking_comparison} we \textbf{compare attention-based masking to string-based} masking with both token- and sentence-level masking. Both attention-level masking and string-based masking strategies perform similarly.

\begin{figure*}[t]
    \centering
    \includegraphics[width=\linewidth]{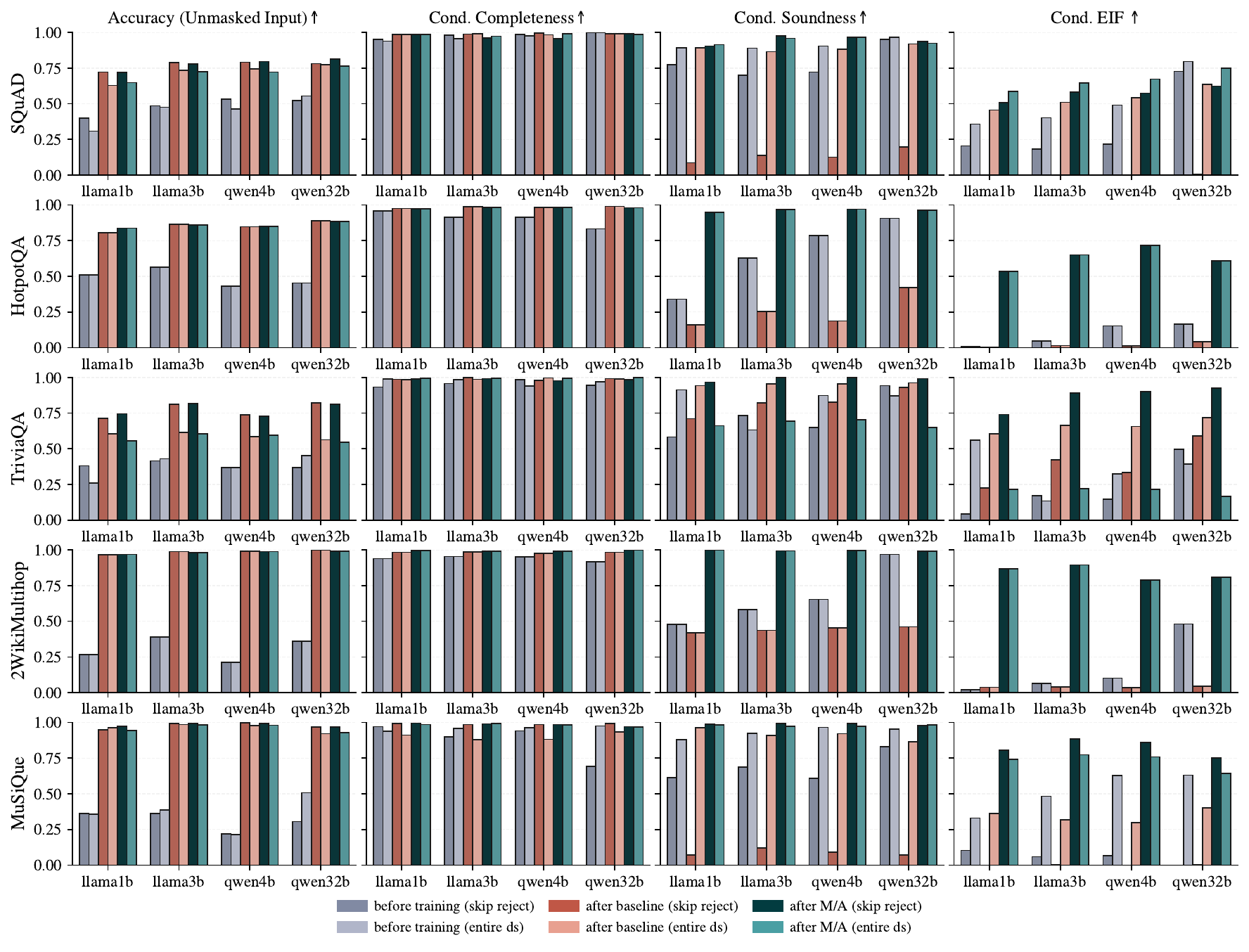}
    \caption{\textbf{With M/A training, completeness, soundness and EIF increase.} We show results for all three models on SQuAD, HotpotQA, and TriviaQA at three stages: (1) before training, (2) after M/A training, and (3) after baseline finetuning. For HotpotQA, which lacks annotated unanswerable samples, the ``skip reject'' evaluation is equivalent to the entire dataset ``entire ds''.}
    \label{fig:RAG_training_barplot_full}
\end{figure*}

\begin{figure*}[ht]
    \centering
    \includegraphics[width=\linewidth]{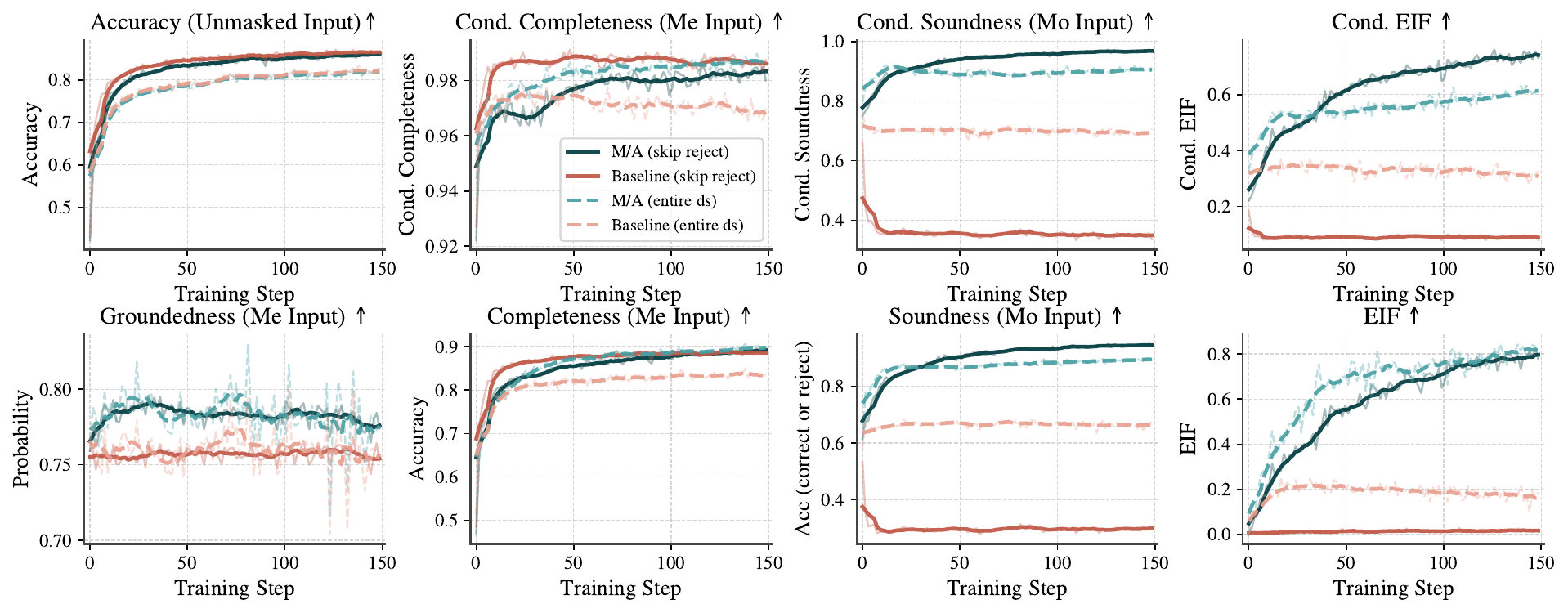}
    \caption{\textbf{M/A leads to measurable lower-bounds on datasets, especially without reject annotations.} 
    Training metrics averaged over all models and datasets. Step~0 is \emph{before} any model updates, reflecting initial performance. Results with all models, metrics and datasets
    % in Fig.~\ref{fig:RAG_train_curves_hotpot_app}. SQuAD and TriviaQA results 
    are in Figs.~\ref{fig:RAG_train_curves_squad_app}-\ref{fig:RAG_train_curves_musique_app_qwen}.}
    \label{fig:RAG_train_curves_app}
\vspace{-7pt}
\end{figure*}

\section{Prompting for Reject Behavior}\label{app:prompt}
All models in our experiments -- both baseline and M/A-trained -- use the same fixed instruction prompt. Since all evaluated models are instruction-tuned, this prompt explicitly specifies the expected reject behavior: the model is instructed to answer a question only when the answer is supported by the provided context and to output \texttt{Reject} otherwise. We keep the prompt identical across training and evaluation to ensure that differences in reject behavior are attributable solely to the training procedure, not to prompt design.

\paragraph{Instruction Prompt}
\begin{quote}\ttfamily
You are a helpful assistant and will answer the user's questions carefully, logically, accurately and well-reasoned.
Use the given context to answer the question faithfully. Answer only if the answer is present in the given context, otherwise answer "Reject" if the answer is not present in the context. 

Context:
\{CONTEXT\}  

Question:
\{QUESTION\}  

The final answer is:
\end{quote}

\clearpage

\section{Licenses of Models and Datasets} \label{app:licenses}
We use only publicly released, openly licensed models and datasets, all properly cited in the bibliography. We respect the terms of each license; in particular, our use is limited to non-commercial academic research and we credit each original author. Below we list, for each asset, the version used, a direct URL, the license name, and a link to the license text where applicable.

\paragraph{Generator Models}
\begin{itemize}[leftmargin=*]
    \item \textbf{Llama-3.2-1B-Instruct} \citep{grattafiori2024llama}. \\
    URL: \url{https://huggingface.co/meta-llama/Llama-3.2-1B-Instruct}. \\
    License: Llama 3.2 Community License Agreement (\url{https://github.com/meta-llama/llama-models/blob/main/models/llama3_2/LICENSE}). \\
    Acceptable Use Policy: \url{https://www.llama.com/llama3_2/use-policy/}.
    \item \textbf{Llama-3.2-3B-Instruct} \citep{grattafiori2024llama}. \\
    URL: \url{https://huggingface.co/meta-llama/Llama-3.2-3B-Instruct}. \\
    License: Llama 3.2 Community License Agreement (\url{https://github.com/meta-llama/llama-models/blob/main/models/llama3_2/LICENSE}). \\
    Acceptable Use Policy: \url{https://www.llama.com/llama3_2/use-policy/}.
    \item \textbf{Qwen3-4B-Instruct-2507} \citep{qwen3technicalreport}. \\
    URL: \url{https://huggingface.co/Qwen/Qwen3-4B-Instruct-2507}. \\
    License: Apache License 2.0 (\url{https://www.apache.org/licenses/LICENSE-2.0}).
    \item \textbf{Qwen2.5-32B-Instruct} \citep{yang2024qwen2technicalreport}. \\
    URL: \url{https://huggingface.co/Qwen/Qwen2.5-32B-Instruct}. \\
    License: Apache License 2.0 (\url{https://huggingface.co/Qwen/Qwen2.5-32B-Instruct/blob/main/LICENSE}).
\end{itemize}

\paragraph{Retriever Model}
\begin{itemize}[leftmargin=*]
    \item \textbf{granite-embedding-small-english-r2} \citep{awasthy2025graniteembeddingr2models}. \\
    URL: \url{https://huggingface.co/ibm-granite/granite-embedding-small-english-r2}. \\
    License: Apache License 2.0 (\url{https://www.apache.org/licenses/LICENSE-2.0}).
\end{itemize}

\paragraph{Datasets}
\begin{itemize}[leftmargin=*]
    \item \textbf{SQuAD~2.0} \citep{rajpurkar-etal-2016-squad,rajpurkar-etal-2018-know}. \\
    URL: \url{https://rajpurkar.github.io/SQuAD-explorer/}. \\
    License: CC BY-SA 4.0 (\url{https://creativecommons.org/licenses/by-sa/4.0/legalcode}).
    \item \textbf{HotpotQA} \citep{yang2018hotpotqa}. \\
    URL: \url{https://hotpotqa.github.io/}. \\
    License: CC BY-SA 4.0 (\url{https://creativecommons.org/licenses/by-sa/4.0/legalcode}).
    \item \textbf{TriviaQA} \citep{2017arXivtriviaqa}. \\
    URL: \url{https://nlp.cs.washington.edu/triviaqa/}; repository: \url{https://github.com/mandarjoshi90/triviaqa}. \\
    License: Apache License 2.0 (\url{https://github.com/mandarjoshi90/triviaqa/blob/master/LICENSE}).
    \item \textbf{2WikiMultihopQA} \citep{ho-etal-2020-constructing}. \\
    URL: \url{https://github.com/Alab-NII/2wikimultihop}. \\
    License: Apache License 2.0 (\url{https://github.com/Alab-NII/2wikimultihop/blob/main/LICENSE}).
    \item \textbf{MuSiQue} \citep{trivedi-etal-2022-musique}. \\
    URL: \url{https://github.com/StonyBrookNLP/musique}. \\
    License: CC BY 4.0 (\url{https://creativecommons.org/licenses/by/4.0/legalcode}).
\end{itemize}

\section{Detailed Description of Datasets} \label{app:datasets}
\textbf{SQuAD2.0} \citep{rajpurkar-etal-2016-squad, rajpurkar-etal-2018-know} contains single-hop questions grounded in a single paragraph. The dataset features precise ground-truth answer-span annotations for all answerable question-context pairs. This dataset tests whether Arthur benefits from Merlin's supportive contexts and rejects confidently when Morgana removes essential evidence.

\textbf{HotpotQA} \citep{yang2018hotpotqa} has multi-hop questions requiring the integration of information across documents. It includes gold-standard labels for supporting-fact sentences, identifying the specific evidence required for justification. Notably, this benchmark consists entirely of answerable questions and provides no native training signals or annotations for unanswerable samples. Here we examine whether masking-based perturbations help Arthur focus on the evidence spans that jointly support the answer.

\textbf{TriviaQA} \citep{2017arXivtriviaqa} contains long-text open-domain questions paired with noisy, loosely relevant retrieval from web data. It is characterized by noisy retrieval and distant supervision, often resulting in distractor-heavy evidence. This setting stresses long-context understanding and Arthur's ability to handle incomplete, distractor-heavy evidence and abstain when the context does not justify an answer.

\textbf{2WikiMultihopQA} \citep{ho-etal-2020-constructing} is a multi-hop reasoning dataset that requires models to navigate structured reasoning chains over Wikipedia and Wikidata. It focuses on the integration of evidence across multiple documents to verify complex queries.

\textbf{MuSiQue} \citep{trivedi-etal-2022-musique} is a complex multi-hop reasoning benchmark. It is designed to test robust information integration while using distractor-heavy contexts to ensure models rely on grounded evidence.

\section{M/A RAG Training}\label{app:RAG_training}
\subsection{Generator Models}\label{app:generator-models}
In our experiments, we use the following LLMs as Arthur, the RAG answer generator:

Llama-3.2-1B-Instruct \citep{grattafiori2024llama}: \url{https://huggingface.co/meta-llama/Llama-3.2-1B},

Llama-3.2-3B-Instruct \citep{grattafiori2024llama}: \url{https://huggingface.co/meta-llama/Llama-3.2-3B},

Qwen3-4B-Instruct \citep{qwen3technicalreport}: \url{https://huggingface.co/Qwen/Qwen3-4B-Instruct-2507},

Qwen2.5-32B-Instruct \citep{yang2024qwen2technicalreport}: \url{https://huggingface.co/Qwen/Qwen2.5-32B-Instruct}.

\subsection{RAG Training Procedure with M/A}

\subsubsection{M/A Generator Training}
We describe RAG generator training in Algorithm~\ref{alg:rag_training}.

\begin{algorithm}[ht]
\caption{RAG Training}
\label{alg:rag_training}
\begin{algorithmic}[1]
    \Require $batch, x\%, \lambda_{\text{Me}}, \lambda_{\text{Mo}}$
    %\Ensure $\mathcal{L}_{\text{total}}$
    
    \State \textbf{Phase 1: Create masked context variants}
    \State $\mathit{augmented\_batch} \gets [\,]$
    \For{each $(q, c, a^{\text{true}})$ in $\textit{batch}$}
        \State $(c_{\text{Me}}, c_{\text{Mo}}) \gets \textsc{MaskContext}(q, c, a^{\text{true}}, x\%)$
        \State $\mathit{augmented\_batch}.\text{append}(q, c, c_{\text{Me}}, c_{\text{Mo}}, a^{\text{true}})$
    \EndFor

    \State \textbf{Phase 2: Compute loss for each context type}
    \State $\mathcal{L}_{\text{total}} \gets 0$
    \For{each $(q, c, c_{\text{Me}}, c_{\text{Mo}}, a^{\text{true}})$ in $\mathit{augmented\_batch}$}
        \State $\mathcal{L}_{\text{util}} \gets \mathrm{CE}(A(q, c), a^{\text{true}})$
        \State $\mathcal{L}_{\text{Me}} \gets \mathrm{CE}(A(q, c_{\text{Me}}), a^{\text{true}})$
        % \State $\mathcal{L}_{\text{Mo}} \gets \frac{1}{2} \left[ \mathrm{CE}(A(q, c_{\text{Mo}}), a^{\text{true}}) + \mathrm{CE}(A(q, c_{\text{Mo}}), a^{\text{reject}}) \right]$
        \State $\mathcal{L}_{\text{Mo}} \gets  \mathrm{CE}(A(q, c_{\text{Mo}}), a^{\text{reject}})$
        
        \State $\mathcal{L} \gets \lambda_{\text{util}} \mathcal{L}_{\text{util}} + \lambda_{\text{Me}} \mathcal{L}_{\text{Me}} + \lambda_{\text{Mo}} \mathcal{L}_{\text{Mo}}$
        \State $\mathcal{L}_{\text{total}} \gets \mathcal{L}_{\text{total}} + \mathcal{L}$
    \EndFor
    
    \State \Return $\mathcal{L}_{\text{total}}$
\end{algorithmic}
\end{algorithm}

%\subsubsection{Training Morgana with both Correct Answer and `Reject'}\label{app:why-not-just-reject}
%Under adversarial Morgana contexts, rejection is not always the correct behavior. Some adversarially masked contexts may still retain sufficient evidence to support the correct answer -- particularly later in training, when Arthur becomes more robust and Morgana has exhausted the viable strategies and its perturbations increasingly resemble uninformed guesses rather than targeted attacks.

%Consequently, during Arthur’s training we treat both the correct answer $a^{\text{true}}$ and the reject response $a^{\text{reject}}$ as valid outcomes under Morgana input. Optimizing solely for rejection would encourage the trivial strategy that rejects whenever evidence is perturbed, thus would increase Arthur's soundness and decrease its completeness. Allowing both outcomes instead encourages Arthur to reject only when evidence is genuinely insufficient, while still answering correctly when adversarial masking fails to eliminate decisive evidence.

\subsection{Metrics recorded during RAG Training} \label{app:rag-metrics}
We compute accuracy using exact match.  We report the following metrics for post-hoc analysis -- they are not used in training: We evaluate groundedness, completeness, soundness (Section~\ref{sec:ma-for-rag}), and the model's reject rate across all datasets.
To ensure high-fidelity reporting of groundedness, calculated as the number of unique target words present in the context, we leverage dataset-specific ground-truth annotations where available: we use answer-span annotations for SQuAD2.0 and supporting-fact sentence labels for HotpotQA. For TriviaQA, which lacks such labels, we approximate groundedness via computing the string-oprelap of the gold answer within the remaining context. 
%To compute groundedness for SQuAD2.0, we use span annotations indicating where the answer appears in the context; for HotpotQA we use its sentence-level supporting-fact annotations; and for TriviaQA -- which provides no ground-truth evidence span -- we approximate groundedness by string-matching the gold answer within the remaining context after masking.

\subsection{M/A RAG Training Results} \label{app:rag-training-results}

We provide the following results:

\begin{itemize}[leftmargin=*]
    \item \textbf{Bar chart} in Fig.~\ref{fig:RAG_training_barplot_full} is the more detailed view on Fig.~\ref{fig:RAG_training_barplot_avg}, presenting accuracy, completeness, soundness and EIF for all three models on SQuAD, HotpotQA, and TriviaQA at three stages: (1) before training, (2) after M/A training, and (3) after baseline finetuning. Discussion below in \ref{app:utility-preservation}.
    \item Aggregated training curves across all datasets and models are in Fig.\ref{fig:RAG_train_curves_app}.
    \item Full set of detailed \textbf{RAG training curves} for \textbf{SQuAD}, including accuracy, completeness, soundness,
    groundedness, reject rates, and the individual Merlin, Morgana, utility, and total loss
    components in Figs.~\ref{fig:RAG_train_curves_squad_app}, \ref{fig:RAG_train_curves_squad_app_qwen}. We compare M/A training to a vanilla finetuning baseline.
    \item Full set of \textbf{RAG training curves} for \textbf{HotpotQA} in Figs.~\ref{fig:RAG_train_curves_hotpot_app}, \ref{fig:RAG_train_curves_hotpot_app_qwen}.
    \item Full set of \textbf{RAG training curves} for \textbf{TriviaQA} in Figs.~\ref{fig:RAG_train_curves_trivia_app}, \ref{fig:RAG_train_curves_trivia_app_qwen}.
    \item Full set of \textbf{RAG training curves} for \textbf{2WikiMultihop} in Figs.~\ref{fig:RAG_train_curves_2wikimultihop_app}, \ref{fig:RAG_train_curves_2wikimultihop_app_qwen}.
    \item Full set of \textbf{RAG training curves} for \textbf{MuSiQue} in Figs.~\ref{fig:RAG_train_curves_musique_app}, \ref{fig:RAG_train_curves_musique_app_qwen}.

    \item Training curves \textbf{comparing}
    (a) \textbf{token}-level versus (b) \textbf{sentence}-level masking during M/A training in Figs.~\ref{fig:RAG_train_curves_squad_sentence_token_app}. See  \ref{app:hotpotqa-soundness} for discussion.

    \item Training curves \textbf{comparing}  (a) \textbf{attention}-based
    masking with (b) \textbf{string}-based masking in Fig.~\ref{fig:RAG_train_curves_squad_atman_vs_topk}. See \ref{app:attention-vs-string-level} for discussion.

\end{itemize}

Note: For the baseline, Merlin and Morgana losses are not optimized for, but just computed and plotted ($\lambda_{\text{util}} = 1$, $\lambda_{\text{Me}} = \lambda_{\text{Mo}} = 0$).

In all training plots, step~0 denotes the metrics computed on the first batch \emph{before} any model updates, reflecting the model's initial performance.

\subsubsection{Utility Preservation with Improved Soundness and Reject Behavior}\label{app:utility-preservation}
Fig.~\ref{fig:RAG_training_barplot_full} compares M/A training to both the pretrained instruct model (no finetuning) and standard supervised finetuning. Accuracy under M/A training consistently increases relative to the instruct model and remains comparable to vanilla finetuning, indicating that M/A supervision does not degrade core model utility. At the same time, M/A training yields substantial gains in completeness, soundness, and Explained Information Fraction (EIF), demonstrating improved robustness and increased evidence-faithful behavior.

\subsubsection{M/A Enables Reject Behaviour without Unanswerable Questions} \label{app:hotpotqa-soundness}
\begin{figure*}[t]
    \centering
    \begin{subfigure}[t]{\linewidth}
        \centering
        \includegraphics[width=\linewidth]{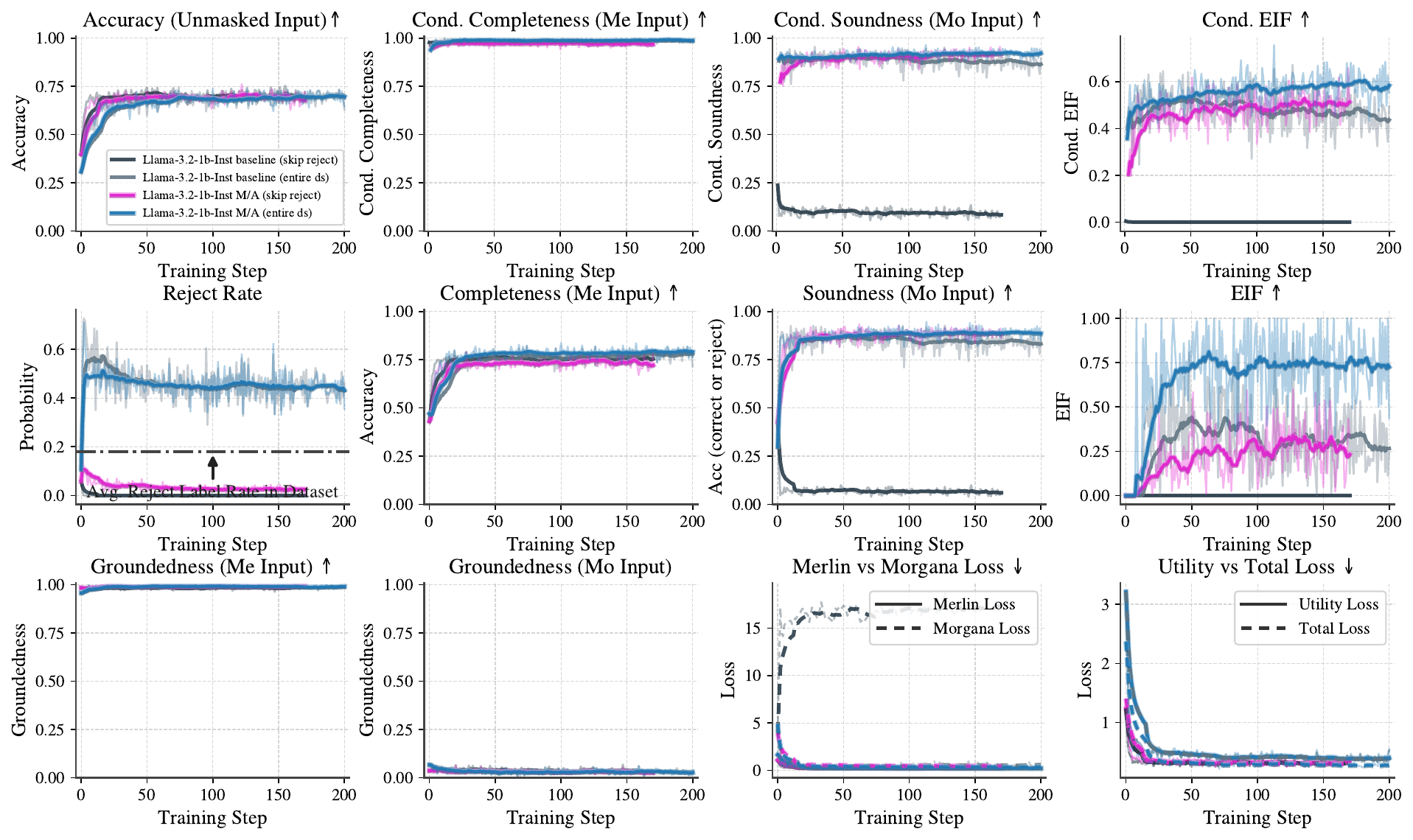}
        \caption{\textbf{Llama-3.2-1B-Instruct}}
    \end{subfigure}

    \begin{subfigure}[t]{\linewidth}
        \centering
        \includegraphics[width=\linewidth]{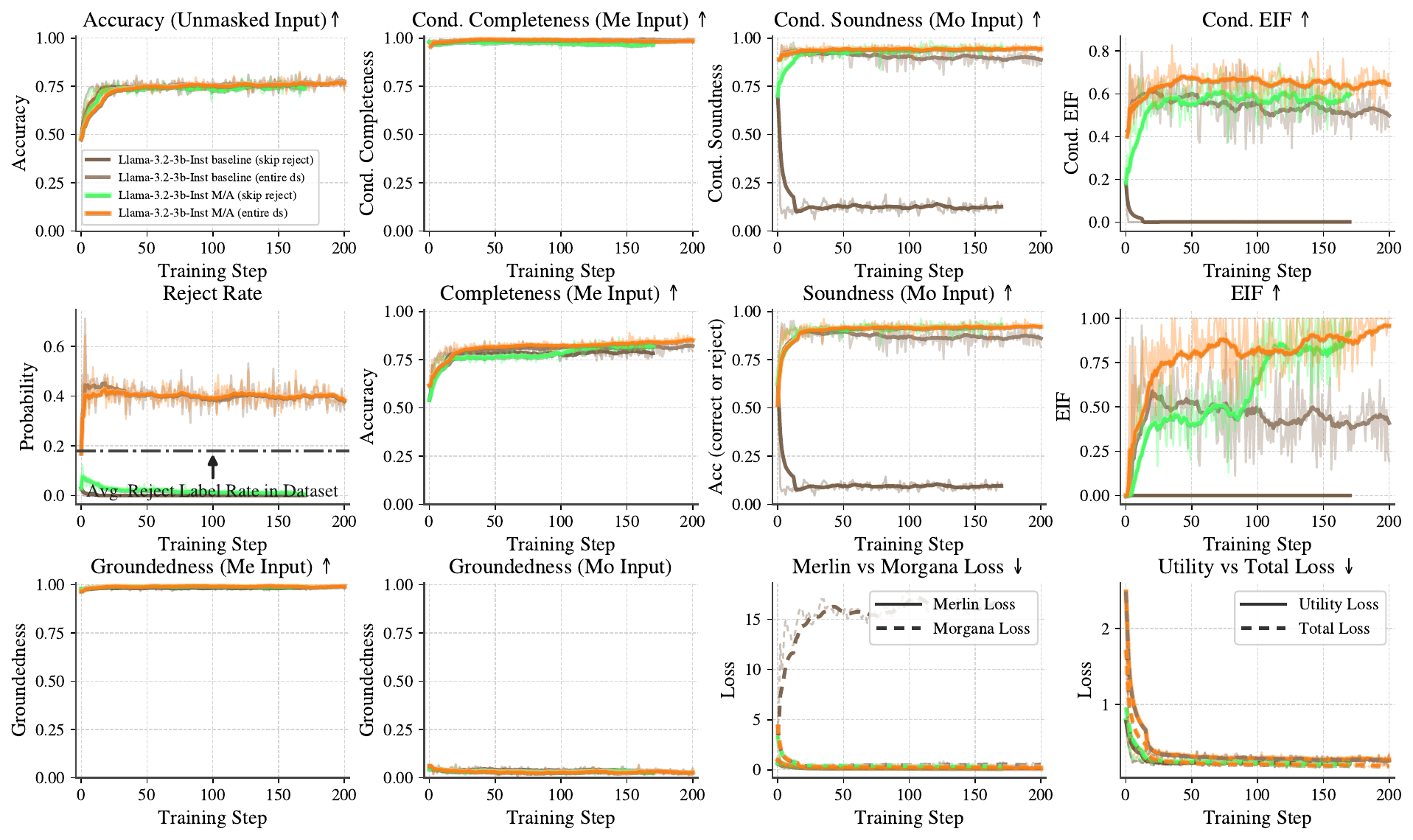}
        \caption{\textbf{Llama-3.2-3B-Instruct}}
    \end{subfigure}
   
    \caption{Training curves on \textbf{SQuAD} comparing \textbf{(i) sentence-level  \textit{M/A training} to (ii) \textit{baseline finetuning}} ($\lambda_{\text{util}} = 1$, $\lambda_{\text{Me}} = \lambda_{\text{Mo}} = 0$). Note that in the baseline setting, the Merlin and Morgana losses are not optimized -- they are only computed and plotted. Qwen models results in Fig.~\ref{fig:RAG_train_curves_squad_app_qwen}. Step~0 are metrics on the first batch \emph{before} any model updates, reflecting initial performance. Strong lines show a rolling-window average over 8\% of the total training datapoints. ``Shadow'' lines represent actual data.}
    \label{fig:RAG_train_curves_squad_app}
\end{figure*}

\begin{figure*}[t]
    \centering
    \begin{subfigure}[t]{\linewidth}
        \centering
        \includegraphics[width=\linewidth]{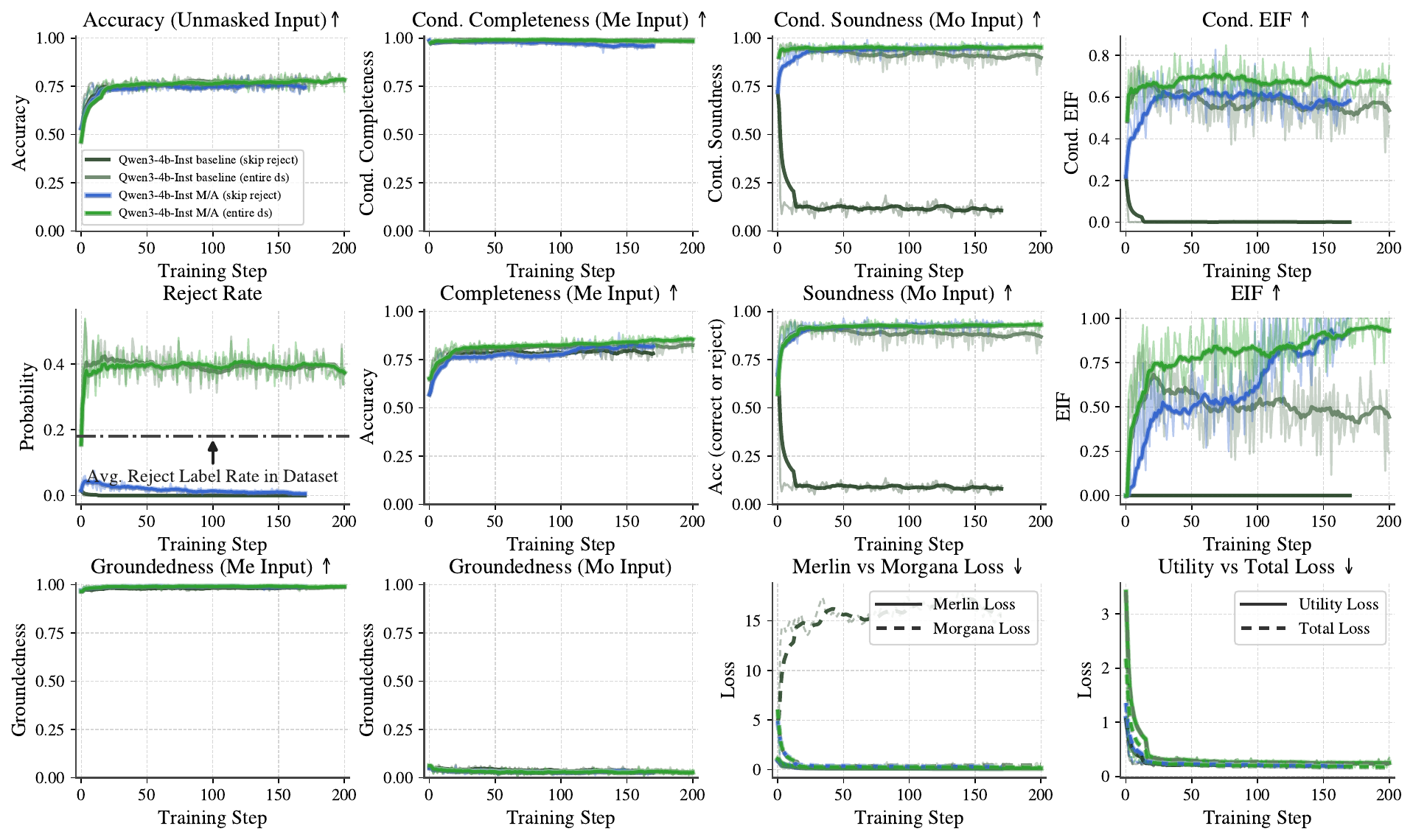}
        \caption{\textbf{Qwen3-4B-Instruct}}
    \end{subfigure}

    \begin{subfigure}[t]{\linewidth}
        \centering
        \includegraphics[width=\linewidth]{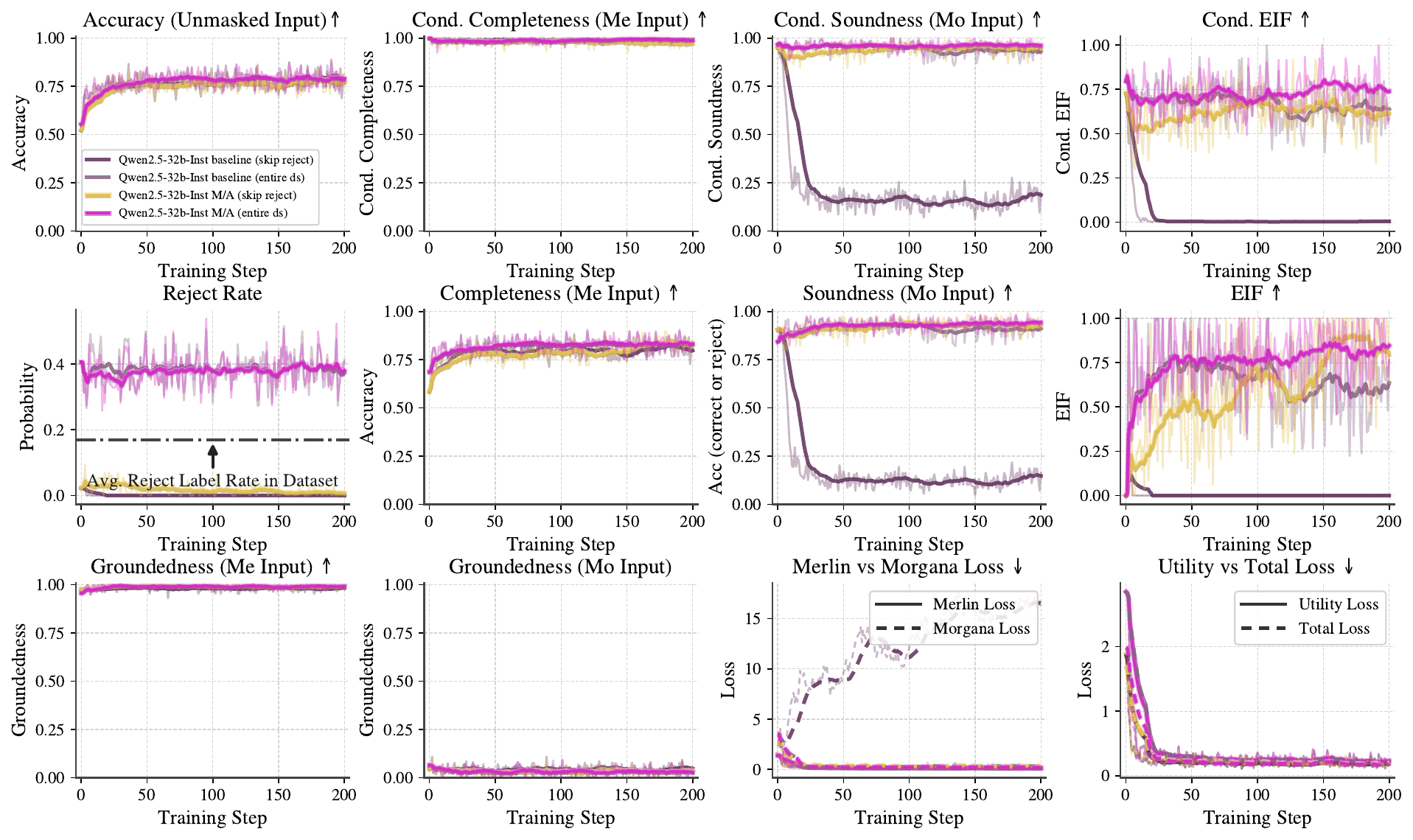}
        \caption{\textbf{Qwen2.5-32B-Instruct}}
    \end{subfigure}
   
    \caption{Training curves on \textbf{SQuAD} comparing \textbf{(i) sentence-level  \textit{M/A training} to (ii) \textit{baseline finetuning}} ($\lambda_{\text{util}} = 1$, $\lambda_{\text{Me}} = \lambda_{\text{Mo}} = 0$). Note that in the baseline setting, the Merlin and Morgana losses are not optimized -- they are only computed and plotted. Step~0 are metrics on the first batch \emph{before} any model updates, reflecting initial performance. Strong lines show a rolling-window average over 8\% of the total training datapoints. ``Shadow'' lines represent actual data.}
    \label{fig:RAG_train_curves_squad_app_qwen}
\end{figure*}

\begin{figure*}[t]
    \centering
    \begin{subfigure}[t]{\linewidth}
        \centering
        \includegraphics[width=\linewidth]{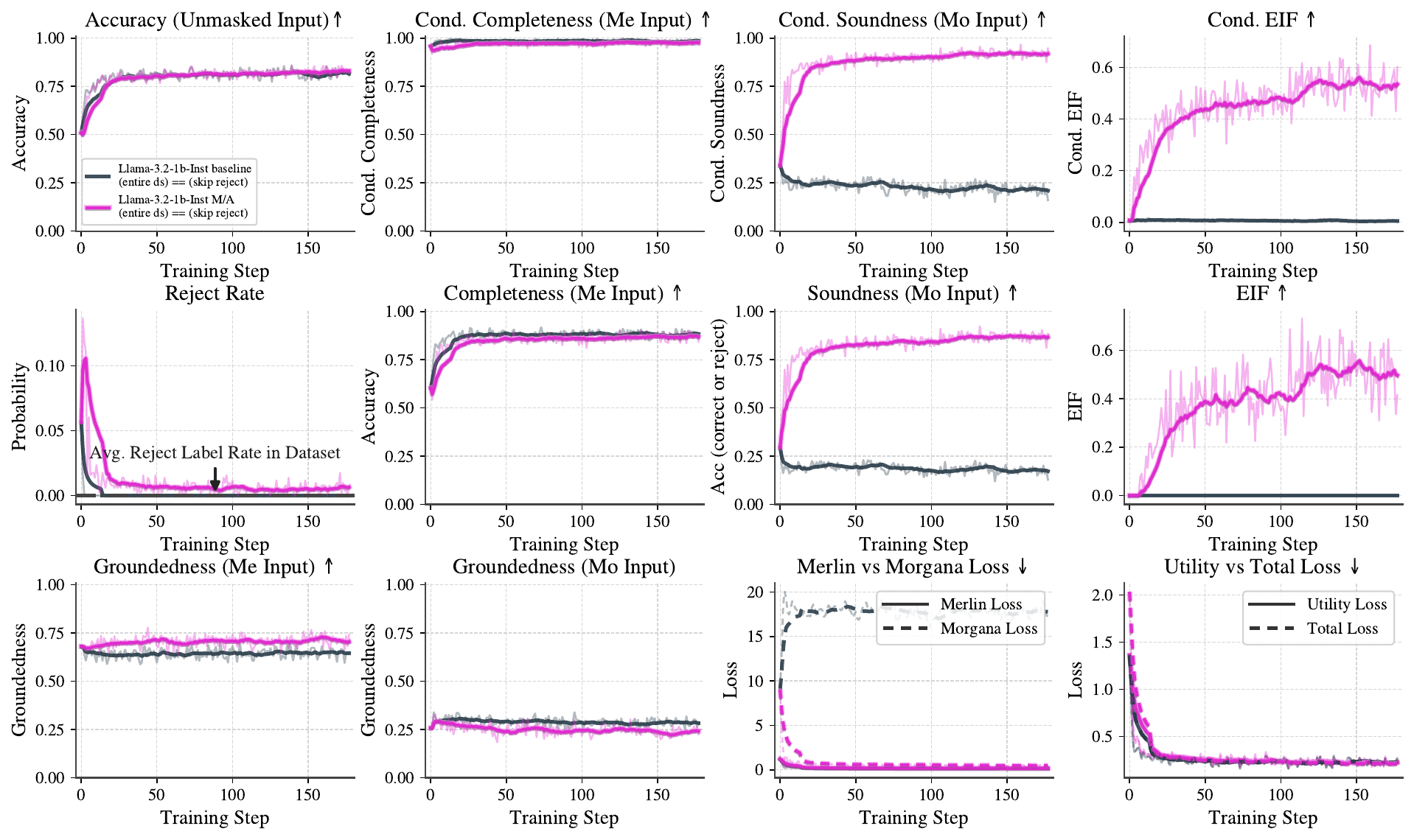}
        \caption{\textbf{Llama-3.2-1B-Instruct}}
    \end{subfigure}

    \begin{subfigure}[t]{\linewidth}
        \centering
        \includegraphics[width=\linewidth]{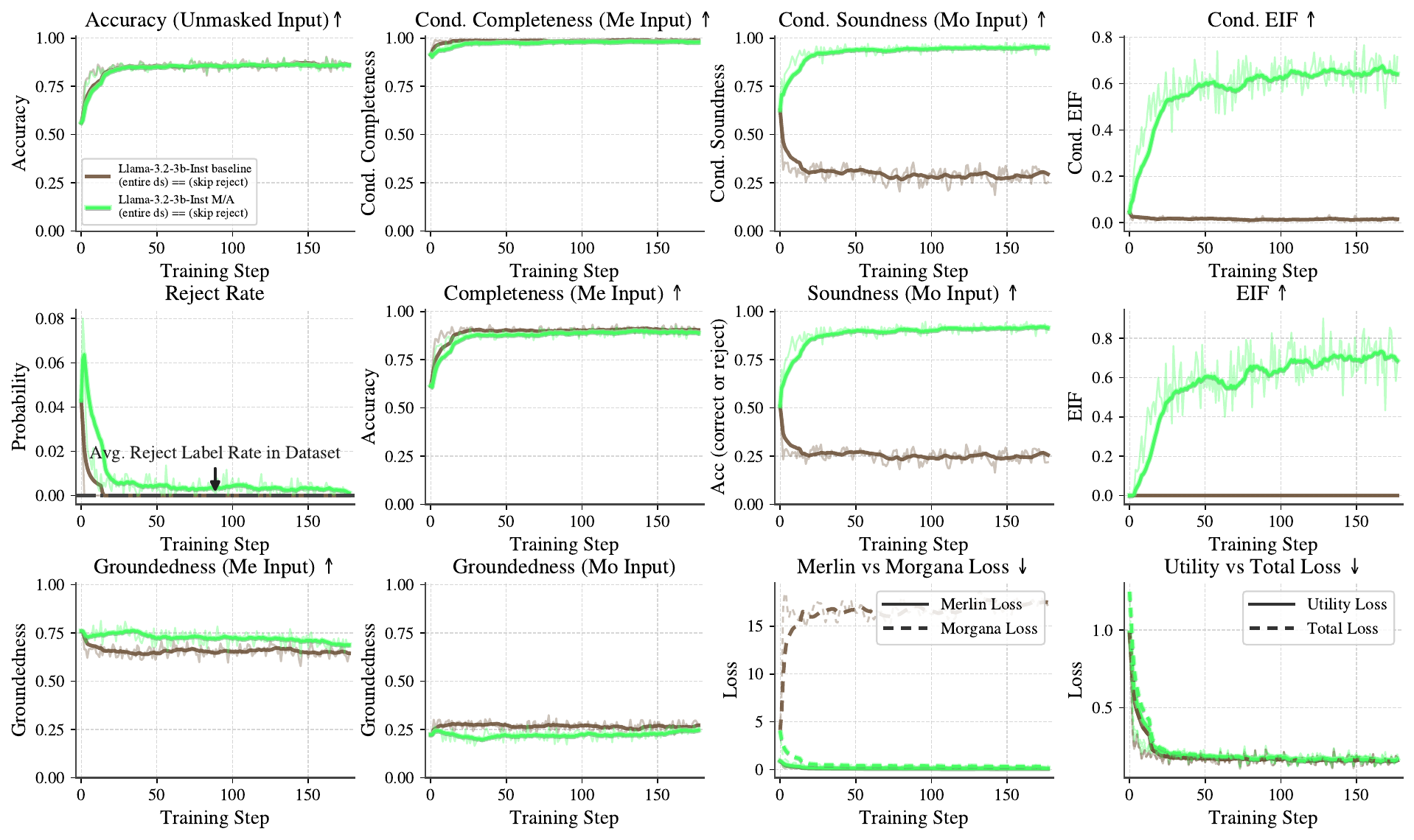}
        \caption{\textbf{Llama-3.2-3B-Instruct}}
    \end{subfigure}
    
    \caption{Training curves on \textbf{HotpotQA} comparing \textbf{(i) sentence-level  \textit{M/A training} to (ii) \textit{baseline finetuning}} ($\lambda_{\text{util}} = 1$, $\lambda_{\text{Me}} = \lambda_{\text{Mo}} = 0$). Note that in the baseline setting, the Merlin and Morgana losses are not optimized -- they are only computed and plotted. Qwen3 model results in Fig.~\ref{fig:RAG_train_curves_hotpot_app_qwen}.  Step~0 are metrics on the first batch \emph{before} any model updates, reflecting initial performance. Strong lines show a rolling-window average over 8\% of the total training datapoints. ``Shadow'' lines represent actual data.}
    \label{fig:RAG_train_curves_hotpot_app}
\end{figure*}

\begin{figure*}[t]
    \centering
    \begin{subfigure}[t]{\linewidth}
        \centering
        \includegraphics[width=\linewidth]{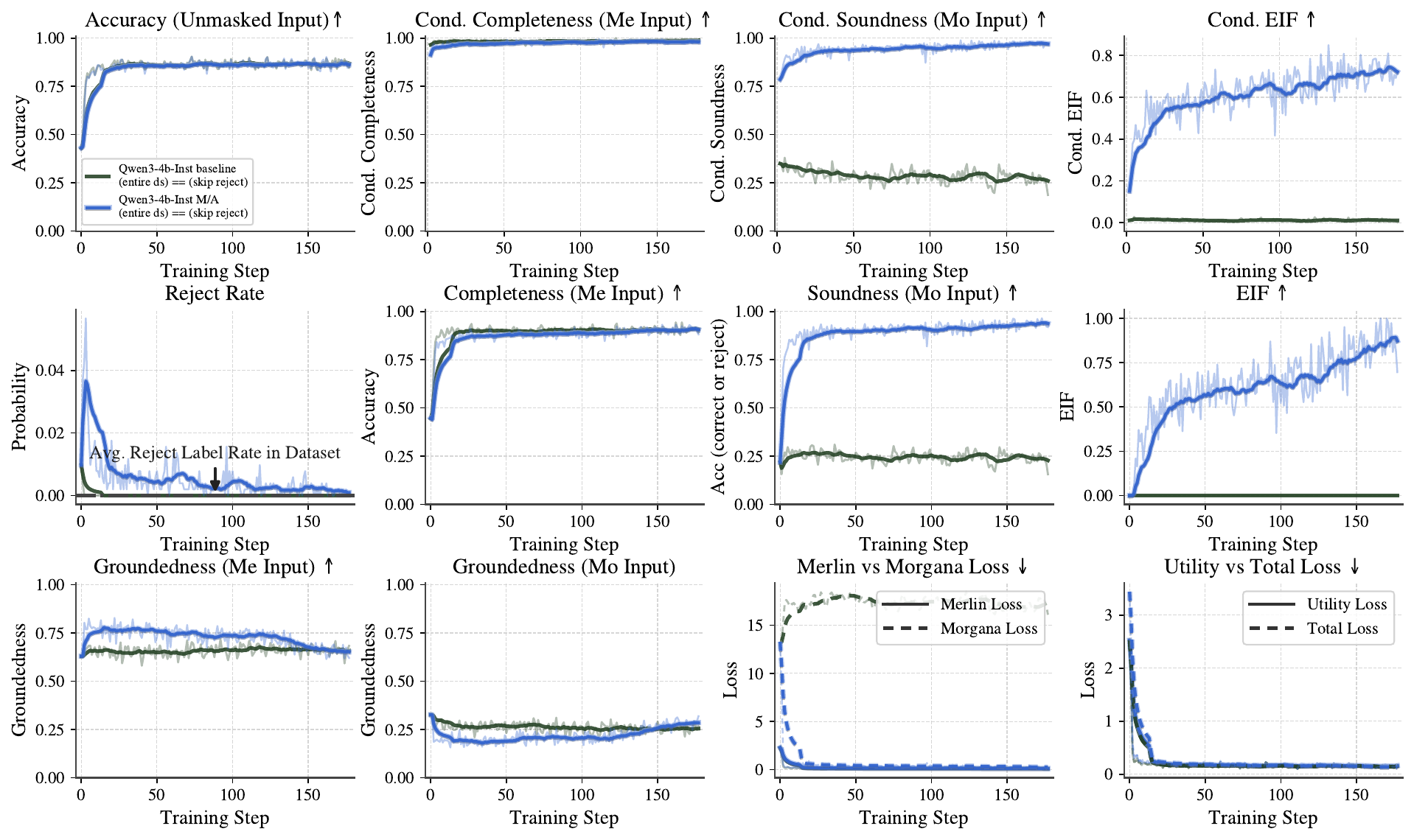}
        \caption{\textbf{Qwen3-4B-Instruct}}
    \end{subfigure}

    \begin{subfigure}[t]{\linewidth}
        \centering
        \includegraphics[width=\linewidth]{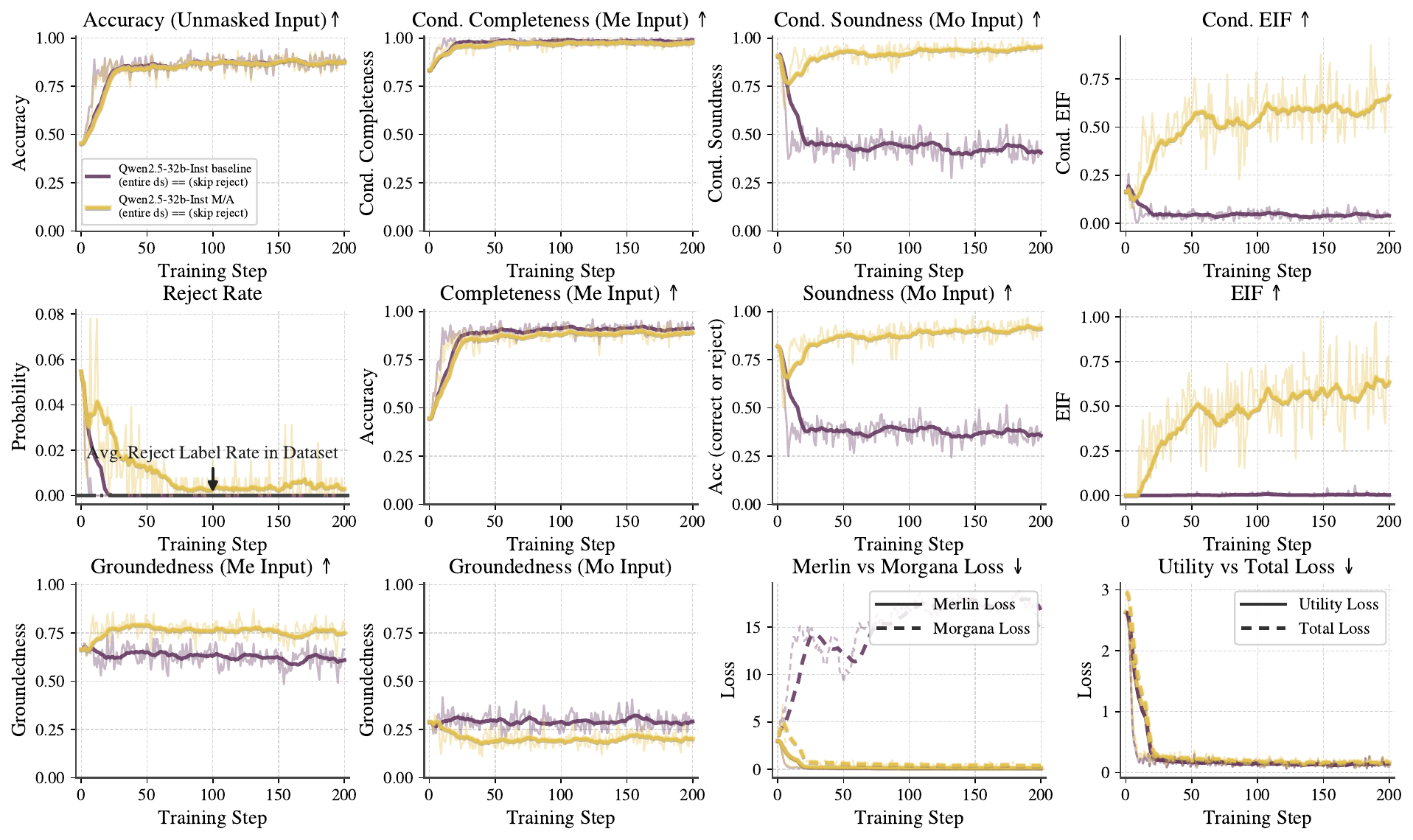}
        \caption{\textbf{Qwen2.5-32B-Instruct}}
    \end{subfigure}
   
    \caption{Training curves on \textbf{HotpotQA} comparing \textbf{(i) sentence-level  \textit{M/A training} to (ii) \textit{baseline finetuning}} ($\lambda_{\text{util}} = 1$, $\lambda_{\text{Me}} = \lambda_{\text{Mo}} = 0$). Note that in the baseline setting, the Merlin and Morgana losses are not optimized -- they are only computed and plotted. Step~0 are metrics on the first batch \emph{before} any model updates, reflecting initial performance. Strong lines show a rolling-window average over 8\% of the total training datapoints. ``Shadow'' lines represent actual data.}
    \label{fig:RAG_train_curves_hotpot_app_qwen}
\end{figure*}

\begin{figure*}[t]
    \centering
    \begin{subfigure}[t]{\linewidth}
        \centering
        \includegraphics[width=\linewidth]{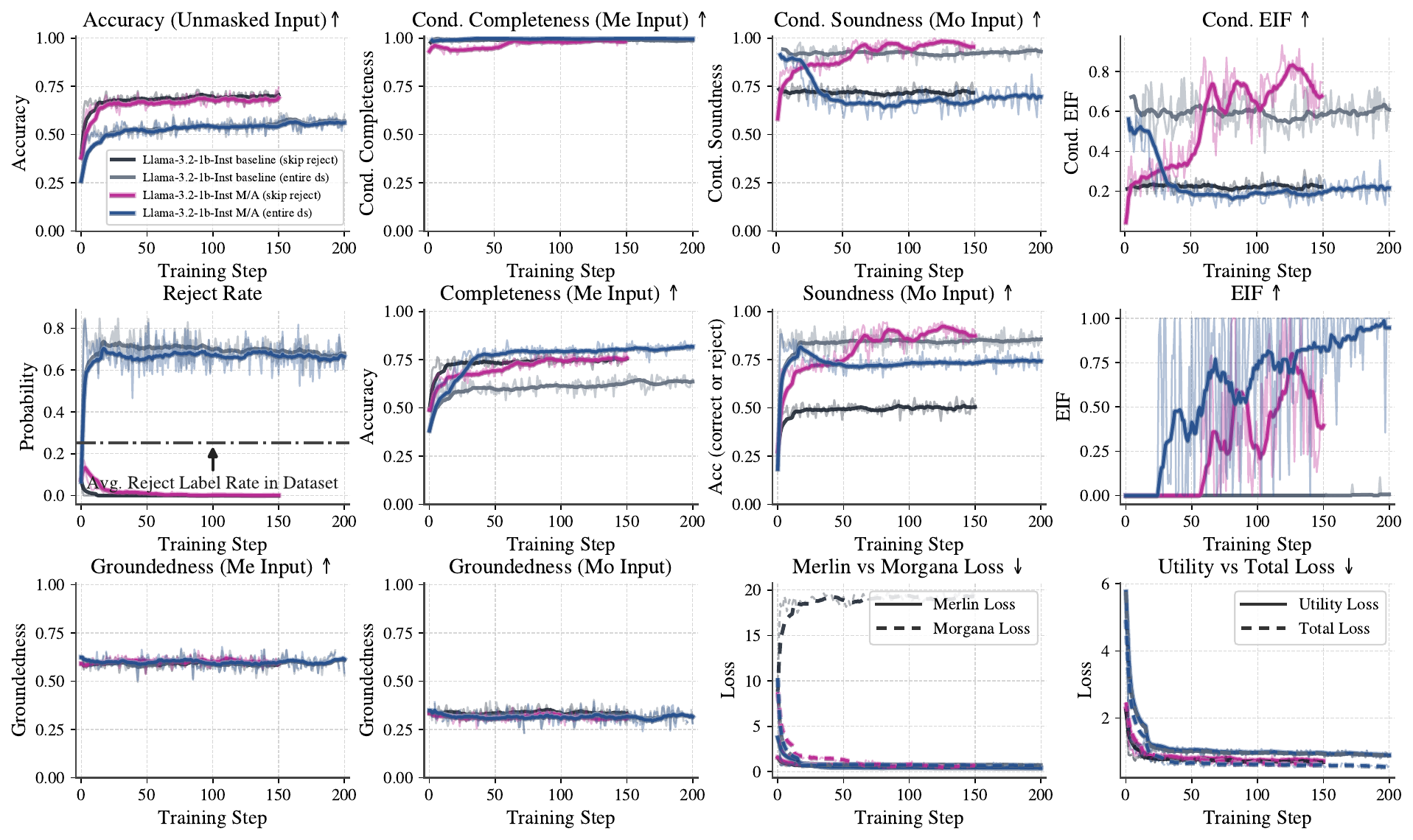}
        \caption{\textbf{Llama-3.2-1B-Instruct}}
    \end{subfigure}

    \begin{subfigure}[t]{\linewidth}
        \centering
        \includegraphics[width=\linewidth]{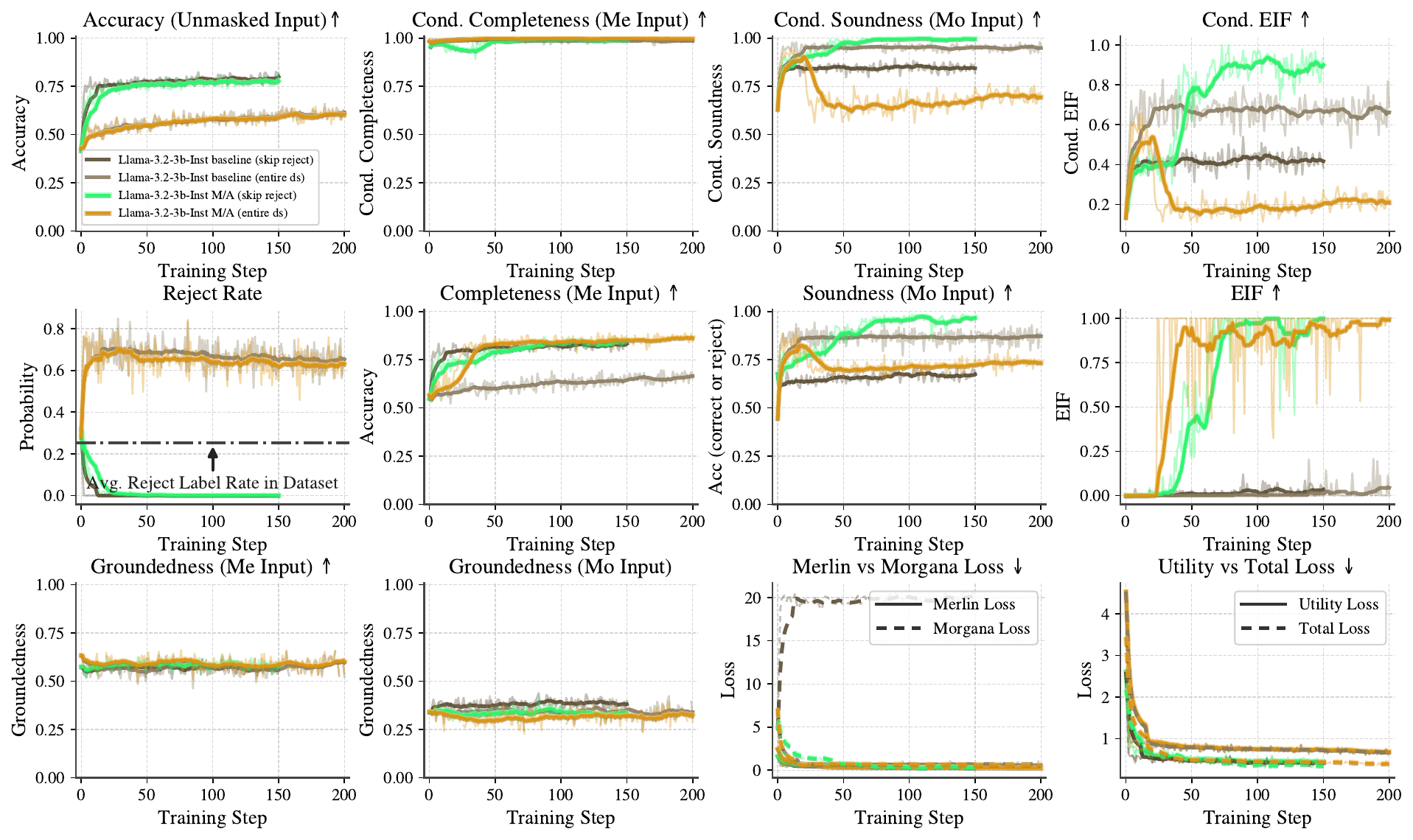}
        \caption{\textbf{Llama-3.2-3B-Instruct}}
    \end{subfigure}
    
    \caption{Training curves on \textbf{TriviaQA} comparing \textbf{(i) sentence-level  \textit{M/A training} to (ii) \textit{baseline finetuning}} ($\lambda_{\text{util}} = 1$, $\lambda_{\text{Me}} = \lambda_{\text{Mo}} = 0$). Note that in the baseline setting, the Merlin and Morgana losses are not optimized -- they are only computed and plotted. Qwen3 model results in Fig.~\ref{fig:RAG_train_curves_trivia_app_qwen}.  Step~0 are metrics on the first batch \emph{before} any model updates, reflecting initial performance. Strong lines show a rolling-window average over 8\% of the total training datapoints. ``Shadow'' lines represent actual data.}
    \label{fig:RAG_train_curves_trivia_app}
\end{figure*}

\begin{figure*}[t]
    \centering
    \begin{subfigure}[t]{\linewidth}
        \centering
        \includegraphics[width=\linewidth]{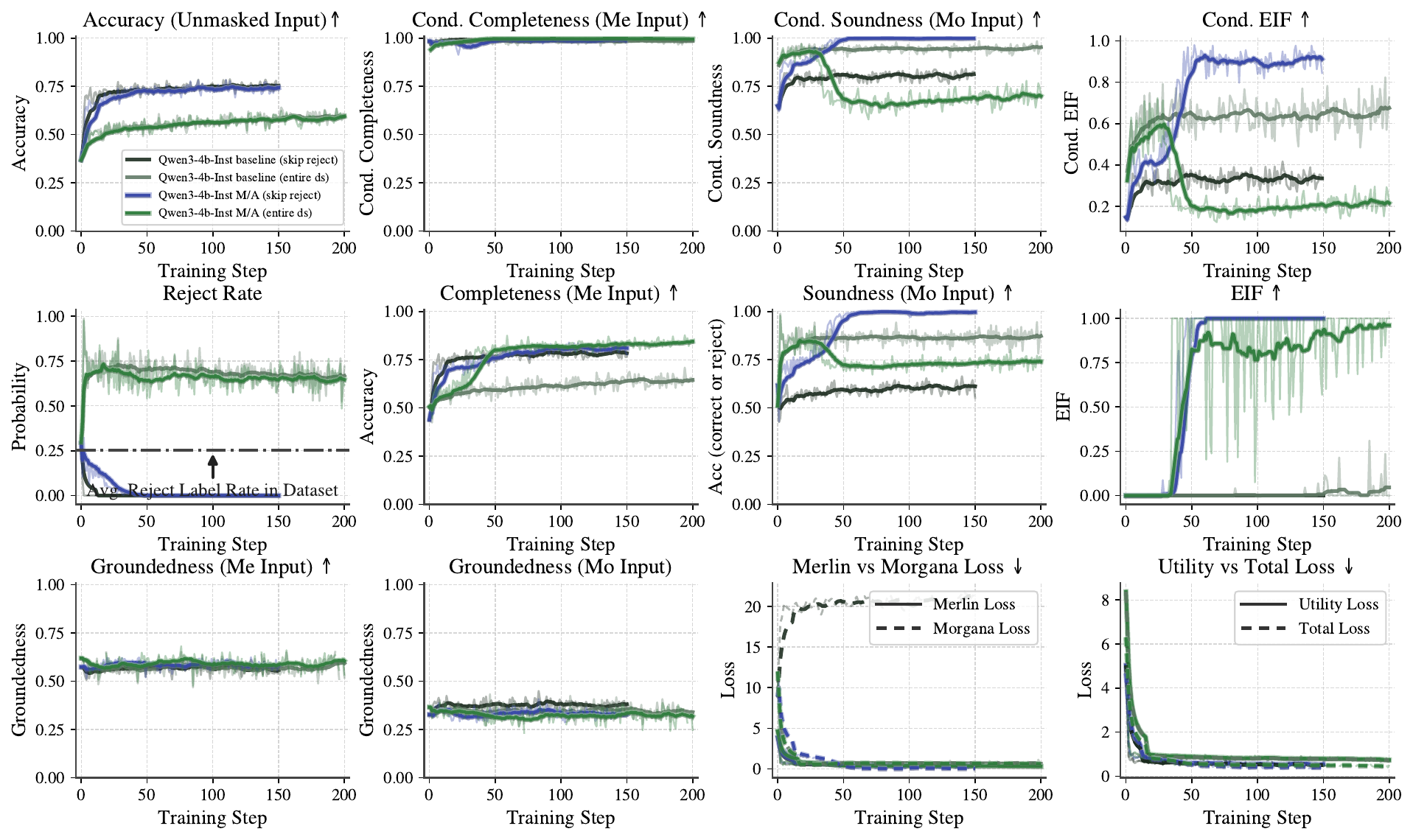}
        \caption{\textbf{Qwen3-4B-Instruct}}
    \end{subfigure}

    \begin{subfigure}[t]{\linewidth}
        \centering
        \includegraphics[width=\linewidth]{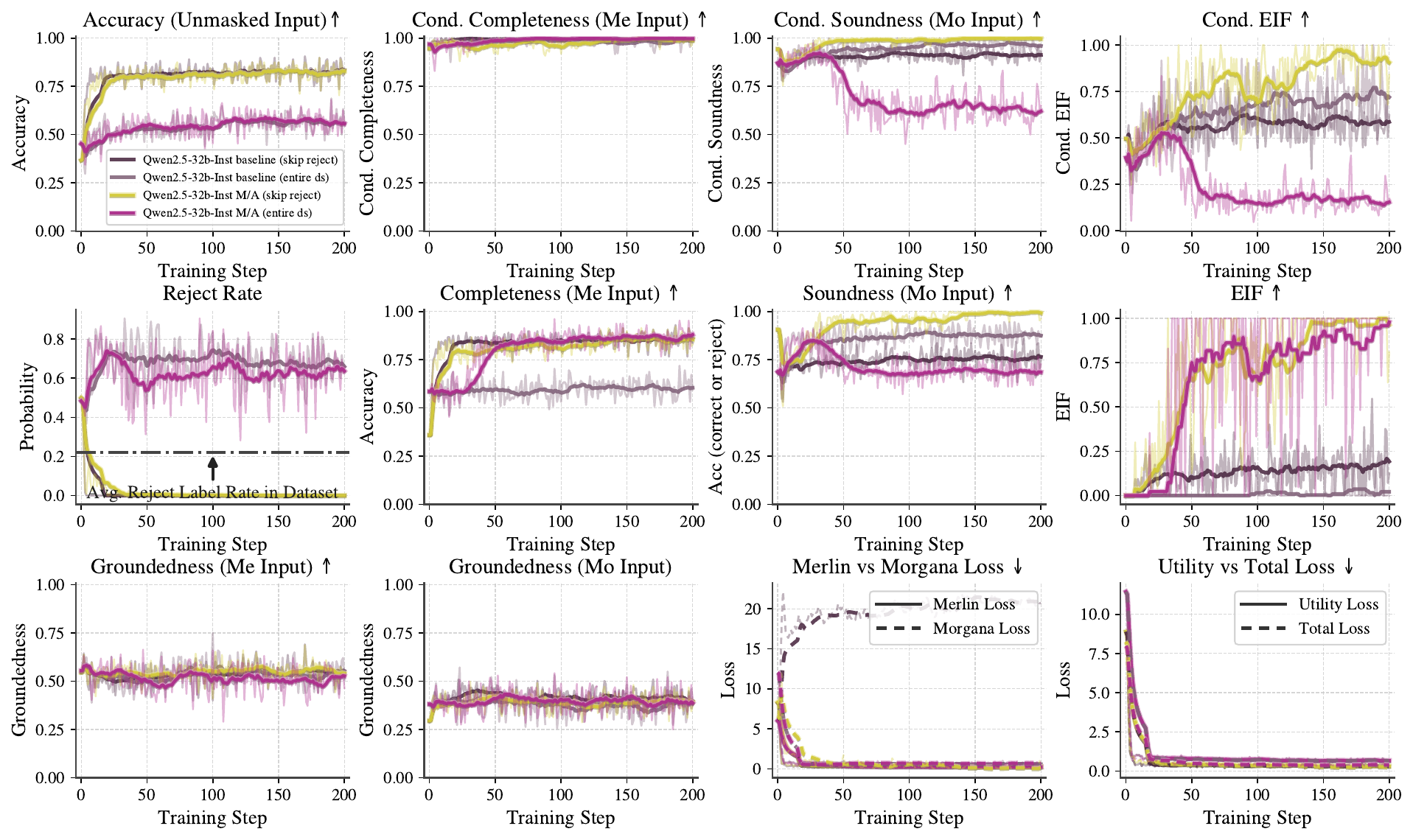}
        \caption{\textbf{Qwen2.5-32B-Instruct}}
    \end{subfigure}
   
    \caption{Training curves on \textbf{TriviaQA} comparing \textbf{(i) sentence-level  \textit{M/A training} to (ii) \textit{baseline finetuning}} ($\lambda_{\text{util}} = 1$, $\lambda_{\text{Me}} = \lambda_{\text{Mo}} = 0$). Note that in the baseline setting, the Merlin and Morgana losses are not optimized -- they are only computed and plotted. Step~0 are metrics on the first batch \emph{before} any model updates, reflecting initial performance. Strong lines show a rolling-window average over 8\% of the total training datapoints. ``Shadow'' lines represent actual data.}
    \label{fig:RAG_train_curves_trivia_app_qwen}
\end{figure*}

\begin{figure*}[t]
    \centering
    \begin{subfigure}[t]{\linewidth}
        \centering
        \includegraphics[width=\linewidth]{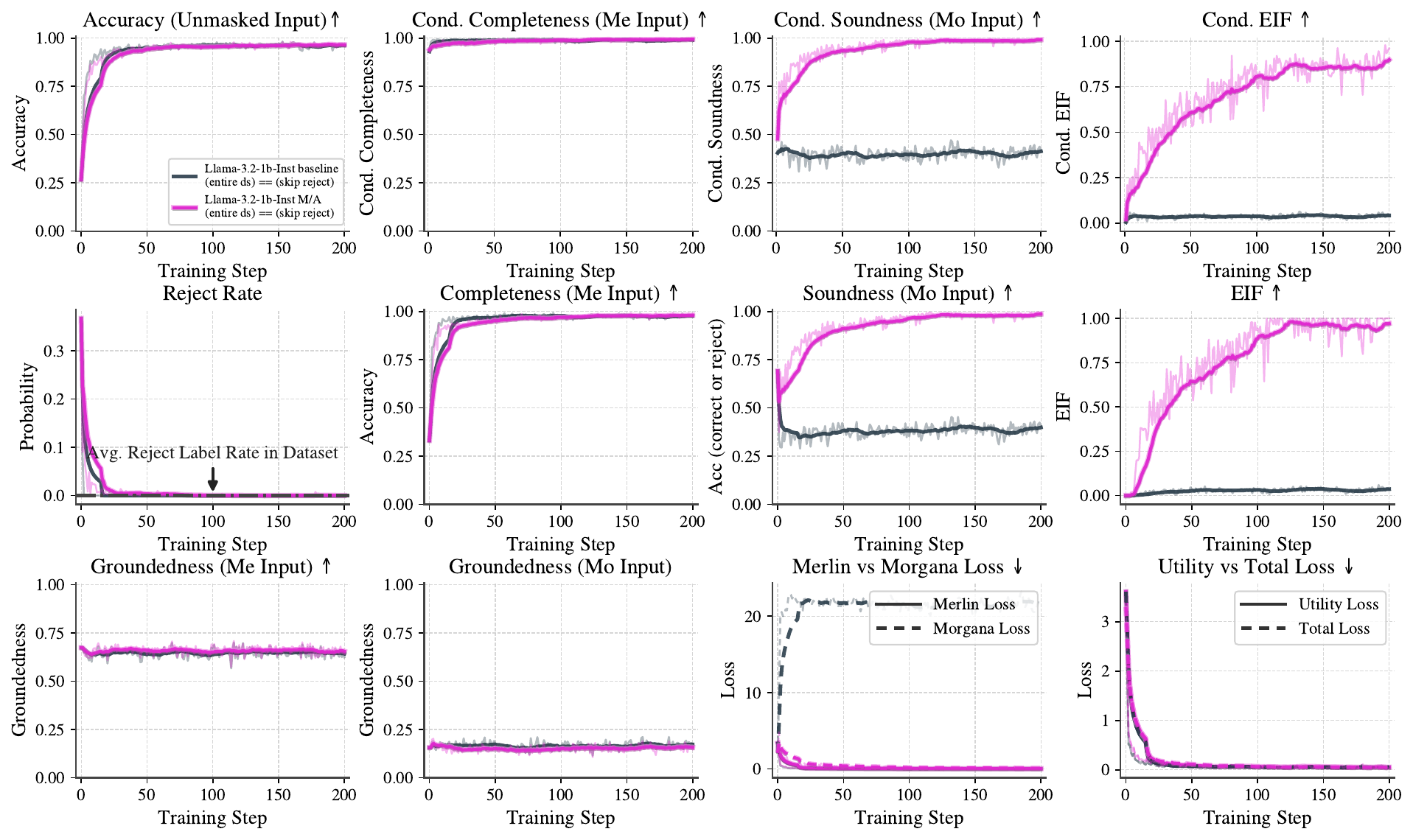}
        \caption{\textbf{Llama-3.2-1B-Instruct}}
    \end{subfigure}

    \begin{subfigure}[t]{\linewidth}
        \centering
        \includegraphics[width=\linewidth]{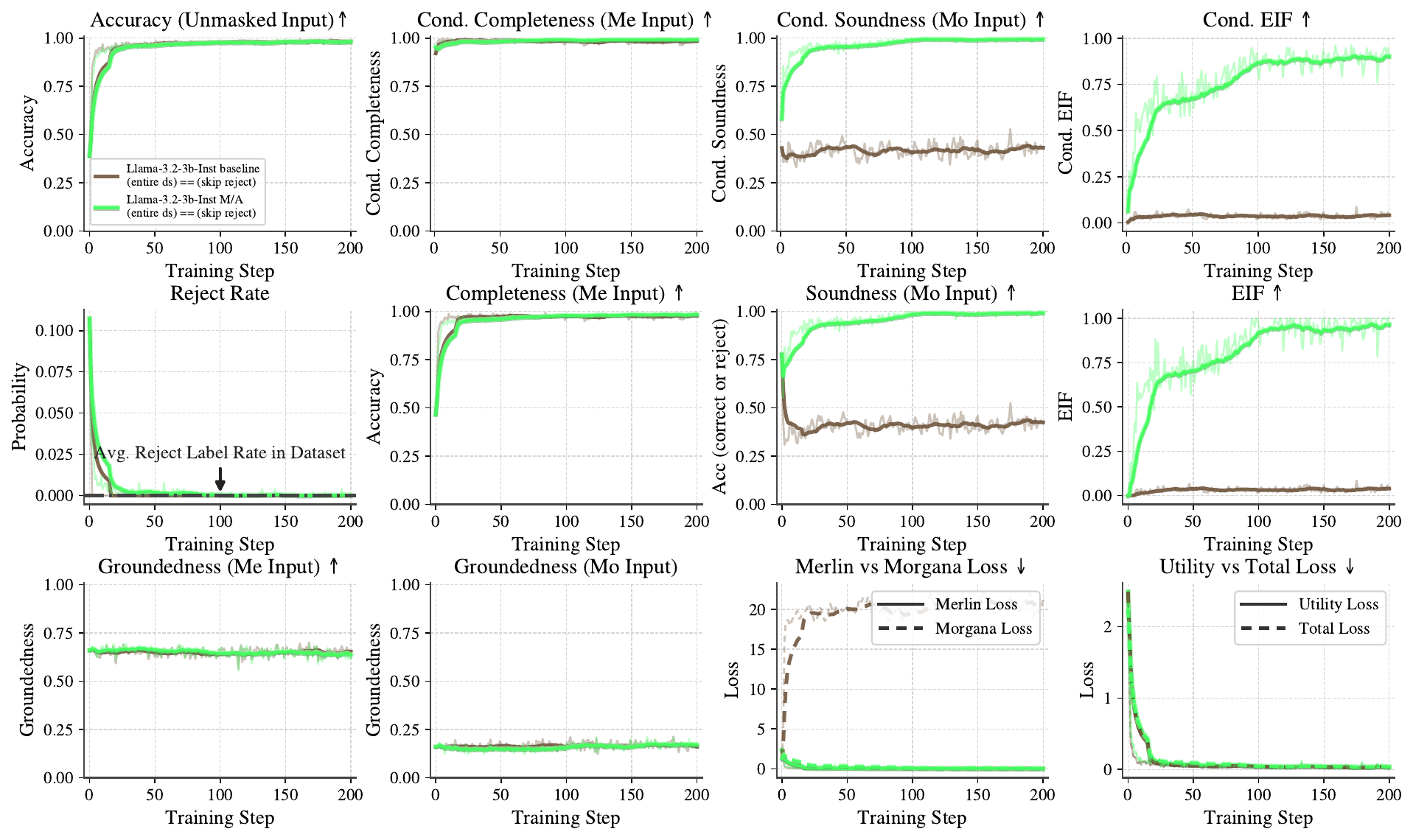}
        \caption{\textbf{Llama-3.2-3B-Instruct}}
    \end{subfigure}
    
    \caption{Training curves on \textbf{2WikiMultihop} comparing \textbf{(i) sentence-level  \textit{M/A training} to (ii) \textit{baseline finetuning}} ($\lambda_{\text{util}} = 1$, $\lambda_{\text{Me}} = \lambda_{\text{Mo}} = 0$). Note that in the baseline setting, the Merlin and Morgana losses are not optimized -- they are only computed and plotted. Qwen3 model results in Fig.~\ref{fig:RAG_train_curves_2wikimultihop_app_qwen}.  Step~0 are metrics on the first batch \emph{before} any model updates, reflecting initial performance. Strong lines show a rolling-window average over 8\% of the total training datapoints. ``Shadow'' lines represent actual data.}
    \label{fig:RAG_train_curves_2wikimultihop_app}
\end{figure*}

\begin{figure*}[t]
    \centering
    \begin{subfigure}[t]{\linewidth}
        \centering
        \includegraphics[width=\linewidth]{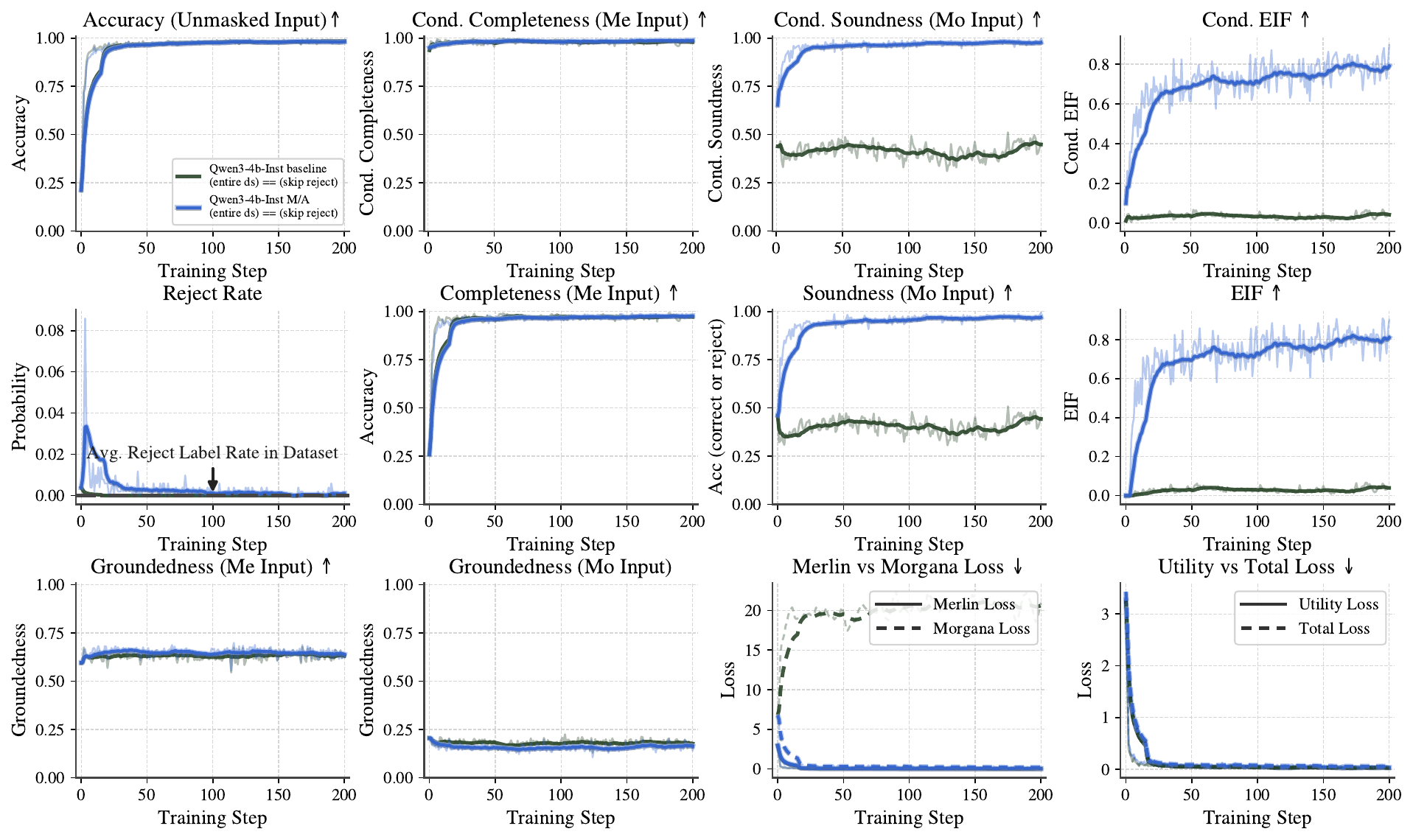}
        \caption{\textbf{Qwen3-4B-Instruct}}
    \end{subfigure}

    \begin{subfigure}[t]{\linewidth}
        \centering
        \includegraphics[width=\linewidth]{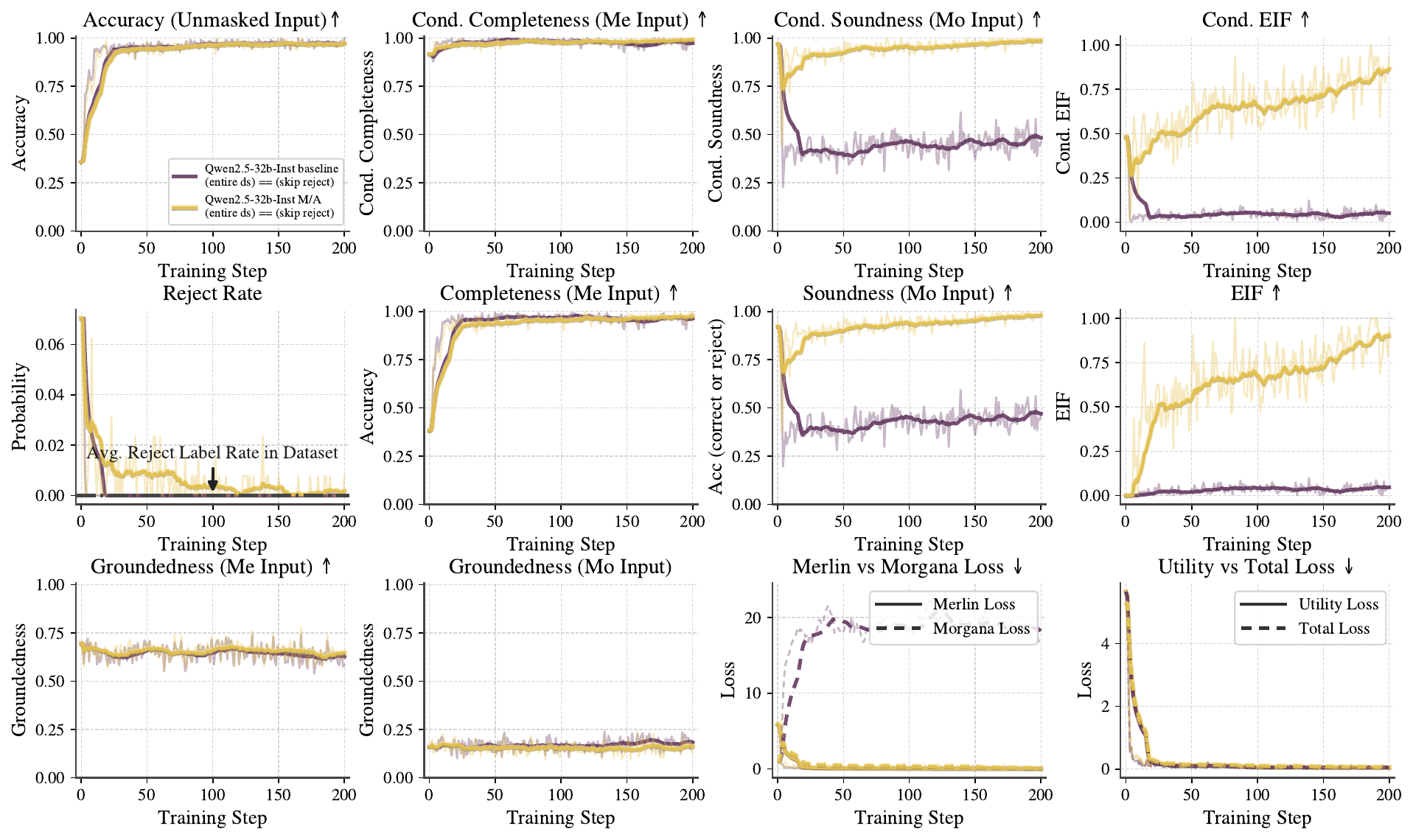}
        \caption{\textbf{Qwen2.5-32B-Instruct}}
    \end{subfigure}
   
    \caption{Training curves on \textbf{2WikiMultihop} comparing \textbf{(i) sentence-level  \textit{M/A training} to (ii) \textit{baseline finetuning}} ($\lambda_{\text{util}} = 1$, $\lambda_{\text{Me}} = \lambda_{\text{Mo}} = 0$). Note that in the baseline setting, the Merlin and Morgana losses are not optimized -- they are only computed and plotted. Step~0 are metrics on the first batch \emph{before} any model updates, reflecting initial performance. Strong lines show a rolling-window average over 8\% of the total training datapoints. ``Shadow'' lines represent actual data.}
    \label{fig:RAG_train_curves_2wikimultihop_app_qwen}
\end{figure*}

\begin{figure*}[t]
    \centering
    \begin{subfigure}[t]{\linewidth}
        \centering
        \includegraphics[width=\linewidth]{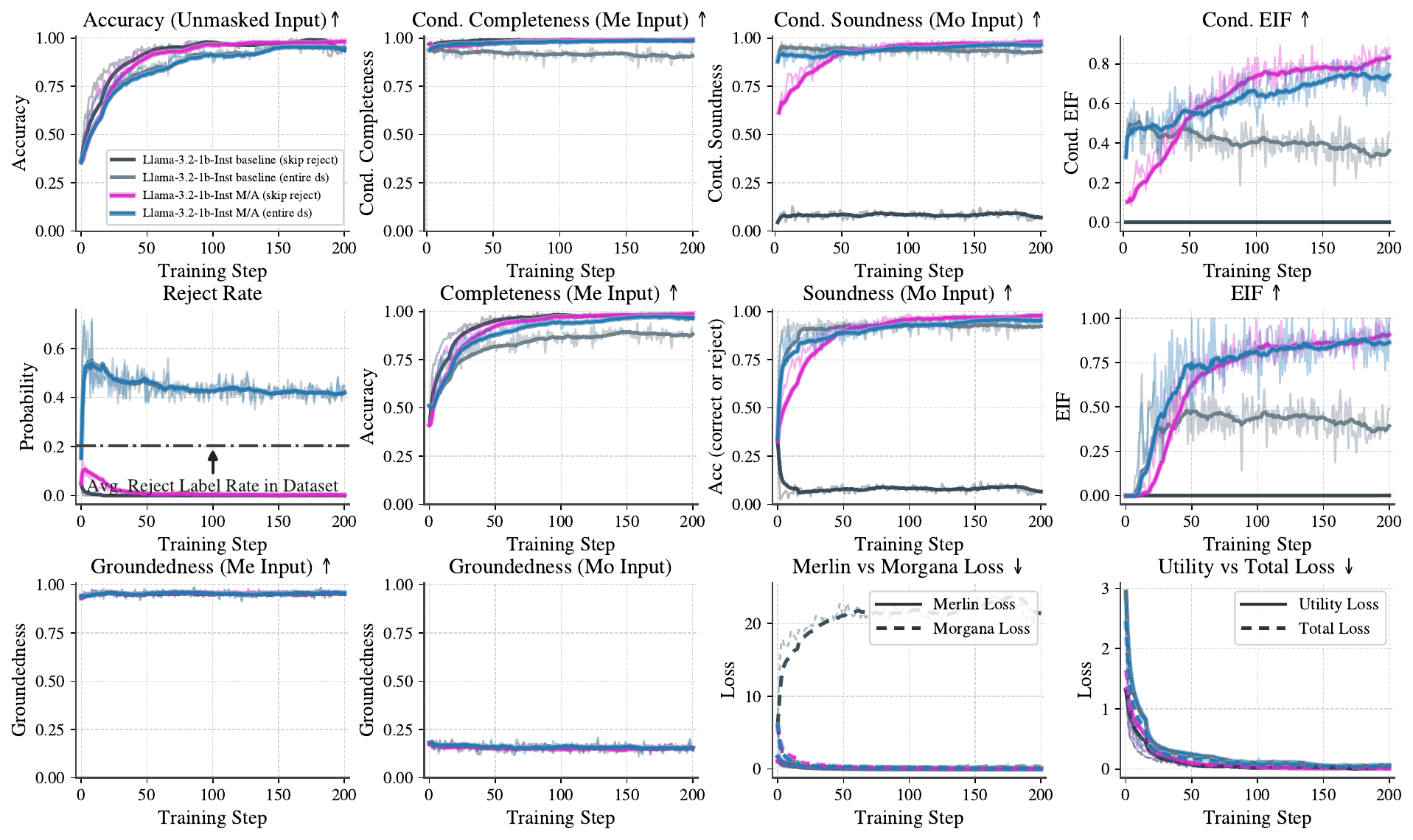}
        \caption{\textbf{Llama-3.2-1B-Instruct}}
    \end{subfigure}

    \begin{subfigure}[t]{\linewidth}
        \centering
        \includegraphics[width=\linewidth]{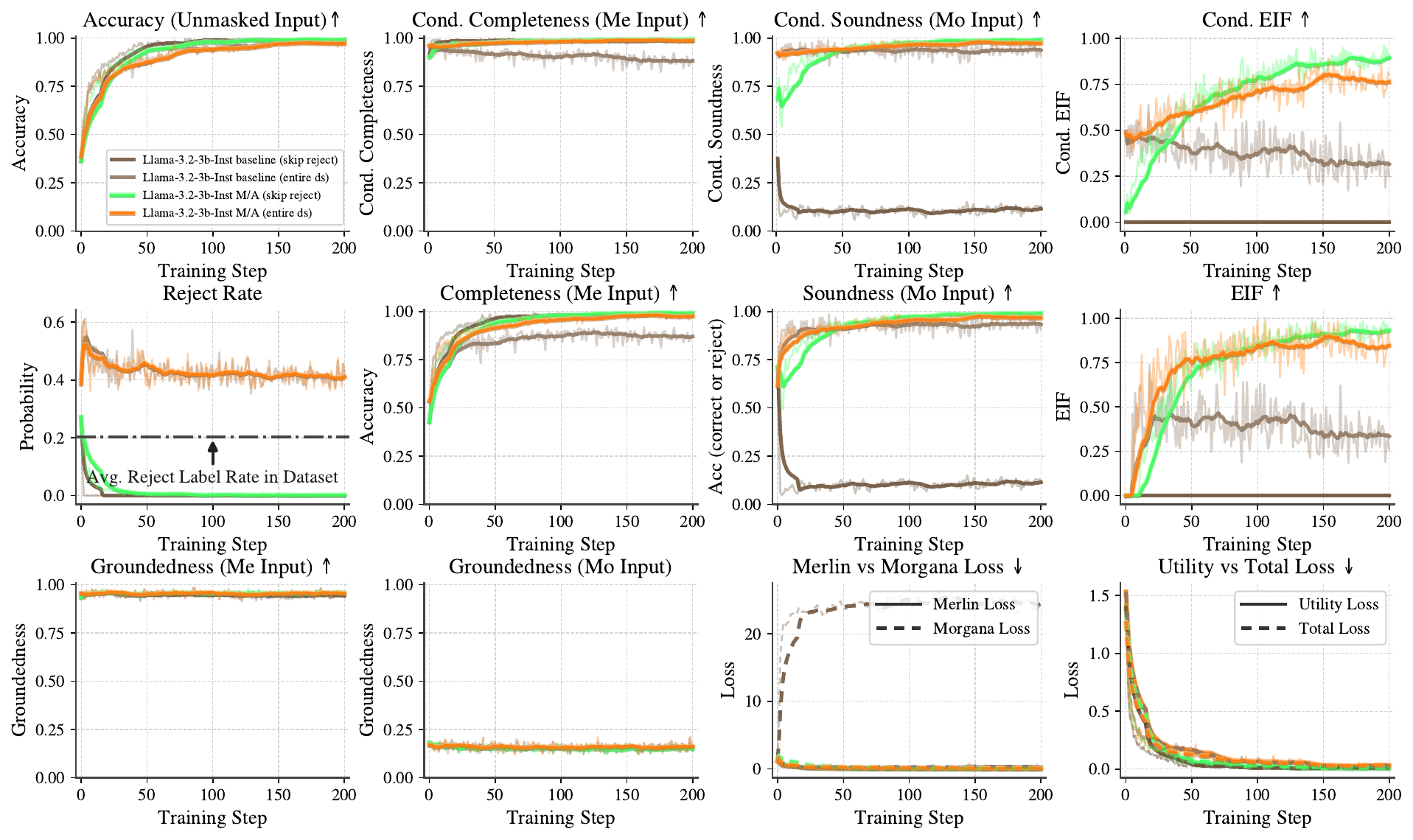}
        \caption{\textbf{Llama-3.2-3B-Instruct}}
    \end{subfigure}
    
    \caption{Training curves on \textbf{MuSiQue} comparing \textbf{(i) sentence-level  \textit{M/A training} to (ii) \textit{baseline finetuning}} ($\lambda_{\text{util}} = 1$, $\lambda_{\text{Me}} = \lambda_{\text{Mo}} = 0$). Note that in the baseline setting, the Merlin and Morgana losses are not optimized -- they are only computed and plotted. Qwen3 model results in Fig.~\ref{fig:RAG_train_curves_musique_app_qwen}.  Step~0 are metrics on the first batch \emph{before} any model updates, reflecting initial performance. Strong lines show a rolling-window average over 8\% of the total training datapoints. ``Shadow'' lines represent actual data.}
    \label{fig:RAG_train_curves_musique_app}
\end{figure*}

\begin{figure*}[t]
    \centering
    \begin{subfigure}[t]{\linewidth}
        \centering
        \includegraphics[width=\linewidth]{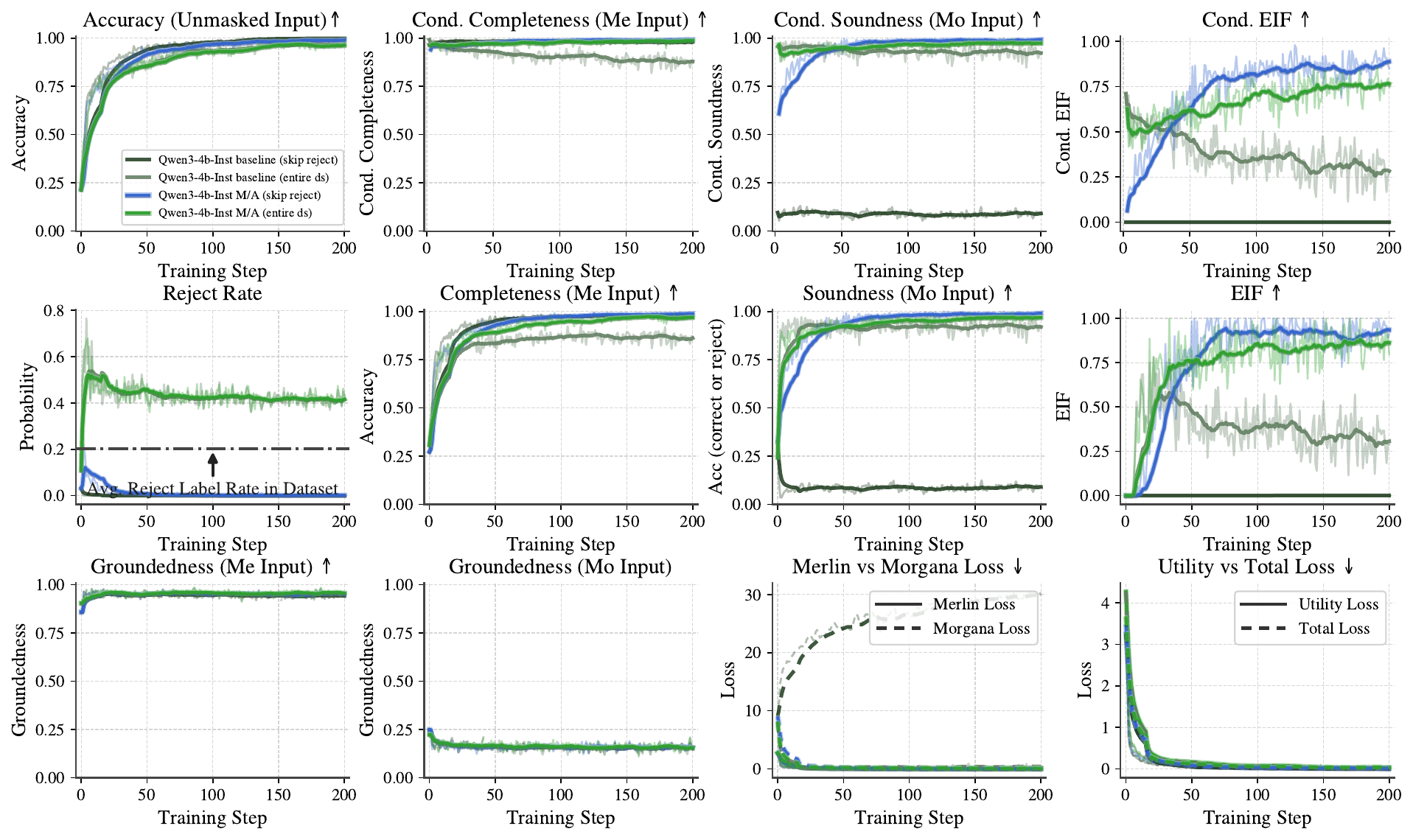}
        \caption{\textbf{Qwen3-4B-Instruct}}
    \end{subfigure}

    \begin{subfigure}[t]{\linewidth}
        \centering
        \includegraphics[width=\linewidth]{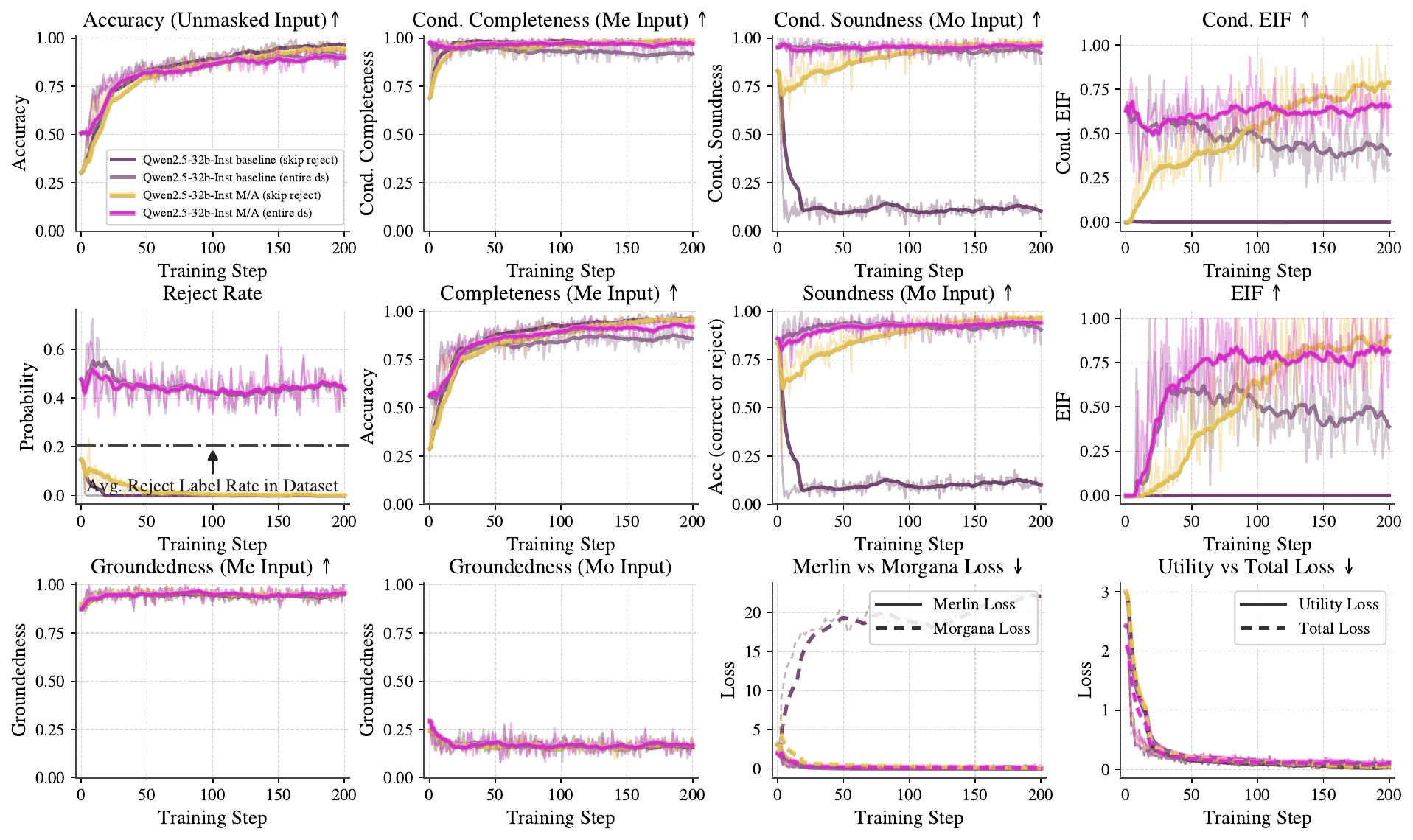}
        \caption{\textbf{Qwen2.5-32B-Instruct}}
    \end{subfigure}
   
    \caption{Training curves on \textbf{MuSiQue} comparing \textbf{(i) sentence-level  \textit{M/A training} to (ii) \textit{baseline finetuning}} ($\lambda_{\text{util}} = 1$, $\lambda_{\text{Me}} = \lambda_{\text{Mo}} = 0$). Note that in the baseline setting, the Merlin and Morgana losses are not optimized -- they are only computed and plotted. Step~0 are metrics on the first batch \emph{before} any model updates, reflecting initial performance. Strong lines show a rolling-window average over 8\% of the total training datapoints. ``Shadow'' lines represent actual data.}
    \label{fig:RAG_train_curves_musique_app_qwen}
\end{figure*}

Notably, HotpotQA and  2WikiMultihopQA do not contain unanswerable questions. Also, we simulate the setting of missing reject annoation by the ''skip reject`` setting in our results (Figures~\ref{fig:RAG_training_barplot_avg},\ref{fig:RAG_training_barplot_full},\ref{fig:RAG_train_curves_squad_app}-\ref{fig:RAG_train_curves_musique_app_qwen}). In such instances, vanilla finetuning (the baseline we compare M/A training to) exhibits particularly low soundness. When datasets do not contain unanswerable questions, and they do not provide any explicit training signal for abstention. As a result, even though we explicitly instruct all models to reject unsupported questions via prompting (cf.~App.~\ref{app:prompt}), the vanilla-finetuned model fails to learn reliable reject behavior, leading to frequent hallucinations under adversarial or insufficient evidence. \textbf{In contrast, M/A training enables it to learn abstention behavior without requiring annotated unanswerable examples}, and induces strong soundness on setting such as HotpotQA and 2WikiMultihopQA, by exposing the model to systematically constructed adversarial contexts through Morgana.

\subsubsection{Training: Comparing Sentence- with Token-level Masking} \label{app:sentence-vs-token-level}
Fig.~\ref{fig:RAG_train_curves_squad_sentence_token_app} compares token-  and sentence-level masking. Both yield similar training dynamics and final accuracy, completeness and soundness. Sentence-level masking reaches higher groundedness, because its smaller search space makes it more likely to target the sentences that contain or disrupt the evidence.

\subsubsection{Training: Comparing Attention-based and String-based Masking} \label{app:attention-vs-string-level}
\begin{figure*}[t]
    \centering
    \begin{subfigure}[t]{\linewidth}
        \centering
        \includegraphics[width=\linewidth]{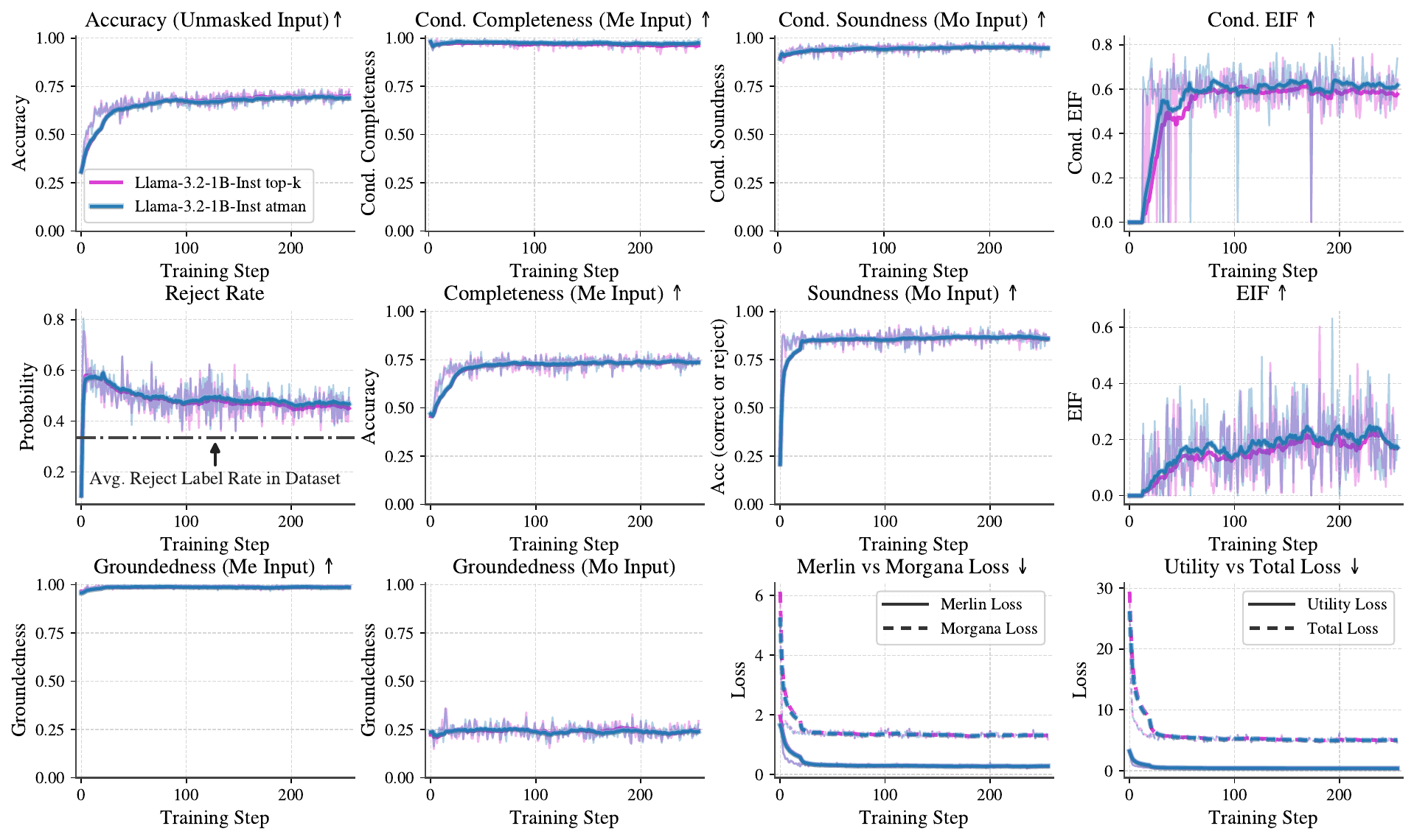}
        \caption{\textbf{Llama-3.2-1B-Instruct:} M/A training in two variants: (a) attention- (blue) and (b) string-level masking (pink). Discussion in \ref{app:attention-vs-string-level}.}
    \end{subfigure}
    
    \vspace{0.5em}

    \begin{subfigure}[t]{\linewidth}
        \centering
        \includegraphics[width=\linewidth]{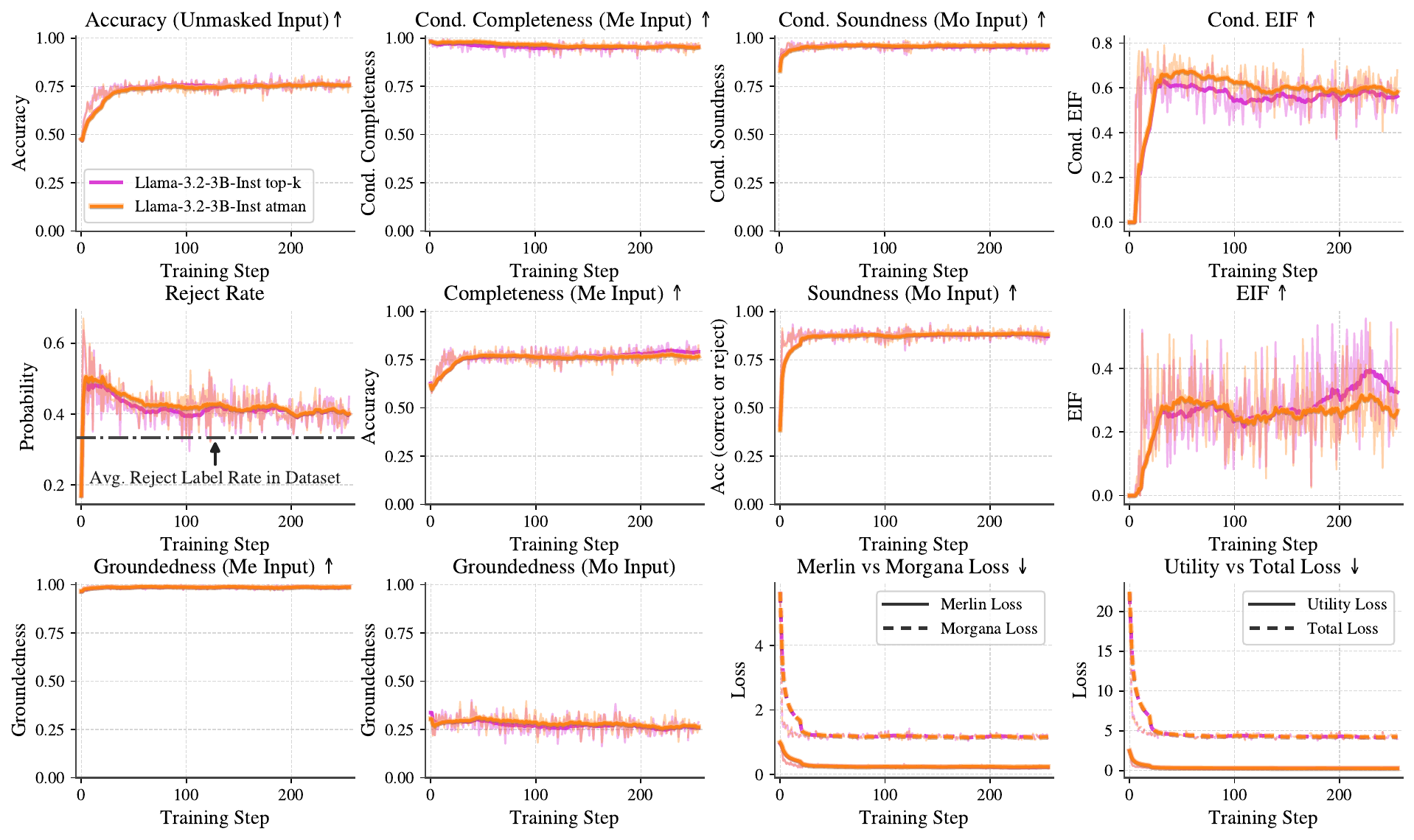}
        \caption{\textbf{Llama-3.2-3B-Instruct:} M/A training in two variants: (a) attention- (orange) and (b) string-level masking (pink). Discussion in \ref{app:attention-vs-string-level}.}
    \end{subfigure}
    
    \caption{\textbf{M/A training in two variants: (a) attention- and (b) string-level masking strategies} on SQuAD. See \ref{app:attention-vs-string-level} for discussion.  Step~0 are metrics on the first batch \emph{before} any model updates, reflecting initial performance. Strong lines show a rolling-window average over 8\% of the total training datapoints. ``Shadow'' lines represent actual data.}
    \label{fig:RAG_train_curves_squad_atman_vs_topk}
\end{figure*}

In Fig.~\ref{fig:RAG_train_curves_squad_atman_vs_topk}, we compare attention-based masking with string masking. Overall, attention-based masking and string-level masking yield comparable training dynamics
across metrics. Attention masking additionally  avoids inserting artificial \texttt{[MASK]} tokens and preserves the positional embeddings of the input.

\subsubsection{Computational Overhead and Autonomous Scaling}
\label{sec:time_estimates}

The M/A framework exemplifies \textbf{model-driven scaling}: rather than relying on static, human-annotated datasets, the supervision signal is dynamically generated to target the model's specific weaknesses as they evolve during training. By autonomously augmenting the training signal, Merlin and Morgana identify and ``patch'' emerging hallucinations in a model-dependent manner. This approach demonstrates that scaling computational investment into autonomous, interactive-proof-style supervision is a highly effective alternative to scaling manual annotation effort in creating static data. It aligns with the broader trend that scaling data and compute-driven interventions yields superior accuracy, particularly when those interventions are targeted at the model's internal causal reasoning rather than simple correlations.

Our alignment phase converges after only a very few hundred steps across all  datasets. The one-time investment is marginal relative to lifecycle inference costs. Current SOTA hallucination-mitigation strategies rely on creating / annotated data with paid inference APIs (e.g. GPT-4), which are orders of magnitude more expensive and capped by the teacher model's capabilities. Our framework provides annotation-free augmentation, especially valuable in specialized domains.

The computational overhead of Merlin-Arthur (M/A) training is primarily governed by the evaluation of three distinct loss objectives and the automated mask generation phase.

\paragraph{Default Training Setting (Sentence-Level).} 
We designate \textbf{sentence-level masking as the default configuration}, as it provides the optimal balance between our core metrics -- model utility (accuracy), soundness, and completeness -- and computational throughput. In this setting, the total training time per step across tested architectures (Llama-3.2-1B, 3B, and Qwen3-4B) is approximately 3.5--4$\times$ that of a baseline SFT step. This factor is composed of the 3$\times$ loss computation for $\mathcal{L}_{\text{util}}, \mathcal{L}_{\text{Me}},$ and $\mathcal{L}_{\text{Mo}}$, supplemented by a marginal \textbf{0.5--0.6$\times$} overhead for the automated mask generation phase.

\begin{figure*}[t]
    \centering
    \begin{subfigure}[t]{\linewidth}
        \centering
        \includegraphics[width=\linewidth]{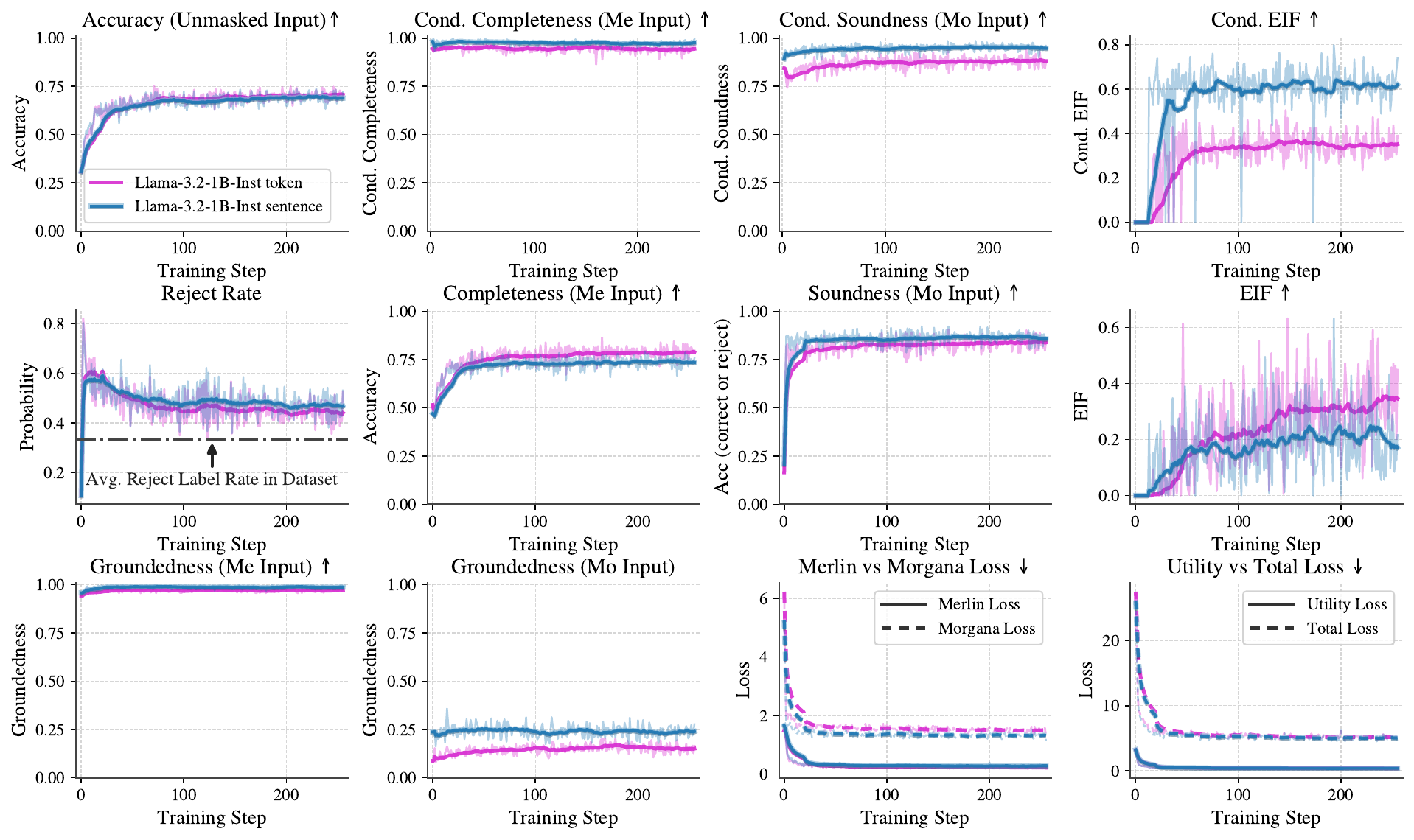}
        \caption{\textbf{Llama-3.2-1B-Instruct:} M/A training in 2 variants: (a) sentence-level (blue) and (b) token-level masking (pink). Discussion in \ref{app:sentence-vs-token-level}.}
    \end{subfigure}
    
    \vspace{0.5em}

    \begin{subfigure}[t]{\linewidth}
        \centering
        \includegraphics[width=\linewidth]{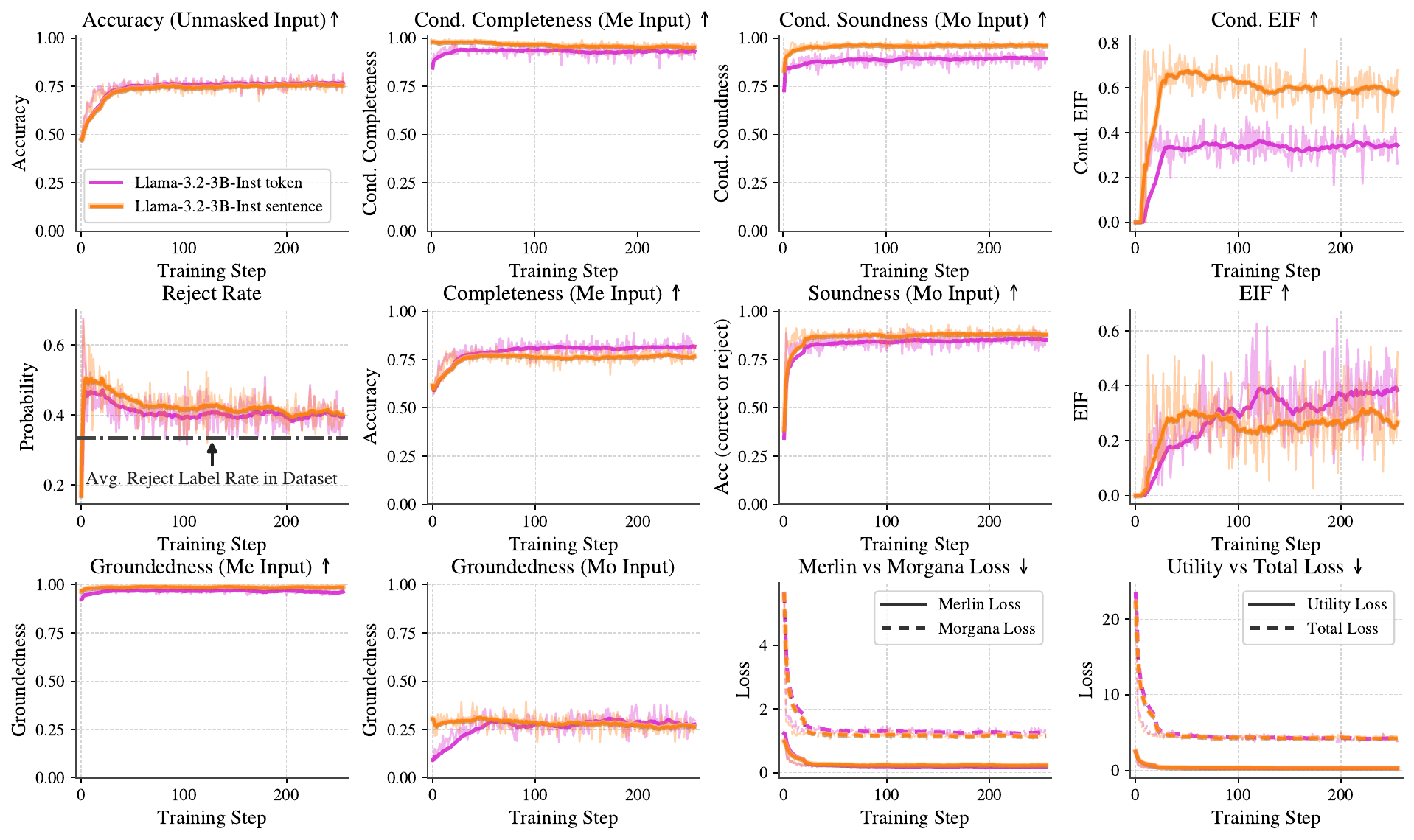}
        \caption{\textbf{Llama-3.2-3B-Instruct:} M/A training in 2 variants: (a) sentence-level (orange) and (b) token-level masking (pink). Discussion in \ref{app:sentence-vs-token-level}.}
    \end{subfigure}
    
    \caption{\textbf{M/A training in two variants: (a) sentence-level and (b) token-level} masking procedures in training on SQuAD. See \ref{app:sentence-vs-token-level} for discussion.  Step~0 are metrics on the first batch \emph{before} any model updates, reflecting initial performance. Strong lines show a rolling-window average over 8\% of the total training datapoints. ``Shadow'' lines represent actual data.}
    \label{fig:RAG_train_curves_squad_sentence_token_app}
\end{figure*}

\paragraph{Comparison with Token-Level Masking.} 
While token-level masking enables fine-grained attribution, it is significantly more intensive due to the expanded search space for the \textsc{AtMan}-based prover. Token-level generation introduces an overhead of 3.5--5$\times$ relative to a baseline step, resulting in total step times roughly 6.5--8.5$\times$ longer than standard SFT. Because sentence-level masking achieves comparable soundness and completeness gains (as shown in~\ref{app:sentence-vs-token-level} Fig.~\ref{fig:RAG_train_curves_squad_sentence_token_app}) with vastly lower latency, it remains our recommended standard for general RAG applications.

\paragraph{Scalability for Long Contexts.} 
M/A training scales effectively through flexible masking granularity. Our experiments on TriviaQA -- which features long contexts -- demonstrate that the prover remains efficient in long-context retrieval settings. Critically, switching granularity from token-level to sentence-level maintains the fidelity of the supervision signal -- see~\ref{app:sentence-vs-token-level} Fig.~\ref{fig:RAG_train_curves_squad_sentence_token_app}. For even larger documents, the prover can be configured to operate on paragraph-level units or arbitrary semantic chunks. Since the \textsc{AtMan} explainer maintains linear complexity relative to the number of units, this hierarchical chunking ensures that M/A supervision remains computationally tractable as the context window grows.

% -------------

\clearpage

\section{M/A Retriever Training}\label{app:retriever_training}

\begin{figure}[th!]
\centering
\includegraphics[width=\linewidth]{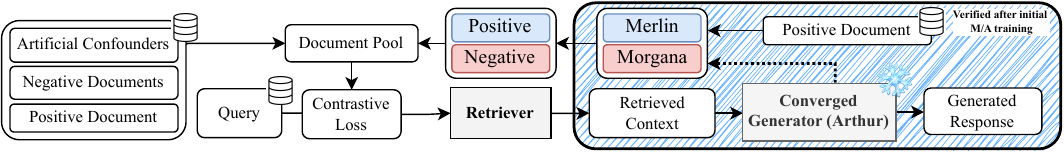}
\caption{\textbf{Automated M/A Data Augmentation for improved Retriever Training.} We use automatically generated masked contexts from a verified model $c_{\text{Me}}$ and $c_{\text{Mo}}$ (cf. Fig.~\ref{fig:ma}) to replace some of the usual hard positives and negatives and improve contrastive training.}
\label{fig:rag}
\end{figure}

The general retriever setup is depicted in Figure~\ref{fig:rag}.

\subsection{Metrics} \label{app:retriever-metrics}
For the retriever, we report Recall@$k$ and Mean Reciprocal Rank (MRR). Recall@$k$ measures how often the correct context appears within the top-$k$ retrieved results, while MRR is the mean reciprocal rank of the correct context across all queries.

\subsection{M/A Retriever Training Details}\label{app:retriever_training-neg-pool}
We provide Algorithm~\ref{alg:retriever_training} describing the retriever training.

For each query, we fill the document pool with:
% of positive and hard-negative documents:
(1) the gold document (positive),
(2) randomly sampled unrelated documents close to gold document (negatives),
% (3) topic-related negatives without the answer (if provided by the dataset),
(3) \emph{artificial confounders} derived from the ground-truth document by modifying the ground-truth context (negatives, details below),
(4) Successfully Merlin-generated contexts $c_{\text{Me}}$ at different masking sizes (positives),
(5) Successfully Morgana-generated contexts $c_{\text{Mo}}$ at different masking sizes (negatives).
While (1)-(3) establish a hard baseline, we substitute samples of the document pool with ones of (4) and (5) for comparable M/A-supported training.

% said this now in the enumeration above.
% The ground-truth document serves as the positive example and is accompanied by a specific number of negatives. If Merlin's masked document achieves a completeness score of~1 when fed to Arthur (generating LLM), we add it to the pool of positive documents. Conversely, if Morgana's masked document receives a soundness score of~0 or elicits a rejection answer from Arthur, we replace 3 negatives with the corresponding Morgana variants.

We create \emph{artificial confounders} for datasets with ground-truth annotations by modifying the sentences containing the answer: (i) removing it, (ii) replacing it with a random sentence from an unrelated context, and (iii) character-scrambling it. These confounders retain high semantic overlap with the original document (cf. App. Fig.~\ref{fig:cosine_sims_confounders}), yielding challenging (hard) negatives and crisp training signals.

% We introduce the \emph{artificial confounders} (hard negatives) to increase the difficulty of the retrieval task and evaluate the retrievers robustness to fine-grained contextual differences. They share high semantic overlap with ground-truth documents but no longer contain the ground-truth answer. We apply this procedure to datasets with explicit answer-span annotations (SQuAD and HotpotQA). Using the annotated answer location, we extract the sentence containing the answer and apply one of three transformations: (i) remove the sentence entirely, (ii) replace it with a random sentence sampled from another context, or (iii) replace every character in the sentence with a random symbol. For each original example, we generate one “removed-sentence” confounder, ten “replaced-sentence” confounders, and ten “character-scrambled” confounders. Because these modified documents remain highly similar to the original, their embeddings typically receive higher similarity scores to the query than ordinary negatives, as shown in Appendix Figure~\ref{fig:retriever_all_context_similarities}.

For a query $q$, let $d_j^+$ denote positive documents and $d_i$ the full candidate set. We train the retriever with a standard contrastive loss:
\begin{equation}
    \mathcal{L}_{\text{InfoNCE}}
    = -\log \left(
        \frac{\sum_{j} \exp(f(q, d_j^{+}))}{
              \sum_{i} \exp(f(q, d_i))}
    \right),
    \label{eq:infonce}
\end{equation}
where $f(q,d)=\mathrm{sim}(q,d)/\tau$ uses cosine similarity between final hidden state of the $\mathbf{CLS}$ token and temperature $\tau$. 
% with the scoring function $f(q, d) = \mathrm{sim}(q, d)/\tau$, where $\tau$ is a temperature parameter, and $\mathrm{sim}(q, d)$ is the cosine similarity between the query and document embeddings. We obtain document embeddings by using the final hidden state of the special \(\mathbf{CLS}\) token as the representation of the entire text.
% Cf. App.~Algo.~\ref{alg:retriever_training} for retriever training pseudocode.

To create the negative pool of retriever documents, we use dataset-specific procedures.

For HotpotQA and SQUAD we use 2 sentence-replacement confounders, 1 sentence-removal confounder, 5 hard negative documents, and 5 random documents. For TriviaQA we use 8 random documents since there are no hard negatives provided by the dataset and artificial confounders are not possible since no answer annotations are provided.

For evaluation of SQUAD and HotpotQA, we use 10 documents with character-scrambled confounders ground-truth sentence, 10 sentence-replacement confounders, 1 sentence-removal confounder, 10 random documents, and 5 dataset hard negatives. For TriviaQA we use 30 random contexts for the evaluation.

To obtain a final evaluation metric, we average over two validation sets containing: (i) original positive and negative contexts and (ii) sentence-level masked negatives generated by masking full sentences with the same ratio used during training.

\begin{algorithm}[ht]
\caption{Retriever Training}
\label{alg:retriever_training}
\begin{algorithmic}[1]
    \Require $\mathcal{B}, \mathbf{m}_{\text{Me}}, \mathbf{m}_{\text{Mo}}$
    %\Ensure $\mathcal{L}$
    
    \State $\mathcal{B}' \gets [\,]$ \Comment{Augmented batch}
    
    \State \textbf{Phase 1: Generate masked contexts for multiple masking sizes}
    \For{each $(q, c, c_{\text{conf}}, a^{\text{true}})$ in $\mathcal{B}$}
        \State $C_{\text{Me}} \gets [\,], \quad C_{\text{Mo}} \gets [\,]$
        \For{each $(p_{\text{Me}}, p_{\text{Mo}})$ in $\text{zip}(\mathbf{m}_{\text{Me}}, \mathbf{m}_{\text{Mo}})$}
            \State $(c_{\text{Me}}, \_) \gets \textsc{MaskContext}(q, c, a^{\text{true}}, p_{\text{Me}})$
            \State $(\_, c_{\text{Mo}}) \gets \textsc{MaskContext}(q, c, a^{\text{true}}, p_{\text{Mo}})$
            \State $C_{\text{Me}}.\text{append}(c_{\text{Me}})$
            \State $C_{\text{Mo}}.\text{append}(c_{\text{Mo}})$
        \EndFor
        \State $\mathcal{B}'.\text{append}(q, c, c_{\text{conf}}, C_{\text{Me}}, C_{\text{Mo}}, a^{\text{true}})$
    \EndFor
    
    \State \textbf{Phase 2: Evaluate Arthur on boundary contexts}
    \State $\mathcal{R} \gets [\,]$ \Comment{Results list}
    \For{each $(q, c, c_{\text{conf}}, C_{\text{Me}}, C_{\text{Mo}}, a^{\text{true}})$ in $\mathcal{B}'$}
        \State $C_{\text{Me}} \gets \textsc{SortByMaskingLength}(C_{\text{Me}})$
        \State $C_{\text{Mo}} \gets \textsc{SortByMaskingLength}(C_{\text{Mo}})$
        \State $c_{\text{Mo}}^{\min} \gets C_{\text{Mo}}[1]$ \Comment{Least masked Morgana context}
        \State $c_{\text{Me}}^{\max} \gets C_{\text{Me}}[|C_{\text{Me}}|]$ \Comment{Most masked Merlin context}
        
        \State $\hat{a}_{\text{Me}} \gets A(q, c_{\text{Me}}^{\max})$
        \State $\hat{a}_{\text{Mo}} \gets A(q, c_{\text{Mo}}^{\min})$
        
        \State $\textit{complete} \gets \mathbbm{1}[\hat{a}_{\text{Me}} = a^{\text{true}}]$
        \State $\textit{reject}_{\text{Mo}} \gets \mathbbm{1}[\hat{a}_{\text{Mo}} = a^{\text{reject}}]$
        \State $\textit{sound} \gets \mathbbm{1}[\hat{a}_{\text{Mo}} = a^{\text{true}}] \lor \textit{reject}_{\text{Mo}}$
        \State $\mathcal{R}.\text{append}(\textit{complete}, \textit{sound}, \textit{reject}_{\text{Mo}})$
    \EndFor
    
    \State \textbf{Phase 3: Compute contrastive loss}
    \State $\mathcal{L} \gets 0$
    \For{$i = 1$ \textbf{to} $|\mathcal{B}'|$}
        \State $(q, c, c_{\text{conf}}, C_{\text{Me}}, C_{\text{Mo}}, a^{\text{true}}) \gets \mathcal{B}'[i]$
        \State $(\textit{complete}, \textit{sound}, \textit{reject}_{\text{Mo}}) \gets \mathcal{R}[i]$
        \State $\mathcal{P} \gets \{c\}$ \Comment{Positive contexts set}
        \State $\mathcal{N} \gets \{c_{\text{conf}}\}$ \Comment{Negative document pool}
        
        \If{$\textit{complete}$}
            \State $\mathcal{N} \gets \mathcal{N} \setminus \textsc{Sample}(\mathcal{N}, |C_{\text{Me}}|)$
            \State $\mathcal{P} \gets \mathcal{P} \cup \{ \text{elements of } C_{\text{Me}} \}$
        \EndIf
        
        \If{$\neg \textit{sound} \land \textit{reject}_{\text{Mo}}$}
        \State $\mathcal{N} \gets \mathcal{N} \setminus \textsc{Sample}(\mathcal{N}, |C_{\text{Mo}}|)$ \Comment{Update negative set}
        \State $\mathcal{N} \gets \mathcal{N} \cup \{ \text{elements of } C_{\text{Mo}} \}$
        \EndIf
        \State $\mathcal{L} \gets \mathcal{L} + \textsc{ContrastiveLoss}(q, \mathcal{P}, \mathcal{N})$
    \EndFor
    
    \State \Return $\mathcal{L}$
\end{algorithmic}
\end{algorithm}

\subsection{Retriever Experimental Details} \label{app:retriever-training-details}
We train a pre-trained retriever (no LoRA) for 4 epochs with AdamW and $4\!\times\!10^{-5}$ learning rate on 32 GPUs and effective batch size  256 (approx. 3,584 documents per step).
%We use the same dataset settings  as in the RAG experiments, additionally
We filter TriviaQA to 100--3500 character contexts. % During training we apply masking with $x=0.6$.

\subsection{M/A Retriever Training Results} \label{}

%Full retriever results are shown in Fig.~\ref{fig:retriever_metrics_full}.
%\begin{figure*}[t]
%    \centering
%     \includegraphics[width=0.65\linewidth]{figures/retriever/metrics_each_dataset_reinit.pdf}
%    \caption{\textbf{Automatically generated M/A masks improve retriever training.} Compared are sentence-level M/A augmentations with baseline training \textit{from scratch} \lp{first 10 layers from pretrained}.}
%    \label{fig:retriever_metrics_full}
%\end{figure*}

%\begin{figure*}[t]
%    \centering
%     \includegraphics[width=0.65\linewidth]{figures/retriever/metrics_avg_datasets_pretrained.pdf}
%    \caption{\textbf{Automatically generated M/A masks improve retriever training.} Compared are sentence-level M/A augmentations with baseline training from the \textit{pre-trained} retriever \lp{full model 12 layers from pretrained}. Averaged over all datasets.}
%    \label{fig:retriever_avg_metrics_pretrained}
%\end{figure*}

\begin{figure*}[t]
    \centering
    \includegraphics[width=0.65\linewidth]{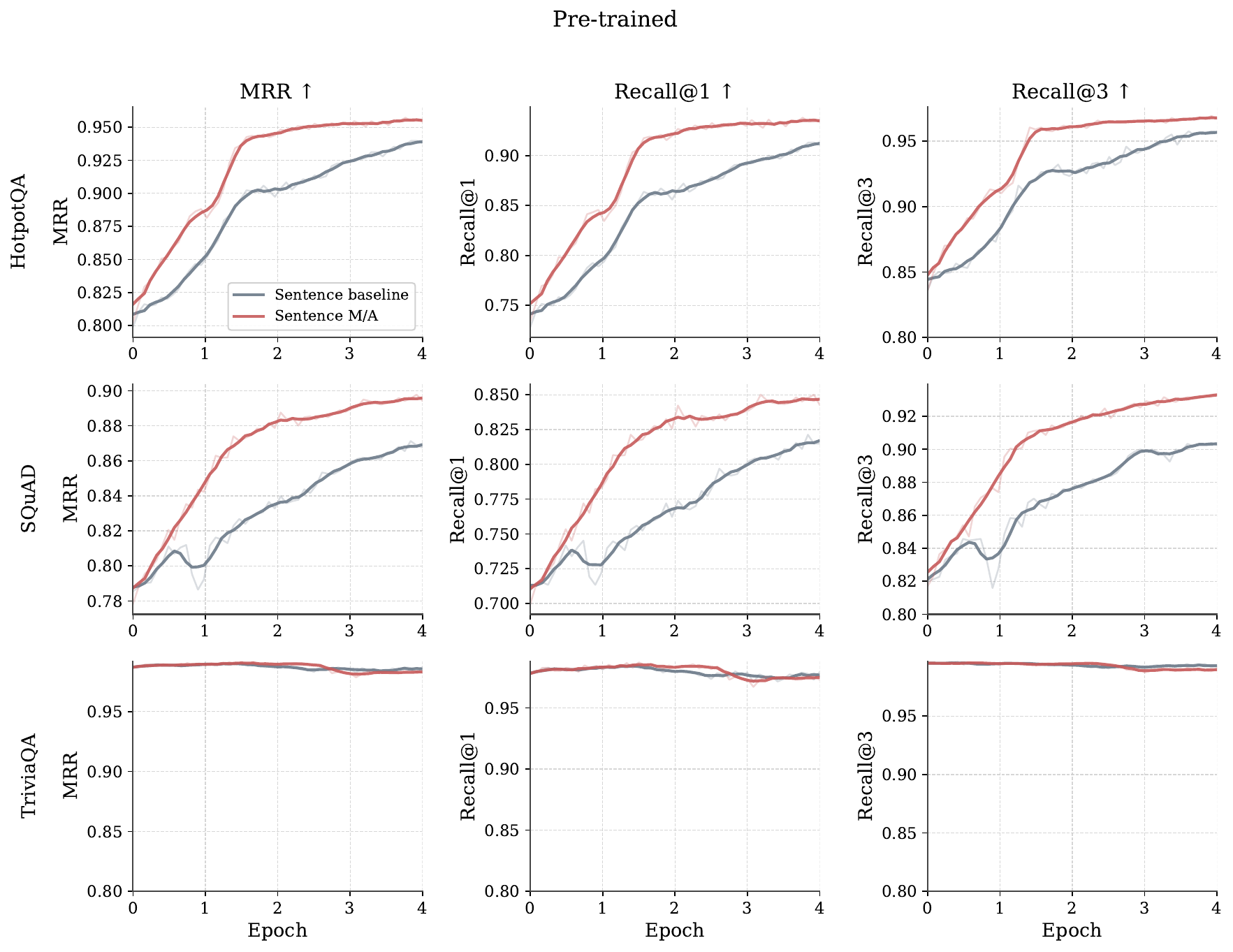}
    \caption{\textbf{Automatically generated M/A masks improve retriever training.} Compared are sentence-level M/A augmentations with baseline training -- starting from the pre-trained retriever.}
    \label{fig:retriever_metrics_full_pretrained}
\end{figure*}
%Fig.~\ref{fig:retriever_metrics_full_pretrained} shows retriever validation performance on HotpotQA, SQuAD, and TriviaQA (MRR, Recall$@1$, Recall$@3$ comparing  sentence-level M/A with baseline sentence-level training.

\begin{figure*}[t]
    \centering
    \includegraphics[width=\linewidth,trim=0.8cm 0cm 0cm 1.5cm, clip]{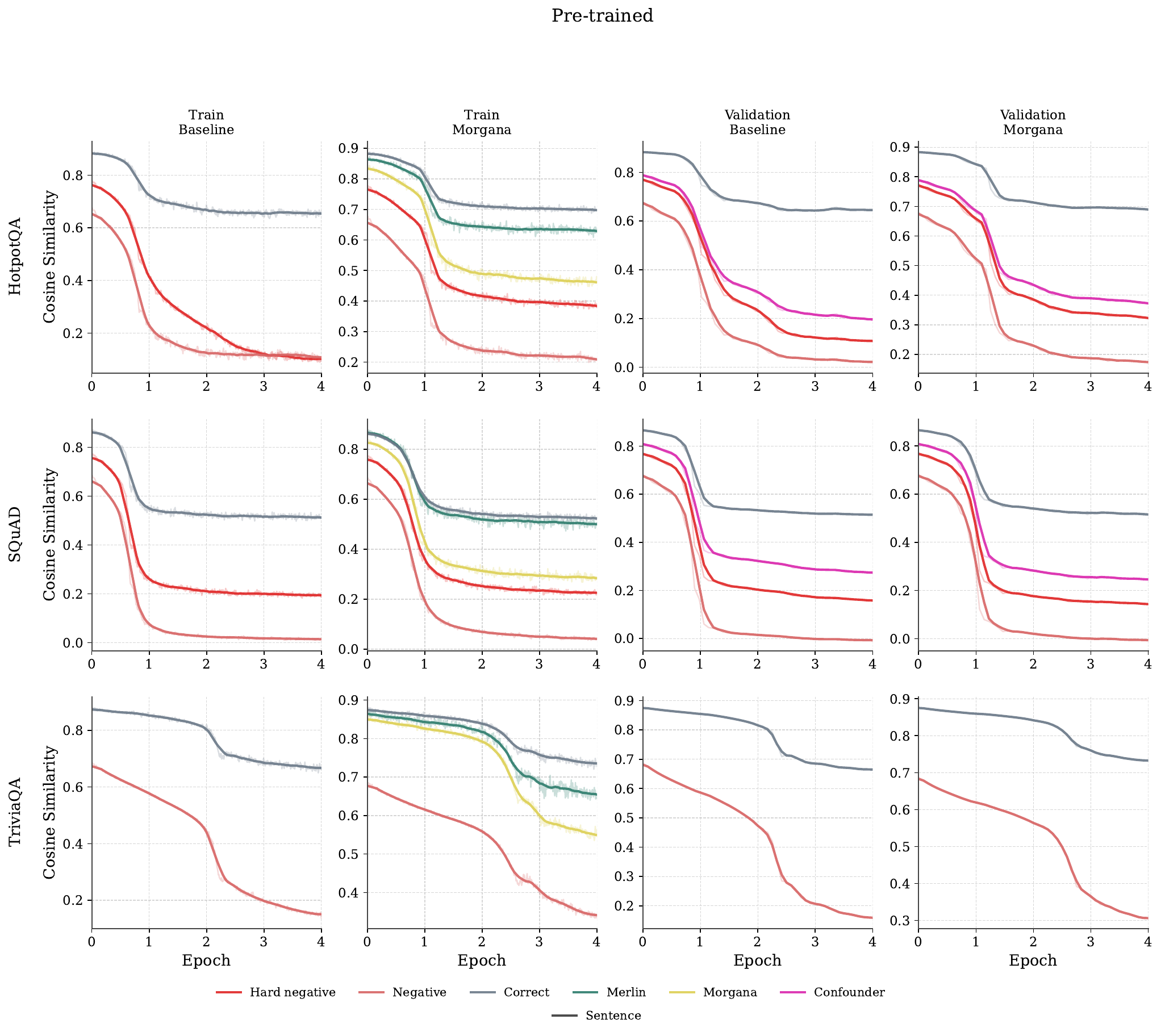}
    \caption{Cosine similarities between the query embedding and different context types during training and validation for M/A versus baseline training for a pretrained retriever. The different context types are: \emph{Correct}, meaning the original positive document for a given query, \emph{Negative}, meaning random unrelated documents; \emph{Hard negative}, meaning related documents with a higher semantic overlap than the negative documents, \emph{Merlin}, meaning the original document masked by Merlin, \emph{Morgana}, meaning the original document masked by Morgana, \emph{Confounder}: meaning the artificially created confounders via sentence manipulation, see Sec.~\ref{sec:retriever-training}. Columns show Baseline vs.\ Morgana training; rows correspond to HotpotQA and SQuAD. \\ \\
    \textbf{Discussion:} M/A training reshapes the retriever’s representation space to better separate evidence from confounders: similarities to correct documents remain high, while similarities to negatives, hard negatives, and confounders are pushed down more strongly and consistently than with baseline training -- both during training and on validation. This effect holds for sentence-level masking and generalizes across HotpotQA and SQuAD, indicating that M/A supervision yields more discriminative, robust retrieval representations rather than merely overfitting to training negatives. }
    \label{fig:retriever_all_context_similarities}
\end{figure*}

Fig.~\ref{fig:retriever_metrics_full_pretrained} compares MRR and Recall@$k$ for (i) a baseline retriever trained on ground-truth positives and standard negatives, and (ii) our M/A retriever trained with Merlin positives and Morgana-generated hard negatives instead of some of the  negatives. 
Across all metrics (Fig.~\ref{fig:retriever_metrics_full_pretrained}), the M/A retriever shows faster and significant improvement during training than the baseline retriever, despite both being trained on the same original documents.
This performance gap suggests that the inclusion of Merlin and Morgana documents provides a more informative training signal. The M/A approach enables the model to learn more robust representations that better distinguish relevant documents from challenging negatives. The Morgana confounders, in particular, appear to help the retriever develop stronger discrimination capabilities by presenting harder negative examples that share surface-level similarities with the ground-truth but lack critical information. This is further supported by the higher cosine similarities of the Morgana documents compared to other negative documents in Fig~\ref{fig:retriever_all_context_similarities}. Additionally, the Merlin documents provide another positive example, offering the retriever a richer signal for learning what constitutes a relevant document and potentially improving its robustness to variations in document quality.

\begin{figure*}[t]
    \centering
    \includegraphics[width=.9\linewidth]{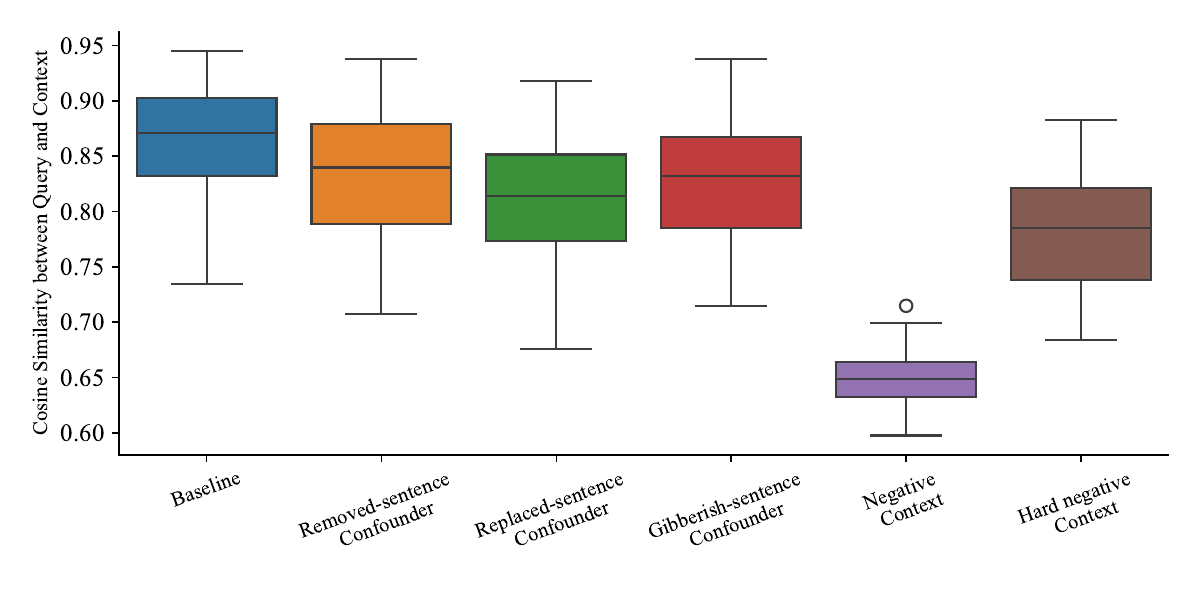}
    \caption{Cosine similarities between the query embedding and different artificial confounder types and the other provided contexts using the base pre-trained retriever model. \\ \\
    \textbf{Discussion:} Artificial confounders (removed-, replaced-, and gibberish-sentence variants) exhibit high similarity to the query,  being close to the true context labeled as \textit{Baseline} in the plot. This makes them much more difficult negatives than random negatives (labeled as \textit{Negative Context} in the plot) and also more difficult than \textit{Hard Negatives} (context sampled from the same book in SQuAD). This confirms that our confounders are genuinely challenging for the retriever and therefore well-suited to evaluate robustness improvements.}
    \label{fig:cosine_sims_confounders}
\end{figure*}

Fig.~\ref{fig:cosine_sims_confounders} shows \textbf{cosine similarities} between query embedding and different context embeddings before training computed by the \textit{granite-embedding-small-english-r2} \citep{awasthy2025graniteembeddingr2models} pre-trained retriever. Discussion in figure caption.

Fig.~\ref{fig:retriever_all_context_similarities} shows \textbf{cosine similarities} between the query embedding and different context embeddings using the \textit{granite-embedding-small-english-r2} during training and evaluation. The different contexts are: \emph{Correct}, meaning the original positive document for a given query, \emph{Negative}, meaning random unrelated documents; \emph{Hard negative}, meaning related documents with a higher semantic overlap than the negative documents, \emph{Merlin}, meaning the original document masked by Merlin, \emph{Morgana}, meaning the original document masked by Morgana, \emph{Confounder}, meaning the artificially created confounders via sentence manipulation, see Sec.~\ref{sec:retriever-training}.
We also compare the two masking settings -- sentence and token introduced in Sec.~\ref{sec:context-masking}. Discussion in figure caption.

\begin{figure*}[t]
    \centering
    \includegraphics[width=.9\linewidth]{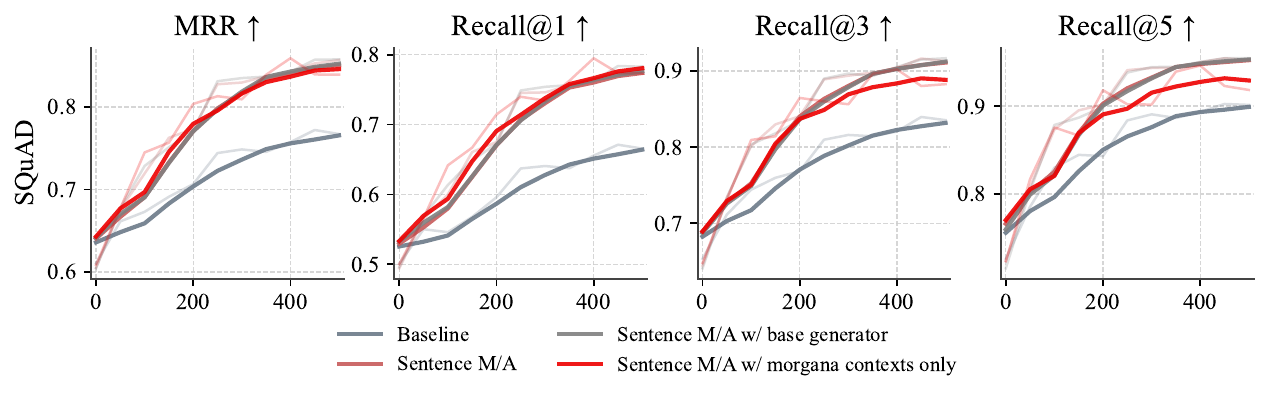}
    \caption{We compared different M/A retriever training settings on SQuAD. The setting used throughout the paper is \emph{Sentence M/A}, which uses two Merlin and two Morgana contexts which are substituting negative contexts if applicable (Morgana contexts are applicable if soundness is 0 or Arthur predicts $a^{\text{reject}}$ with the Morgana input and Merlin contexts are applicable if completeness is 1). The \emph{Sentence M/A w/ morgana only} setting only uses substituted Morgana contexts and no Merlin contexts at all. The \emph{Sentence M/A w/ base generator} setting uses the base model to generate the masks instead of a finetuned M/A model. \\ \\
    \textbf{Discussion:} Retriever training benefits most from joint Merlin-Morgana supervision: using both supportive and adversarial contexts consistently outperforms baseline training and ablations that include only one of the two. }
    \label{fig:retriever_ablation}
\end{figure*}

Figure~\ref{fig:retriever_ablation} presents an \textbf{ablation study} of retriever training on SQuAD, comparing baseline contrastive training to several variants of M/A-based retriever training. Specifically, we vary the composition of the document pool used to compute the contrastive loss by including different combinations of Merlin-generated positive contexts and Morgana-generated adversarial contexts, while keeping all other training settings fixed. Discussion in figure caption.

\begin{comment}
\begin{figure*}[t]
    \centering
    \includegraphics[width=.9\linewidth]{figures/retriever/all_metrics_all_ds.pdf}
    \caption{All retriever metrics for each dataset across the three validation dataset settings. We compare sentence- and token-level M/A with baseline sentence- and token-level  training.}
    \label{fig:retriever_all_metrics}
\end{figure*}
\end{comment}


\section{M/A aids Interpretability}

We present the following masks and attribution maps comparing three model states: (i) after token-level M/A training, (ii) before training, and (iii) after vanilla finetuning.

\begin{figure*}[t]
    \centering
    \includegraphics[width=\linewidth]{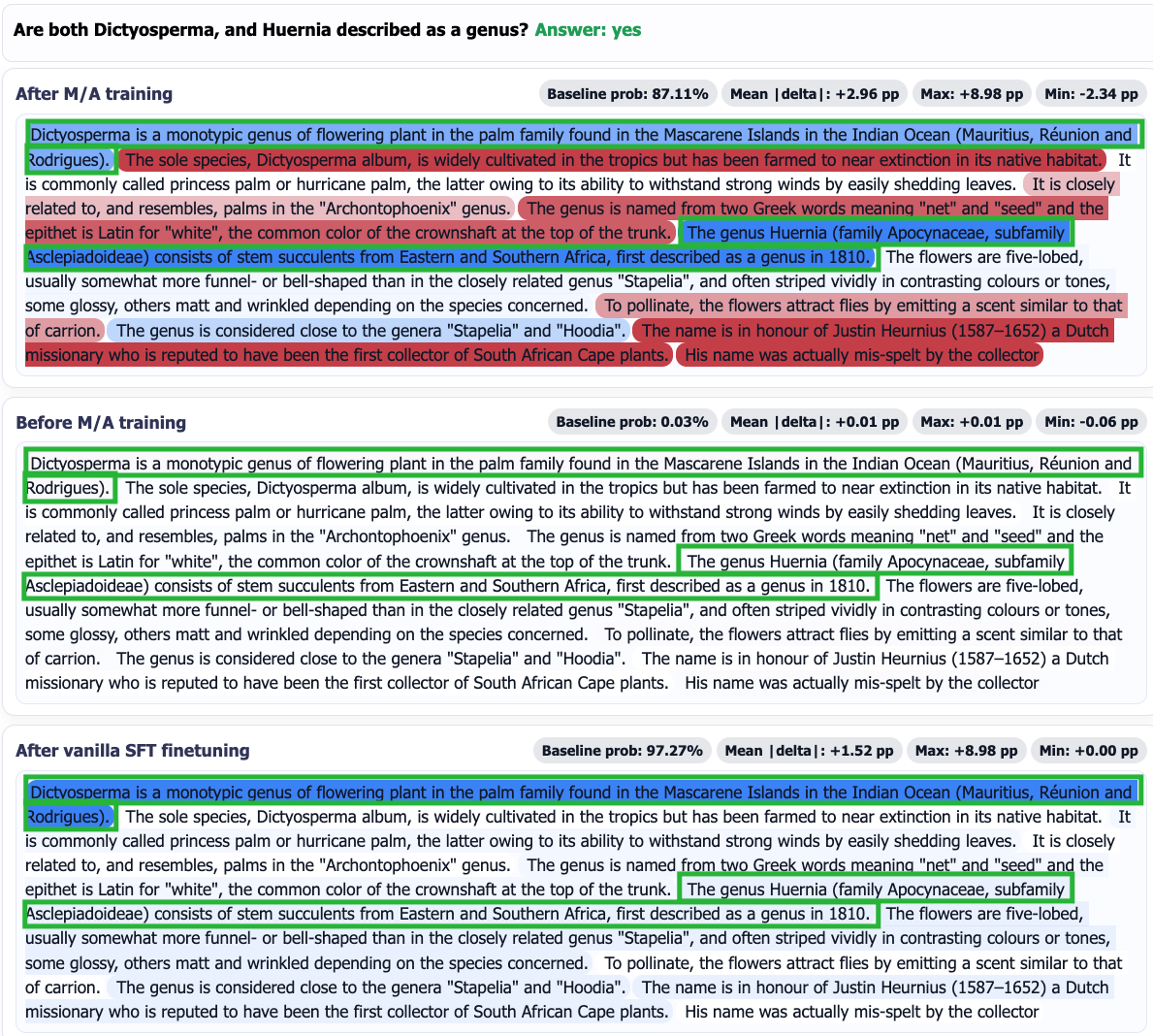}
    \caption{\textsc{AtMan} sentence level interpretation of the correct answer of  Llama-3.2-1B-Instruct on a multiple-choice version of HotpotQA. \textbf{Red}: probability increase when token is suppressed (distracting evidence).
    \textbf{Blue}: probability decrease when suppressed (helpful evidence). We compare the model (i) after \textbf{token-level} M/A training, (ii) before training, (iii) after vanilla finetuning. The \emph{baseline prob} indicates the probability the model assigned to the correct answer given the context. The \emph{mean $\Delta$} is the average change in this probability when masking tokens one after the other. We also list the minimum and maximum probability change induced by any single-token mask.\\ \\
    \textbf{Discussion:} Before training, the model fails to select any of the two relevant sentences to answer the question. After vanilla finetuning, the model selects only one of the two relevant sentences. After M/A training, both relevant sentences are selected (blue), demonstrating the model's increased grounding.}
    \label{fig:atman_hotpot_sentence}
\end{figure*}

\begin{figure*}[t]
    \centering
    \includegraphics[width=\linewidth]{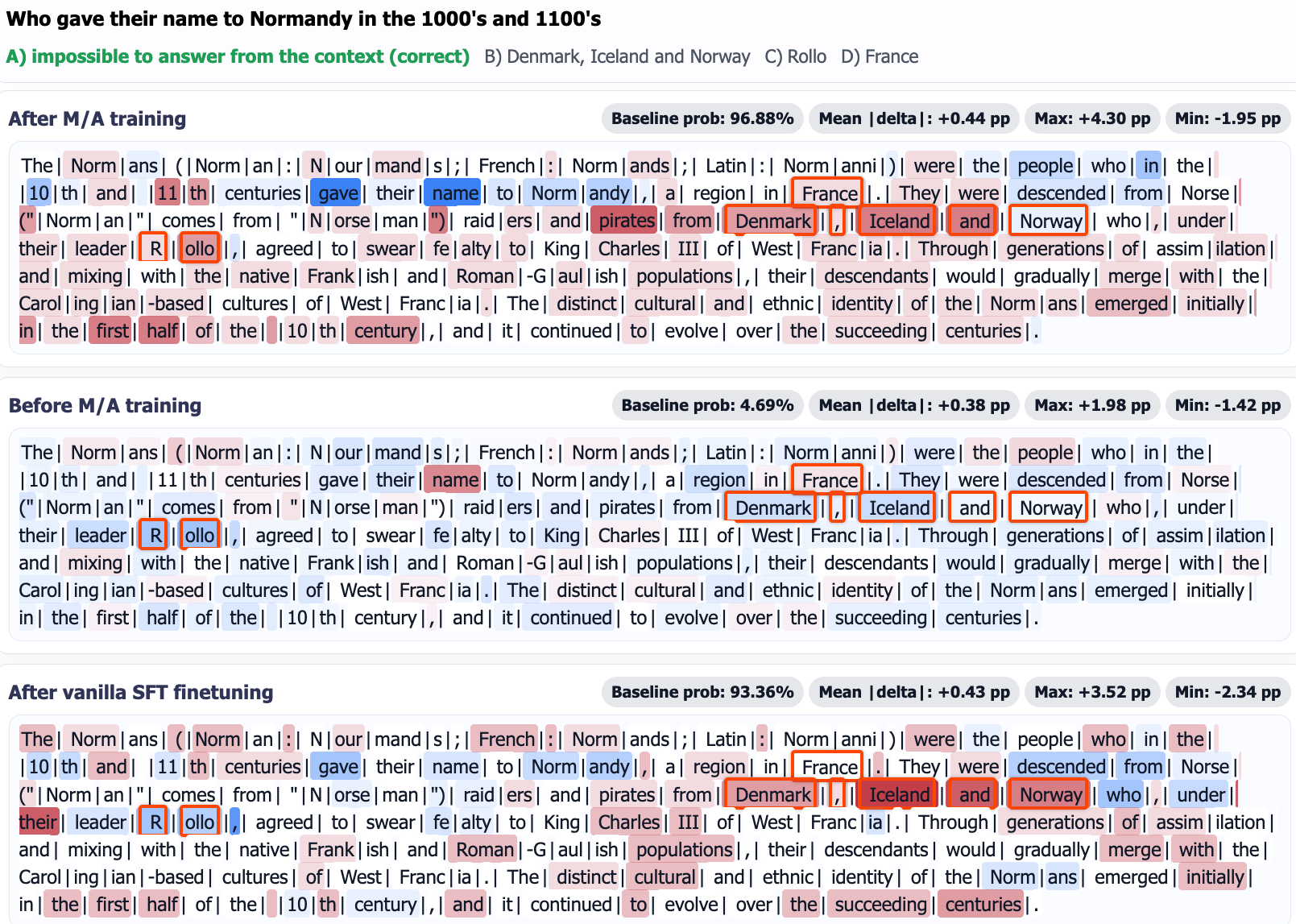}
    \caption{\textsc{AtMan} interpretation of the correct answer (\emph{reject}) of  Llama-3.2-1B-Instruct on a multiple-choice version of SQuAD. \textbf{Red}: probability increase when token is suppressed (distracting evidence).
    \textbf{Blue}: probability decrease when suppressed (helpful evidence). We compare the model (i) after \textbf{sentence-level} M/A training, (ii) before training, (iii) after vanilla finetuning. The \emph{baseline prob} indicates the probability the model assigned to the correct answer given the context. The \emph{mean $\Delta$} is the average change in this probability when masking tokens one after the other. We also list the minimum and maximum probability change induced by any single-token mask.\\ \\
    \textbf{Discussion:} This multiple-choice question is unanswerable. (1) \textbf{Before training}, the model increases the probability of the correct \textit{Reject} response, but does so for the wrong reason: it relies on misleading evidence. Specifically, confounding answer options -- \emph{Denmark, Iceland and Norway} and \emph{Rollo} (highlighted with red boxes) -- should reduce the probability of \emph{Reject}. Instead, they receive \emph{blue} attributions, indicating that the model treats these confounders as supportive evidence. (2) \textbf{After vanilla finetuning}, this behavior only partially improves: one confounder (\emph{Rollo}) is still treated as supporting the \emph{Reject} decision, while another becomes neutral, indicating incomplete disentanglement of misleading cues. (3) \textbf{After M/A training}, all confounders consistently decrease the probability of predicting \textit{Reject}, demonstrating that the model no longer relies on misleading context and instead learns to reject based on the absence of valid evidence.}    \label{fig:atman_squad_mc_token_reject}
\end{figure*}

\begin{figure*}[t]
    \centering
    \includegraphics[width=\linewidth]{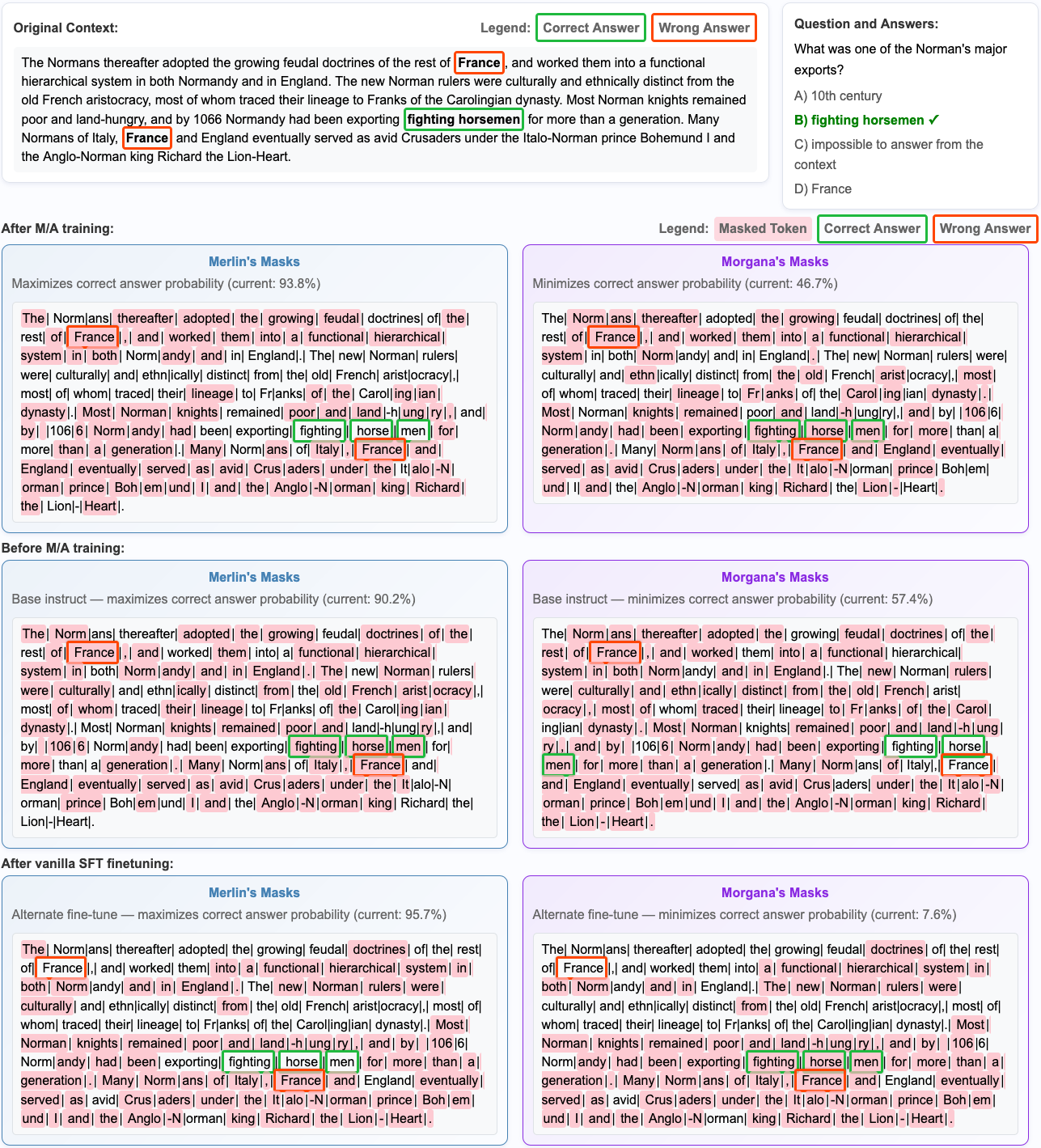}
    \caption{Merlin and Morgana \textbf{token}-level masks after token-level M/A training on a SQuAD validation example using Llama-3.2-1B-Instruct. 
    Top: original context, query, and highlighted correct and incorrect answers. 
    Middle-to-bottom: (i) Merlin and Morgana masked contexts \emph{after} M/A training, (ii) corresponding masks \emph{before} training, (iii) masks \emph{after} vanilla finetuning (baseline). 
    Red highlights denote masked tokens; green boxes indicates the correct answer span. Each panel reports the model's probability of predicting the correct answer under the shown mask. \\ \\
    \textbf{Discussion}: After M/A training, Merlin isolates evidence-bearing spans and masks confounders, reflecting improved grounding (cf. “\emph{fighting horsemen}”). Before training and after vanilla finetuning, Merlin does not remove all misleading evidence. Before training, Morgana fails suppressing the correct evidence.}
    \label{fig:squad_mc_token_sample_horsemen}
\end{figure*}

\begin{itemize}[leftmargin=*]
    % \item Fig.~\ref{fig:atman_squad_mc_token}: \textsc{AtMan} interpretation of the correct answer produced by Llama-3.2-1B-Instruct on a multiple-choice version of SQuAD. 
    
    \item Fig.~\ref{fig:atman_hotpot_sentence}: \textbf{\textsc{AtMan} interpretation} of a correct prediction 
    by Llama-3.2-1B-Instruct on a validation sample of HotpotQA.

    \item Fig.~\ref{fig:atman_squad_mc_token_reject}: \textbf{\textsc{AtMan} attribution} of a correct \textit{reject} answer on a multiple choice variant of SQuAD validation example as predicted by Llama-3.2-1B-Instruct.

    \item Fig.~\ref{fig:squad_mc_token_sample_horsemen}: \textbf{Masks as produced by Merlin and Morgana} at token-level on a SQuAD validation example using Llama-3.2-1B-Instruct.

    % \item Fig.~\ref{fig:squad_mc_token_sample_reject-rollo} Merlin and Morgana \textbf{token}-level masks on a SQuAD validation sample where the correct answer is \emph{reject} using Llama-3.2-1B-Instruct.

    % \item Fig.~\ref{fig:squad_mc_token_sample_deabolis} Merlin and Morgana \textbf{sentence}-level masks on a SQuAD validation example using Llama-3.2-3B-Instruct.
\end{itemize}

\textbf{Discussion in Figure captions.}

% \begin{figure*}[t]
%     \centering
%     \includegraphics[width=\linewidth]{figures/interpretability/atman_squad_llama1b_open_token_v2.png}
%     \caption{\textbf{M/A focuses on true evidence (green boxes), while baselines mix relevant and irrelevant context.}
%     \textsc{AtMan} attributions for  Llama-3.2-1B-Inst SQuAD (multiple-choice). Colors reflect probability change when suppressing a token (\textbf{blue}: helpful, \textbf{red}: distracting). \emph{Baseline prob} is the correct answer probability; \emph{mean $\Delta$} and \emph{range} report the average and extrema in probability changes.
%     }
%     \label{fig:atman_squad_mc_token}
% \end{figure*}

\clearpage

\end{document}